\RequirePackage{fix-cm}
\documentclass[smallcondensed]{svjour3}     
\smartqed  
\usepackage{graphicx}
\usepackage[misc]{ifsym}
\usepackage{amsmath,amssymb,amsfonts}
\usepackage{algpseudocode}
\usepackage[linesnumbered,ruled,vlined]{algorithm2e}
\usepackage[title]{appendix}
\usepackage{comment}
\usepackage{subcaption}
\usepackage[table,xcdraw]{xcolor}
\usepackage{graphicx}
\usepackage[american]{babel}
\usepackage{csquotes}
\usepackage[utf8]{inputenc}
\usepackage[T1]{fontenc}
\usepackage{siunitx}
\usepackage{romannum}
\usepackage[round,authoryear]{natbib}
\usepackage{hyperref}
\def\makeheadbox{\relax}

\newcommand\repolink{\url{https://github.com/Data-Science-in-Mechanical-Engineering/SPACE}}
\newcommand*{\affaddr}[1]{#1} 
\newcommand*{\affmark}[1][*]{\textsuperscript{#1}}

\begin{document}

\title{Scale-Preserving Automatic Concept Extraction (SPACE)}



\author{
Andrés Felipe Posada-Moreno\protect\affmark[*,1]
\and Lukas Kreisköther\protect\affmark[*,1]
\and Tassilo Glander\protect\affmark[2]
\and Sebastian Trimpe\protect\affmark[1]
}



\institute{
\Letter $\;\;$ Andrés Felipe Posada-Moreno \\ \email{andres.posada@dsme.rwth-aachen.de}\\\\
Lukas Kreisköther \\ \email{lukas.kreiskoether@dsme.rwth-aachen.de}\\\\
Sebastian Trimpe \\ \email{trimpe@dsme.rwth-aachen.de}\\\\
Tassilo Glander \\ \email{tassilo@deevio.ai}\\\\
\affaddr{\affmark[1]Institute for Data Science in Mechanical Engineering (DSME), RWTH Aachen University, Germany.}\\
\affaddr{\affmark[2]Deevio GmbH, Germany.}\\
{\textsuperscript{*}Both authors contributed equally to this work.}
}

\date{Received: date / Accepted: date}

\maketitle

\begin{abstract}
Convolutional Neural Networks (CNN) have become a common choice for industrial quality control, as well as other critical applications in the Industry 4.0.
When these CNNs behave in ways unexpected to human users or developers, severe consequences can arise, such as economic losses or an increased risk to human life.
Concept extraction techniques can be applied to increase the reliability and transparency of CNNs through generating global explanations for trained neural network models.
The decisive features of image datasets in quality control often depend on the feature’s scale; for example, the size of a hole or an edge.
However, existing concept extraction methods do not correctly represent scale, which leads to problems interpreting these models as we show herein.
To address this issue, we introduce the Scale-Preserving Automatic Concept Extraction (SPACE) algorithm, as a state-of-the-art alternative concept extraction technique for CNNs, focused on industrial applications.
SPACE is specifically designed to overcome the aforementioned problems by avoiding scale changes throughout the concept extraction process.
SPACE proposes an approach based on square slices of input images, which are selected and then tiled before being clustered into concepts.
Our method provides explanations of the models’ decision-making process in the form of human-understandable concepts.
We evaluate SPACE on three image classification datasets in the context of industrial quality control.
Through experimental results, we illustrate how SPACE outperforms other methods and provides actionable insights on the decision mechanisms of CNNs.
Finally, code for the implementation of SPACE is provided.

\end{abstract}
\section{Introduction}
\label{sec:intro}
Convolutional Neural Networks (CNN) are increasingly being used in critical applications such as self-driving cars \citep{10.1145/3003715.3005465} and automatic quality control \citep{8954181}. Their outstanding success arises from their powerful approximation capabilities \citep{Lin2017}, which traditionally has come at the cost of missing transparency and the hazard of unexpected behaviors. As the consequences in critical systems can be severe, ensuring a better understanding of the CNNs and their reliability becomes essential. 

We focus on the domain of automatic quality control. In this domain, image classification of defective and non-defective parts ensures quality in mass production processes \citep{LandingAI}. As an example, in the process of metal casting, CNNs can be used to classify good parts from parts containing specific defects such as pinholes. These defects are directly related to the structural integrity of the parts, and a misdetection of a faulty part can lead to economic consequences (e.g., a broken machine or a collapsed structure) or even put human life at risk (e.g., a faulty brake disk in a car). We refer to unexpected behavior when these misdetections are caused by biases in the models which are not aligned with human expectations. For example, a detection model could classify a part as defective by focusing on the background or the serial number of a part instead of focusing on the defect. This is clearly undesirable and dangerous.

To detect unexpected behaviors in a human-understandable way, the field of eXplainable Artificial Intelligence (XAI) \citep{DBLP:journals/corr/abs-2006-11371,arrieta2020explainable,10.1613/jair.1.12228} proposes the usage of \textit{concepts}. In this work, a \textit{concept} refers to an abstract idea, represented by a set of images, sharing a specific semantic meaning. As an example, consider the concept of a scratch. Scratches can take many appearances and forms, yet they still share the same semantic meaning. The task of concept extraction consists in analyzing a model to obtain sets of images representing concepts, which are of importance for the model. These concepts can then be presented to a human to better understand, which useful representations were learned by a model to solve a specific task. Therefore, it is an important tool in providing insights into neural networks, which often are considered black boxes.

In the context of automated quality control, current methods have two considerable flaws. First, they focus on analyzing scale-agnostic models and thus rely on interpolation techniques (e.g., bicubic interpolation) for resizing features. In contrast, quality control setups often provide static perspectives of parts, yielding data and models where scale is an important factor of the emerging image features. For example, holes or curves of different sizes may correspond to normal features or defects depending on their size. Second, existing methods rely on segmentation/superpixel (e.g., SLIC \citep{6205760}) techniques during the concept extraction process. 
These techniques segment patches alongside intensity and color boundaries (e.g, bents, curves, edges), which generates a loss of context around the segmented areas, and can introduce artifacts when padding is applied afterwards. As our empirical results highlight, both issues are significant, leading to unreliable results. We address this gap by proposing a concept extraction method that can deal with typical features of industrial datasets (e.g., small defects such as holes) and performs correctly when applied to models sensitive to changes in scale.

\subsection{SPACE: Main idea and use}
\label{sec:intro-space}

\begin{figure*}[!t]
\begin{subfigure}{\textwidth}
\centering
\includegraphics[width=1.0\textwidth]{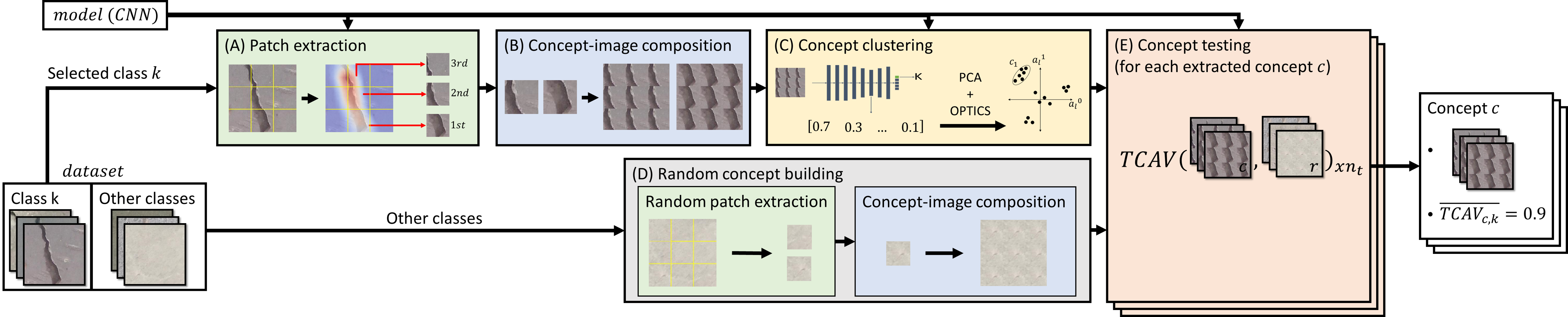}
\caption{SPACE}
\label{fig:SPACE}
\end{subfigure}


\begin{subfigure}{\textwidth}
\centering
\includegraphics[width=1.0\textwidth]{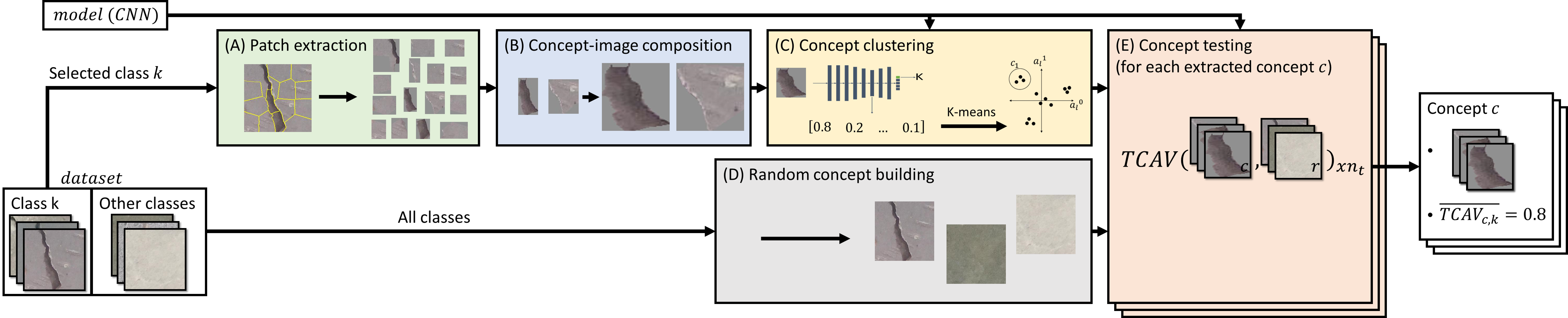}
\caption{ACE}
\label{fig:ACE}
\end{subfigure}

\caption{Visual comparison between (a) SPACE (proposed method) and (b) ACE (state-of-the-art). Both follow a similar structure of (A) \textit{patch extraction}, (B) \textit{concept-image composition}, (C) \textit{concept clustering}, (D) \textit{random concept building}, and (E) \textit{concept testing}. However, for steps (A) and (B) SPACE introduces new principles to avoid scaling through interpolation. In (C), SPACE introduces the usage of PCA and OPTICS to extract non-spherical clusters. Finally, (D) is modified to be consistent with (A) and (B).}
\label{fig:ace-space}
\end{figure*}

In this work, we propose the novel algorithm \textit{Scale-Preserving Automatic Concept Extraction} (SPACE). Our algorithm builds up on the state-of-the-art for image-based concept extraction, ACE, \citep{NEURIPS2019_77d2afcb} and introduces significant modifications presented in Section \ref{sec:SPACE}. Through experiments, we show that it achieves superior results in the context of automatic quality control. SPACE takes as inputs a CNN model, a dataset containing images for all classes that the model was trained on, and a set of (hyper-)parameters (including, for example, the class index as defined in the model output), as described in Figure \ref{fig:SPACE}. SPACE returns a set of meaningful concepts, where each concept consists of a set of example images (concept-images) and a corresponding importance score.

As an application example, consider the case of detecting pinholes on a metal part (details follow in Section \ref{sec:results-metal casting}). 
First, an image dataset of non-defective and defective parts is obtained. Second, a CNN is trained to perform the intended classification task. Third, an expert may use SPACE to validate which concepts are being used to classify images as defective parts. The SPACE algorithm takes the trained CNN, the training dataset, and the chosen (hyper-)parameters as inputs and returns multiple concepts represented through example images and importance scores. Then, a human domain expert may visually inspect the examples of each concept to interpret what concepts are important for the model to solve the classification task. Next, the human can verify if the concepts with high importance are aligned with the intended task (e.g., pinholes), or point towards biases which can generate unexpected behaviors (e.g., background colors, or thickness of an unrelated line). These results can be used by the human expert to identify biases, data leakages and overfitting. Finally, the dataset can be adjusted to remove the discovered biases or data leakages.

In more detail, the five steps of SPACE are (c.f., Figure \ref{fig:ace-space}): (A) first, \textit{patch extraction} takes as input the images of the class under inspection; then, the images are sliced into square patches, and a subset of the patches is selected based on their aggregated pixel importance. (B) Then, \textit{concept-image composition} takes each extracted patch, and stacks it vertically and horizontally (tiling) to obtain images of the input size of the model. (C) Next, in \textit{concept clustering}, concept-images are encoded in the latent space of the model, then the dimensionality of the encoding is reduced through \textit{principal component analysis} (PCA) \citep{10.1613/jair.1.12228}, and clusters are extracted using \textit{ordering points to identify the clustering structure} (OPTICS) \citep{10.1145/304182.304187}. (D) After this, during \textit{random concept building}, the rest of the dataset (excluding the class under inspection) is used to randomly extract square patches similar to step (A) and compose concept-images as in step (B). These sets of random concept-images are needed for the importance score calculation of the next step. (E) Finally, in \textit{concept testing}, each extracted and random concept is sampled multiple times and tested using \textit{testing with concept activation vectors} (TCAV) \citep{kim2018interpretability}, obtaining a mean TCAV score for each concept. The higher the importance score, the more sensitive the model is to the concept for predicting the class under inspection.

\subsection{Contributions}
\label{sec:intro-contribution}

In essence, this paper makes the following contributions:
(\romannum{1}) We propose the SPACE algorithm as a new concept extraction method for the global interpretability of CNNs.
(\romannum{2}) We compare SPACE to the state-of-the-art method, ACE, using three quality control use cases.
(\romannum{3}) We present an application on real-world manufacturing data and show how SPACE allows for a better understanding of trained models, as well as a more reliable detection of defects.

In summary, this paper tackles the gap of creating a concept extraction method capable of analyzing scale-sensitive models and data containing features that are small or related to the morphological structure of the images. Our method provides a tool for concept extraction in the context of automatic quality control.

A Python implementation of the SPACE algorithm based on TCAV \citep{kim2018interpretability} is available at:\\ \repolink.

\subsection{Outline}
\label{sec:intro-outline}

The paper continues with the related work, exposing the current alternative methods and the overall scientific context (Section \ref{sec:relatedwork}). Then, the required background of concept extraction is described (Section \ref{sec:background}). Next, the algorithm SPACE is introduced, and the underlying mathematical ideas explained (Section \ref{sec:SPACE}). Consecutively, SPACE is compared with ACE through empirical studies over three datasets (Section \ref{sec:results}). Finally, conclusions and future perspectives are presented (Section \ref{sec:conclusions}).

\section{Related work}
\label{sec:relatedwork}

This work falls into the general field of XAI, which we briefly discuss next. We start by introducing the general objective of the field, and the taxonomy of explanations. Within the field of XAI, we locate the methods of concept extraction, which encompass the current work. Afterward, we give a brief introduction to concept extraction methods and position our work in this context.

\subsection{eXplainable Artificial Intelligence (XAI)}
\label{sec:relatedwork-xai}

The field of XAI proposes either the creation of AI models which are human-understandable, or methods that allow humans to better understand the decision-making processes of existing models \citep{DBLP:journals/corr/abs-2006-11371,arrieta2020explainable,10.1613/jair.1.12228}. XAI methods can be classified in terms of their scope (local or global) and usage (ante-hoc or post-hoc) \citep{DBLP:journals/corr/abs-2006-11371}. The current work focuses on algorithms that explain a model as a whole (global scope) and are used to analyze trained models without having to modify their architecture (post-hoc usage).
In this context, global explanation methods for CNNs have focused on extracting decision-making patterns that can describe the behavior of a model in a human-understandable way \citep{arrieta2020explainable}. The two paradigms that have been proposed are to distill a simpler model which is human-understandable (e.g. rule-based \citep{Augasta2012,Zhou2003}, decision trees \citep{8953917,8999213,10.1007/978-3-319-46307-0_29,938448}, fuzzy systems \citep{DBLP:journals/corr/abs-2010-04974}), or to extract sets of inputs which generate similar activations inside the models and have global relations with resulting predictions such as ACE \citep{NEURIPS2019_77d2afcb}, CaCe \citep{goyal2019explaining}, and VRX \citep{DBLP:journals/corr/abs-2105-00290}. The current work focuses on the latter, more specifically on the extraction of concept-based explanations \citep{NEURIPS2019_77d2afcb,goyal2019explaining,NEURIPS2020_ecb287ff}.

\subsection{Concept extraction}
\label{sec:relatedwork-conceptextraction}

The extraction of concepts to make AI models explainable has only recently been proposed.
A first approach in this direction was proposed as an ante-hoc method named ProtoPNet \citep{DBLP:conf/nips/ChenLTBRS19}, where the CNN architecture was specifically designed to provide prototypical regions as explanations, which are analog to concepts.
On the field of post-hoc methods, the first work in this direction is by Ghorbani et al., with the method ACE \citep{NEURIPS2019_77d2afcb}. This work was directly linked to the concept testing algorithm TCAV \citep{kim2018interpretability}. Both methods (ACE and TCAV) will be discussed in more detail in Section \ref{sec:background}. Since its recent proposal, three main working principles have been introduced for post-hoc concept extraction and concept-based explanations. The main approaches have been based on autoencoders (e.g. CaCe \citep{goyal2019explaining}, VAEs \citep{DBLP:conf/fruct/UtkinDKK21}), changing a model's architecture to constrain certain layers (e.g. concept bottleneck models \citep{pmlr-v119-koh20a}, concept whitening \citep{DBLP:journals/corr/abs-2002-01650}), or patch extraction (e.g. ACE \citep{NEURIPS2019_77d2afcb}, VRX \citep{DBLP:journals/corr/abs-2105-00290}, conceptShap \citep{NEURIPS2020_ecb287ff}). 

Our work is directly related to the patch extraction approaches, which encompass methods based on the extraction of regions/patches from the input images, then clustering and testing the obtained concepts. The first introduction to region-based explanations was first proposed in XRAI \citep{9008576}. Afterwards, ACE \citep{NEURIPS2019_77d2afcb} focused on extracting patches through superpixel techniques such as SLIC \citep{6205760}. Similarly, VRX \citep{DBLP:journals/corr/abs-2105-00290} proposes the usage of gradients to guide the patch discovery of ACE. Other similar approaches such as EFC-CAM \citep{9405672} have introduced the usage of gradients to obtain regions specific to a class. Our current work also proposes a step of patch extraction, yet, we argue that by changing the way patches are extracted, encoded, and tested, better results can be obtained in applications related to quality control.

From the mentioned algorithms, ACE will be taken as the state-of-the-art of concept extraction techniques. Thus, it will be further introduced alongside the testing method TCAV in Section \ref{sec:background}. Additionally, ACE will be used as an experimental baseline in Section \ref{sec:results} to obtain an insightful comparison.
\section{Background}
\label{sec:background}

In this section, we introduce three fundamental components required for the correct understanding of SPACE. Specifically, we describe the general methods for concept testing and automatic extraction.
First, we explain the method TCAV for concept testing, which is a supervised approach for testing whether a CNN is sensitive to a concept or not. 
Then, we lay out the general mechanisms of the ACE algorithm, introduced by Ghorbani et al. \citep{NEURIPS2019_77d2afcb}.



\subsection{Testing with Concept Activation Vectors (TCAV)}
\label{sec:background-tcav}

TCAV was proposed \citep{kim2018interpretability} to provide an interpretation of a CNNs internal state employing human-understandable concepts. 
The key idea is to assess whether two sets of images (containing or not a human-defined concept) generate different activations within the CNN, and how much this difference contributes to the prediction process of a CNN.
For this, the method proposes to encode each image as a flattened activation map $a_l=f_l(x)$ (obtained through the partial evaluation of the CNN until layer $l$). 
Then, a linear classifier is trained to differentiate the two groups of encoded images. As a result, the fitted parameters of the linear classifier denote a vector normal to the decision boundary (hyperplane) of the classification. This vector is then used to define the concept and is named Concept Activation Vector (CAV).
Finally, a sensitivity metric $S_{C,k,l}(x)$ is introduced to express how sensible the prediction of the class $k$ for the image $x$ is towards the defined concept $C$,
\begin{equation}
\label{eqn:TCAVsensitivity}
S_{C,k,l}(x) = \nabla h_{l,k}(f_l(x))\cdot v_{C}^{l} .
\end{equation}
Here, $h_{l,k}(f_l(x))$ is the predicted logit for the class $k$, obtained by evaluating the activation map $a_l=f_l(x)$ from the layer $l$ onwards. Finally, a score $TCAV_{Q_{C,k,l}}$ is introduced to quantify the global importance of a concept for the predictions of a specific class. This score is defined as the ratio of images $x$ of class $k$ ($X_k$) with a positive sensitivity for a selected concept $C$,
\begin{equation}
\label{eqn:TCAVscore}
TCAV_{Q_{C,k,l}}= \frac{\left | \left \{ x \in X_{k} : S_{C,k,l}(x) > 0 \right \} \right |}{\left |X_{k}\right |}  .
\end{equation}

\subsection{Automatic Concept-based Explanations (ACE)}
\label{sec:background-ace}

Based on TCAV, the first automatic concept extraction method was introduced by Ghorbani et al. \citep{NEURIPS2019_77d2afcb}. The general method of ACE is described in Figure \ref{fig:ACE}, and is composed of five steps. (A) First, a set of images from a class are given, and each image is segmented with multiple resolutions using SLIC \citep{6205760}. (B) Then, each one of the extracted patches/superpixels is padded to a rectangular shape with the mean pixel values and then resized by bicubic interpolation to the original size of the image. (C) Later, the activation maps of the resized patches are flattened and clustered using k-mean to obtain a defined number of concepts. Next, (D) sets of random concepts are created by selecting a subset from a pool of random images. Finally, (E) the importance score of each concept is calculated using TCAV \citep{kim2018interpretability}.

\section{SPACE}
\label{sec:SPACE}

We propose the technique Scale-Preserving Automatic Concept Extraction (SPACE) to tackle specific challenges of industrial datasets related to applications such as quality control and predictive maintenance. SPACE, as described in Figure \ref{fig:SPACE}, represents an alternative method for the automatic extraction of concept-based explanations using a distinctive patch extraction technique, coupled with saliency maps to assess local importance of extracted patches and the usage of tiling for patch resizing. A central algorithmic requirement that considered for the SPACE algorithm was the preservation of scale of all features while extracting and evaluating concepts. Because the scale of potentially meaningful concepts of industrial datasets often critically determines their actual semantic meaning. 

As described previously, SPACE follows the process of concept extraction as shown in Figure \ref{fig:ace-space}. SPACE takes as input: a dataset $E:\left\{ (x_{1},y_{1}),...,(x_{N},y_{N}) \right\}$ of labeled data points; a trained CNN $f_\mathrm{M}$; the index $k$ of the class to be analyzed; the number $n_\mathrm{s}$ of horizontal and vertical slices used for patch extraction; the percentage $n_\mathrm{p}$ of patches to be extracted from each image; the number of PCA components $n_\mathrm{pca}$ to reduce activation maps dimensionality before clustering; the layer $l_\mathrm{gradcam}$ to perform Grad-CAM; the layer $l_\mathrm{activ}$ used to extract activations for the clustering and testing. As output, SPACE returns a set of concepts $\{ (\varepsilon_{0}, \overline{S}_0),(\varepsilon_{1}, \overline{S}_1), ...,(\varepsilon_{n-1}, \overline{S}_{n-1})\}$, where $\varepsilon_{i}$ is a set of examples for the concept $c_i$, and $\overline{S}_i$ is the importance of said concept with relation to the class $k$. In this section, we introduce SPACE's five functional steps as shown in the Algorithm \ref{alg:SPACE_short}.
\begin{algorithm}
    \caption{Scale-Preserving Automatic Concept Extraction, \texttt{SPACE}}
    \label{alg:SPACE_short}
    \SetAlgoLined
    \SetKwInOut{Input}{input}
    \SetKwInOut{Output}{output}
    \DontPrintSemicolon
    \Input{
    $E:\left\{ (x_{1},y_{1}),...,(x_{N},y_{N}) \right\}$,
    $f_\mathrm{M}$,
    $k$,
    $n_\mathrm{s}$,
    $n_\mathrm{p}$,
    $n_\mathrm{pca}$,
    $l_\mathrm{gradcam}$,
    $l_\mathrm{activ}$
    }
    
    \Output{
    Set of concepts $\{ (\varepsilon_{0}, \overline{S}_0),(\varepsilon_{1}, \overline{S}_1), ...,(\varepsilon_{n-1}, \overline{S}_{n-1})\}$
    }
    
    $E_k = \{(x_i, y_i) \in E \mid \mathrm{max\_index}(y_i)=k\}$
    
    $P \gets \mathrm{Patch\_extraction}(E_k, k, f_\mathrm{M}, n_\mathrm{s}, n_\mathrm{p}, l_\mathrm{gradcam})$
    
	$X^{*} \gets \mathrm{Concept\-image\_ composition}(P, n_\mathrm{s})$
    
	$C^{*},\{\varepsilon_{c_i} \mid c_i \in C^{*}\} \gets \mathrm{Concept\_clustering}(X^{*},f_\mathrm{M},k,l_\mathrm{activ})$
	
    $R^{*} \gets \mathrm{Random\_ concept\_ building}(E \setminus E_k, n_\mathrm{s})$
    
    $\{\overline{S}_i \mid c_i \in C^{*}\} \gets \mathrm{Concept\_testing}(C^{*},\{\varepsilon_{c_i} \mid c_i \in C^{*}\},R^{*}, l_\mathrm{activ})$
    
    return ($\left \{(\varepsilon_{c_0}, \overline{S}_0), ..., (\varepsilon_{n_c-1}, \overline{S}_{n_c-1}) \right \}$)
\end{algorithm}

\subsection{(A) Patch extraction}
The \textbf{Patch extraction} function produces a set $P$ of patches from the subset $E_k$ of images of the target class $k$. For each image $x_j$, SPACE first performs Grad-CAM \citep{8237336} to obtain the saliency map $S_j$ (pixel-wise importance). Afterwards, the image $x_j$ is sliced into \(n_\mathrm{s} \, \mathrm{x} \, n_\mathrm{s}\) windows to obtain a set $P^{*}_{x_j}$ of patches $p_{x_j,o}$, where $o$ denotes the position of the patch. Similarly, a binary mask $g_o$ the size of $x_j$ is created for each patch $p_{x_j,o}$, obtaining the set $G^{*}$. Then, the aggregated patch importance score $\psi_{j,o}=f_{\psi}(S_j, g_o)$ of each patch $p_{x_j,o}$ is computed by applying the mask $g_o$ over the saliency map $S_j$ (element-wise multiplication), summing the importance of each pixel $(a,b)$, and then dividing by the number of non-zero importance pixels,
%
%
\begin{align} \label{eq:patch-importance}
\psi_{j,o}=f_{\psi}(S_j, g_o)=\frac
{\sum_{a,b} (S_j \odot g_o)_{(a,b)}}
{\sum_{a,b} \mathit{H}(S_j \odot g_o)_{(a,b)}}
\end{align}
with
\begin{align}
\mathit{H}(x) =
  \begin{cases}
    1       & \quad x>0\,,\\
    0  & \quad  else.
  \end{cases}
\end{align}
Then, the patches of $P^{*}_{x_j}$ are ranked based on $\psi_{j,o}$, and only the top $n_\mathrm{p}$ percent of the most important patches are selected for the set $P_{x_j}$. Finally, the set $P$ of extracted patches is obtained as the union of all $P_{x_j}$ for $x_j$ in $E_k$.


In contrast with ACE, the proposed patch extraction has two advantages. First, the borders of the extracted patches are uncorrelated with intensity boundaries (e.g, edges) of the input images. This enables a more coherent extraction in the cases where these types of features are important. Second, the ranking and selection of patches centers the analysis towards the patches that have actual impact on the decision-making process of the model. The proposed aggregation approach leads to preferring patches with few pixels of high pixel-importance compared to several pixels with lower pixel-importance, leading to more robustness against noisy saliency maps and the chosen patch size. As an example, a homogeneous yet unimportant background would be filtered out in this step, instead of becoming noisy points in the context of the concept clustering step.


\subsection{(B) Concept-image composition}


The \textbf{Concept-image composition} function transforms each extracted patch $p_{t}$ from $P$ into an image $x_{p_{t}}$, yielding the set $X^{*}$. To do this, SPACE proposes to tile each square patch $p_{t}$ vertically and horizontally $n_\mathrm{s}$ times to obtain $x_{p_{t}}$. The resulting concept-image aims to trigger similar activation patterns as the images of $E$ containing similar features.


The main argument behind the use of tiling is the fact that convolutions are not scale invariant.
This translates in CNNs learning features which are not scale invariant, unless the dataset, training process, or architecture of the CNN are explicitly modified for this goal.
Other works have tried to deal with this limitation, either by adding specific architectures such as ensembles \citep{van2017learning}, using multiple columns/backbones for different scales \citep{xu2014scale}, or specific architectures such as scale pyramids for object detection \citep{kim2018parallel}.
These same works have studied how activation maps and predictions shift as the scale of input images changes. 
From another perspective, only a subset of an image is required to compute the activations at an arbitrary layer. This region is the receptive field, and, as long as the tiles are bigger than the receptive filed required to encode the important features of the dataset, tiling will allow a similar encoding for a patch in comparison to an original image.

In contrast with other methods, the resulting concept-images are not transformed with respect to their scale. This is specially important when analyzing models that are trained from a single perspective (e.g., quality control of a metal piece where the scale of the features is semantically meaningful). In these cases, the internal representations of the models are not scale-invariant and thus, any activations generated by re-scaled concept-images can differ significantly from the activations of the dataset. 

During the tiling process, discontinuities are introduced. 
This can present a challenge for our methodology in instances where classification relies on similar features
Yet, discontinuities are rarely defining features for quality control classification tasks.
In quality control tasks, images of products or objects are scrutinized to identify specific local features. 
In most instances, the visual cues associated with each class are local morphological features and not discontinuities. 
Hence, the models that are developed are typically not tuned to detect or be sensitive to such features. 
This is especially true when these discontinuities are introduced during the training process as a result of data augmentations (e.g. by using random crops of mix-based augmentations).

\subsection{(C) Concept clustering}


The \textbf{Concept clustering} function extracts clusters from the set $X^{*}$ of concept-images which are meaningful for the model. First, SPACE computes the activation maps $a_{l_\mathrm{activ},x_{p_t}}$ by partially evaluating each concept-image $x_{p_t}$, in the model $f_\mathrm{M}$ until the layer $l_\mathrm{activ}$. It then flattens the activation maps, composing the set $A^{*}$. As a result, the flattened activation maps become the encoding of the image, representing the perception of the model. Second, we perform a PCA over  $A^{*}$, and reduce the dimensionality of each element to $n_\mathrm{pca}$ components, obtaining the set $A^{*}_\mathrm{pca}$. Then, the clustering algorithm OPTICS \citep{10.1145/304182.304187} is used to identify a set $C^{*}$ of clusters based on the Manhattan distance between their elements. Each one of the extracted clusters $c_i$ becomes an extracted concept. Finally, all concept-images $x_{p_t}$ whose $a_{l_\mathrm{activ},x_{p_t}}$ belong to a concept $c_i$, are used to compose the example set $\varepsilon_{i}$.


This function implements a step of dimensionality reduction to improve the effectivity of the clustering algorithms as well as to reduce redundant information caused by previous steps. The joint application of PCA, OPTICS and the usage of the Manhattan distance allows for an effective extraction of density-based clusters consisting of coherent concept-images leading to meaningful concepts.

We chose OPTICS over alternatives such as k-means due to several factors. OPTICS can identify complex, non-spherical clusters and handle varying densities and quantities. In contrast, k-means extracts a known number of spherical clusters, struggling with density differences, or a mismatch of the assumed cluster quantity. With OPTICS, we avoid making assumptions about cluster shapes, densities, and quantities within the CNN's latent space. Prior to OPTICS, we implement PCA to manage our data's high dimensionality, thereby improving computational speed, reducing redundancy, and noise.

\subsection{(D) Random concept building}

The \textbf{Random concept building} function assembles example sets of random concepts. The items of these sets are chosen randomly, and do not have any shared meaning. SPACE proposes to use the images of the analyzed dataset, excluding the ones from the class that is being analyzed. From each image $x_j$ in $E \setminus E_k$, a random patch $p_j$ is cropped, considering the dimensions of the patches in step (a). Then, each random patch $p_j$ is used to compose concept-images $x_{p_j}$ using the same approach as step (B). The resulting set of concept-images $x_{p_j}$, becomes the example set $\varepsilon_{r_i}$ of a random concept $c_{r_i}$. As an output of this function, a defined number of random concepts $c_{r_i}$ and their example sets $\varepsilon_{r_i}$ are extracted.

In the case of ACE, the random concepts are composed by complete images, randomly selected from the dataset. This means that during the concept testing step, complete images are being compared with resized patches of other images. We argue that this comparison is flawed, and the composition of the random example sets $\varepsilon_{r_i}$ should be similar to the composition of the extracted concept example sets $\varepsilon_{i}$. It is desired to have similar transforms in the main features of the concept-images and the random images, this will minimize the risk of testing the importance of the transforms instead of the importance of the actual concepts. This is specially important for models that are not robust to changes in scale, as the activation maps generated from these images can differ significantly or, even worse, generate representations that are completely disconnected from the nominal behaviors of the model.

\subsection{(E) Concept testing}

The \textbf{Concept testing} function  is performed for each concept $c_i$ in accordance with the TCAV algorithm \citep{kim2018interpretability}, using the layer $l_\mathrm{activ}$. For each analyzed concept $c_i$, a sample of the example set $\varepsilon_{i}$ is compared with a sample of a random example set $\varepsilon_{r}$. 
The basic idea behind the testing in TCAV, is to compute multiple CAV and TCAV scores for a subset of a group \textbf{A} of patches ($\varepsilon_{i}$), and for a group of random patches \textbf{B} ($\varepsilon_{r}$). This generates a set of TCAV scores for each group, which are later compared through a two-sided t-test. It is to be noted that the expected value of the TCAV score for a set of random patches is 0.5. This 0.5 TCAV score means that for half of the images in a class, the concept affects the prediction of the CNN positively, and half of the time it affects it negatively. This procedure was introduced in detail by Kim et al. in TCAV \citep{kim2018interpretability}.
This process is repeated multiple times, obtaining the CAV and TCAV score of each repetition. Afterwards, a control group is obtained by repeatedly comparing two random concepts. Then, the two populations of TCAV scores are used to verify the significance of the findings through a two-sided t-test. The TCAV algorithm yields as result an average TCAV score, also referred as average sensitivity $\overline{S}_i$. This method of testing concepts is also used in ACE, as it allows for the testing of whether a concept described by a set of images sharing a semantic meaning is influential in the decision-making process of the model.

\section{Results}
\label{sec:results}

In this section, we present the aggregated experimental results of the proposed algorithm SPACE and its comparison with ACE.
We compare SPACE and ACE through multiple industrial datasets, three CNN architectures and ten random seeds.
We focus on datasets where scale is semantically meaningful, data is scarce, unbalanced, or the intended features are small in comparison with the main features of images.
First, we present the results of executing both SPACE and ACE over two open datasets as mentioned before, and aggregating the results based on the ratio of extracted concepts aligned with the semantic meaning of the analyzed classes, as shown in Figure \ref{fig:boxplots}. A complimentary dataset on leather classification in discussed in Appendix \ref{apd:Extended leather}.
Then, we further explore the datasets through representative cases, highlighting the main issues observed in the experimental setup. As an additional representative case, we present a real world use case on quality control for metal casting. 
We show that SPACE outperforms ACE when extracting concepts in the above defined settings.

\subsection{Concept extraction runs (SPACE and ACE)}

In our setup, a run refers to the execution of either SPACE or ACE, with a specific set of parameters, to analyze a single model. 
Each run is defined by the parameters of the concept extraction method, the parameters of the analyzed model, the dataset used for the training of the model, as well as the random seed used for the run itself.

On our experimental setup, we use two open datasets, which highlight different properties common in industrial environments.
First, the concrete crack dataset \citep{conc} provides a test case for defects spanning complete images without other major features. 
Second, the metal nut anomaly detection dataset \citep{8954181} is a test case where a specific object (roughly the size of the image) may contain a defect of a size comparable to other features in the image.
A third complementary dataset about leather anomaly detection\citep{8954181} will be discussed in Appendix \ref{apd:Extended leather}.

The used CNN architectures where VGG-16 \citep{DBLP:journals/corr/SimonyanZ14a}, ResNet-18 \citep{DBLP:journals/corr/HeZRS15}, and Densenet-121 \citep{DBLP:journals/corr/HuangLW16a}.
These architectures have significant differences in the interconnection of their layers, which influences how information propagates within them.
Each architecture was trained with ten different random seeds on each dataset.
The training and validation sets were used for training and verification of convergence, respectively, while the test data was used to assess the model accuracy after the training was over. For both training and testing the dataset was sampled using a weighted random sampler, to mitigate the effects of the unbalanced datasets. After training, the best model of each training was analyzed with both SPACE and ACE to extract its learned concepts. The overall performance metrics of these models are presented in Appendix \ref{apd:Extended concrete cracks} and \ref{apd:Extended metal nut}.

\begin{figure}[h]
\centering
\begin{minipage}{1.0\textwidth}
    \centering
    \begin{subfigure}[t]{0.99\linewidth}
        \centering
        \begin{subfigure}[t]{0.49\linewidth}
            \centering
            \includegraphics[width=1.0\linewidth]{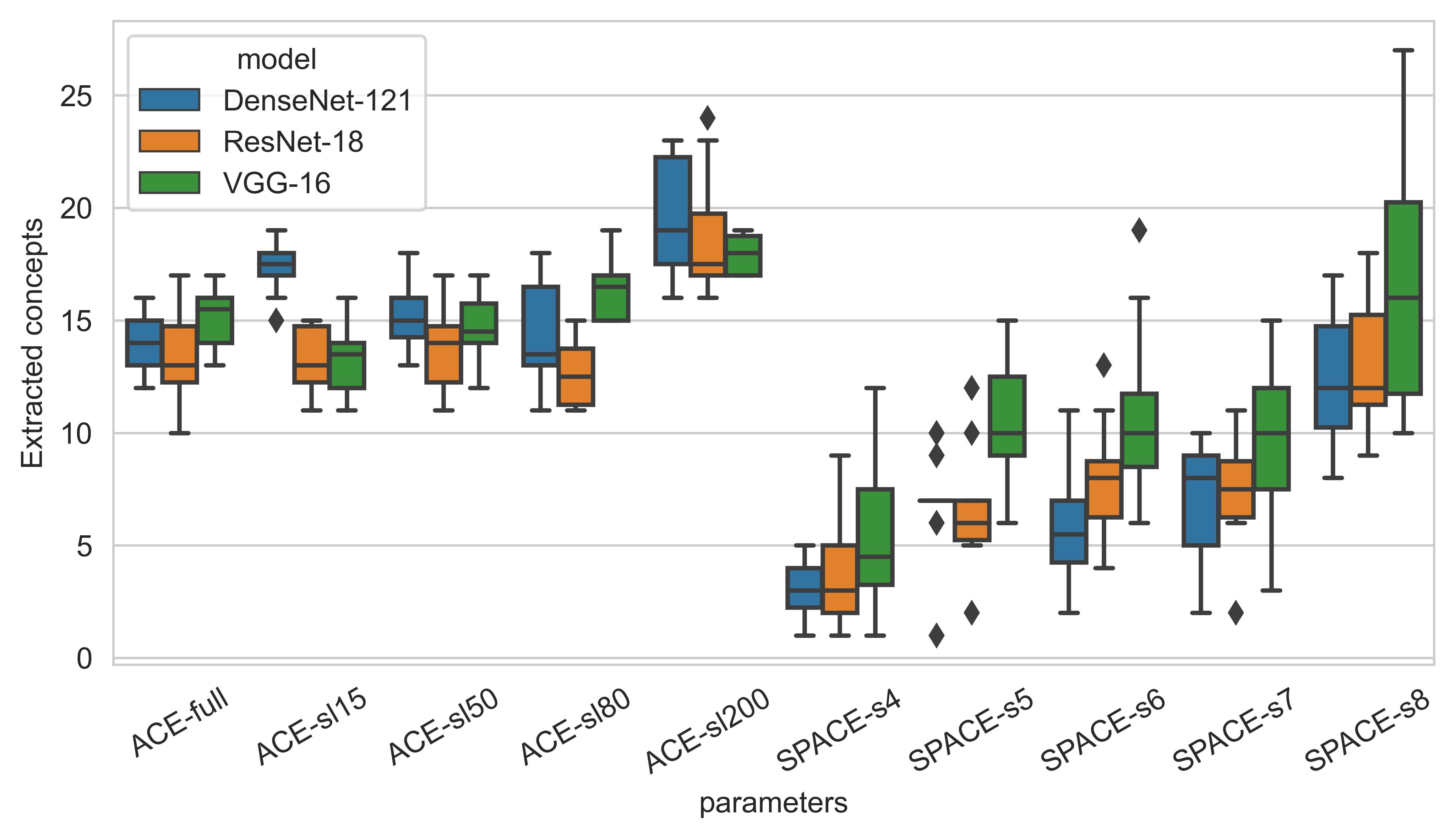}
            \caption{\textbf{Concrete crack}, extracted concepts.}\label{fig:boxplot concrete crack extracted}
        \end{subfigure}
        \begin{subfigure}[t]{0.49\linewidth}
            \centering
            \includegraphics[width=1.0\linewidth]{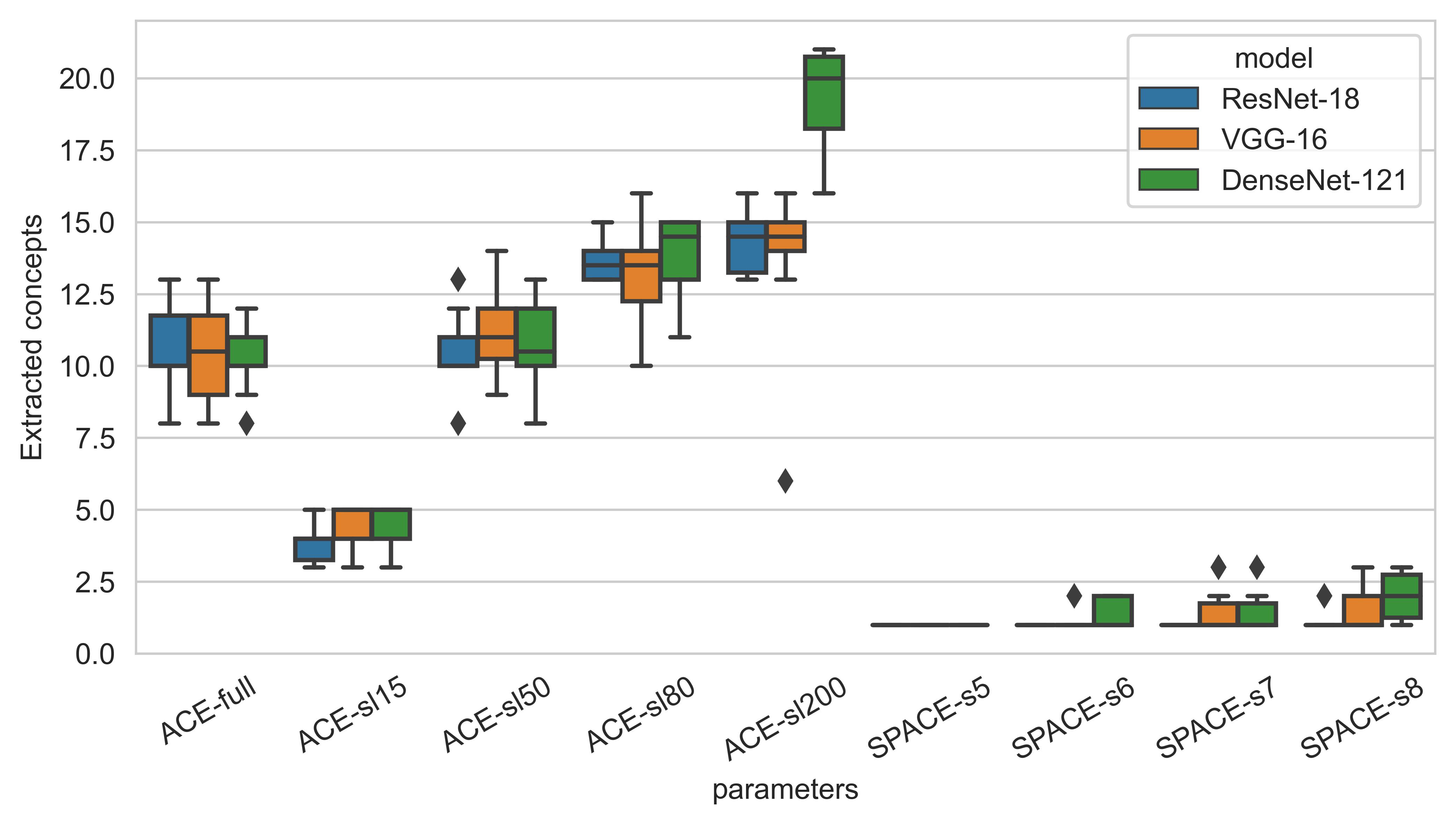}
            \caption{\textbf{Metal nut}, extracted concepts.}\label{fig:boxplot metal nut extracted}
        \end{subfigure}
        
    \end{subfigure}

    \begin{subfigure}[t]{0.99\linewidth}
        \centering
        \begin{subfigure}[t]{0.49\linewidth}
            \centering
            \includegraphics[width=1.0\linewidth]{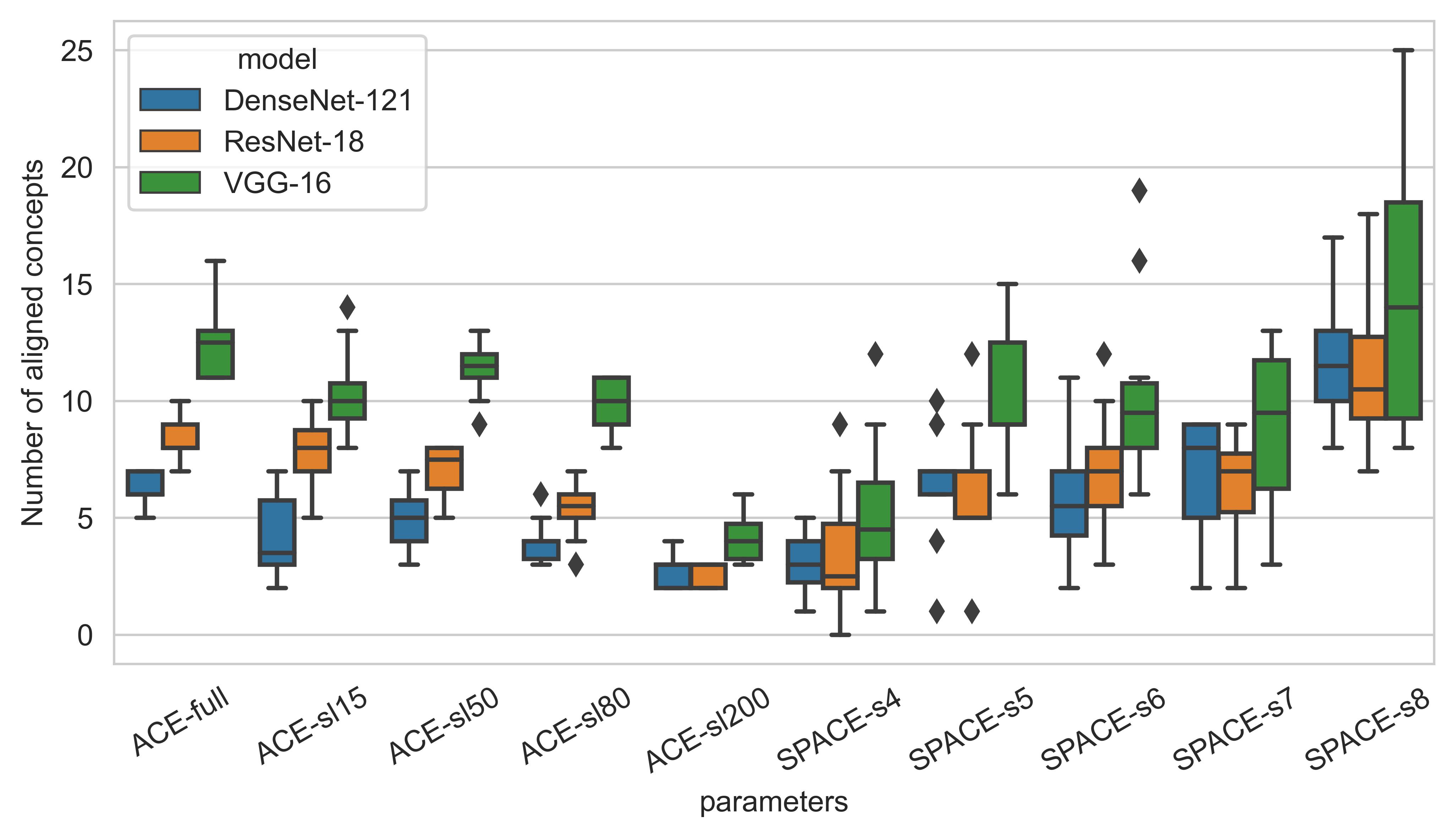}
            \caption{\textbf{Concrete crack}, aligned concepts.}\label{fig:boxplot concrete crack Number of aligned}
        \end{subfigure}
        \begin{subfigure}[t]{0.49\linewidth}
            \centering
            \includegraphics[width=1.0\linewidth]{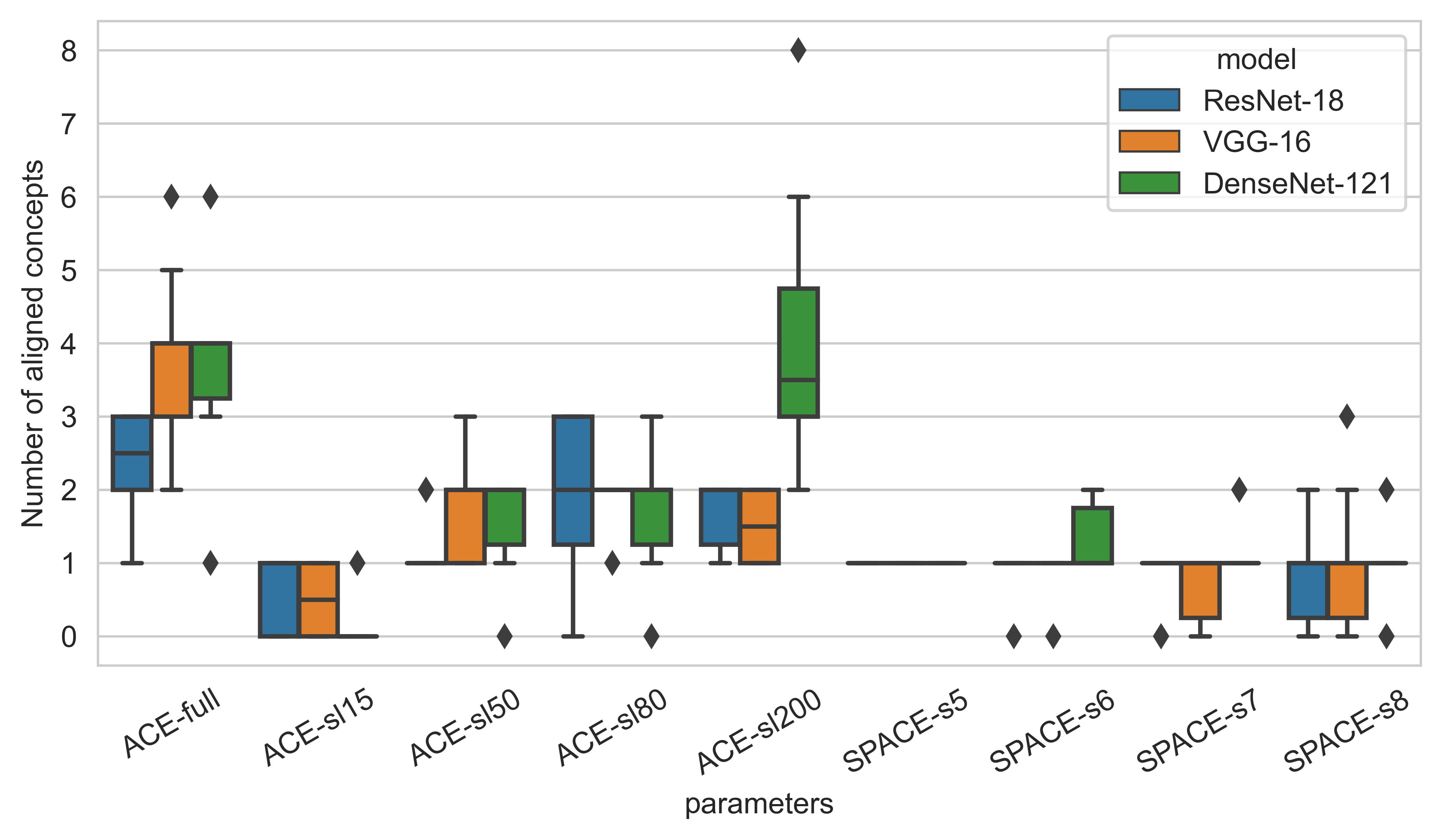}
            \caption{\textbf{Metal nut}, aligned concepts.}\label{fig:boxplot metal nut Number of aligned}
        \end{subfigure}
    \end{subfigure}

    \begin{subfigure}[t]{0.99\linewidth}
        \centering
        \begin{subfigure}[t]{0.49\linewidth}
            \centering
            \includegraphics[width=1.0\linewidth]{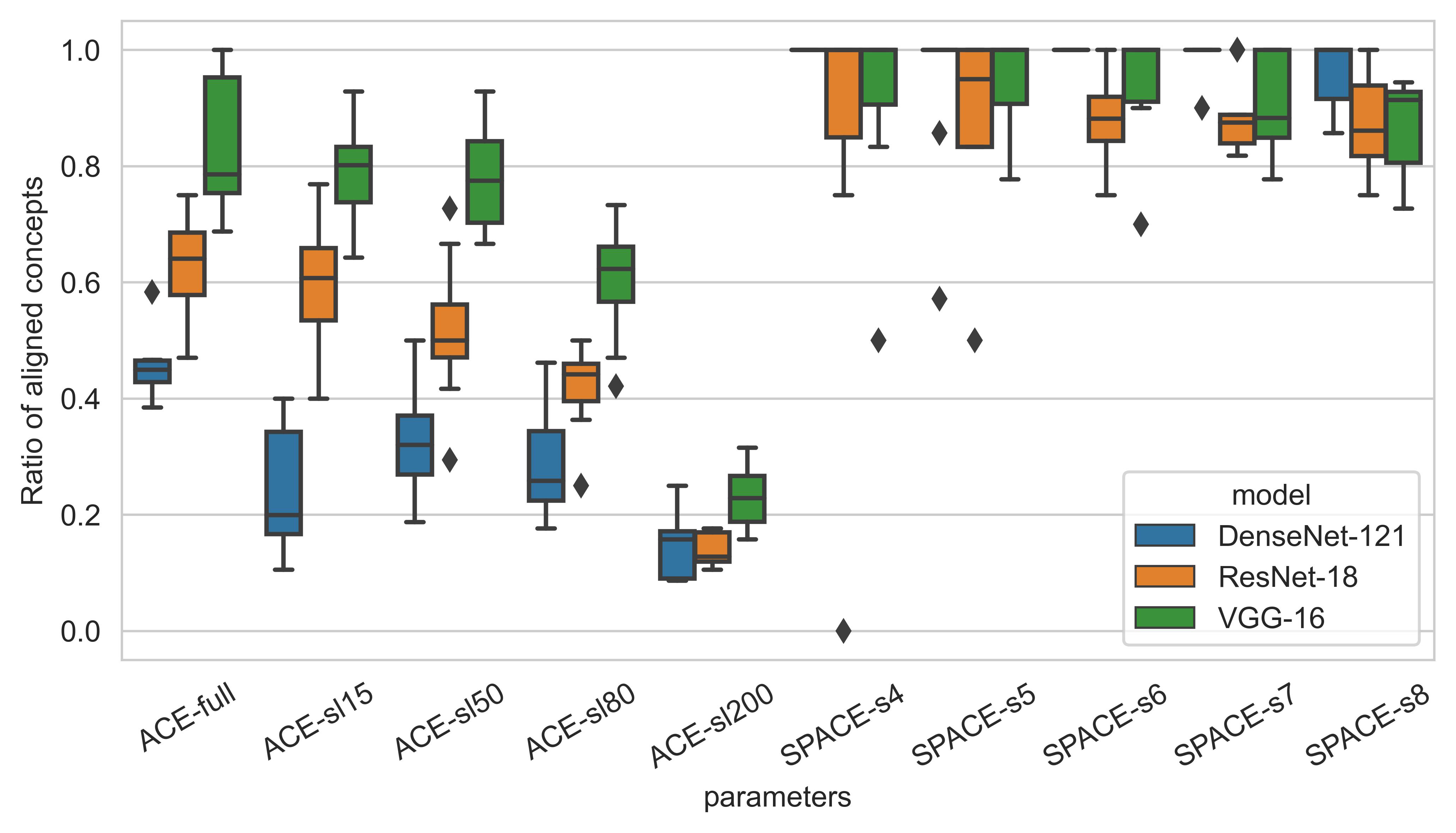}
            \caption{\textbf{Concrete crack}, aligned ratio.}\label{fig:boxplot concrete crack aligned ratio}
        \end{subfigure}
        \begin{subfigure}[t]{0.49\linewidth}
            \centering
            \includegraphics[width=1.0\linewidth]{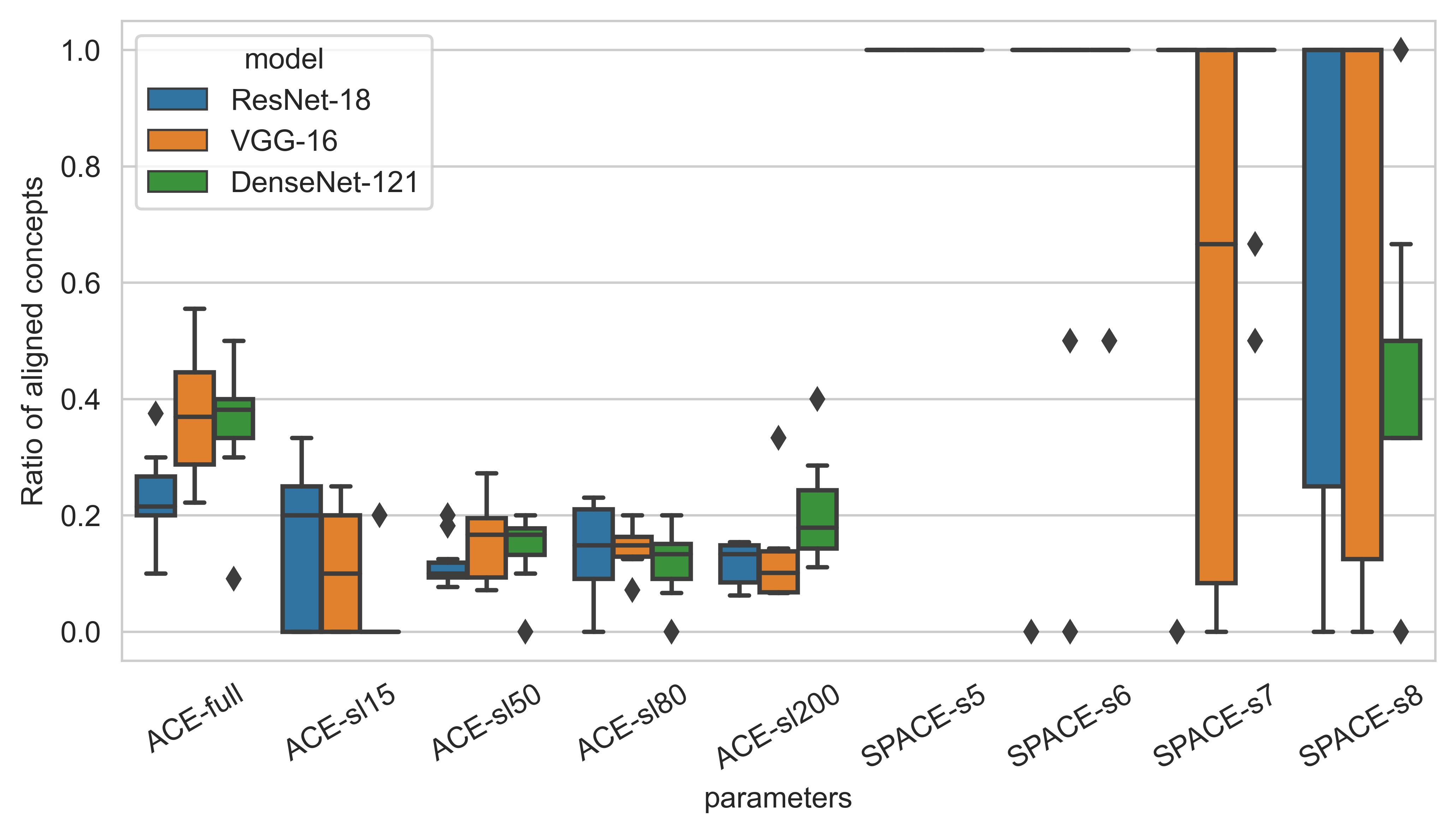}
            \caption{\textbf{Metal nut}, aligned ratio.}\label{fig:boxplot metal nut aligned ratio}
        \end{subfigure}
    \end{subfigure}
    
\end{minipage}
\caption{Aggregated results from SPACE and ACE runs. First we present the number of concepts extracted with more than three items in Fig. \ref{fig:boxplot concrete crack extracted}, and \ref{fig:boxplot metal nut extracted}. Second, the number of aligned concepts extracted in each run are presented in Fig. \ref{fig:boxplot concrete crack Number of aligned}, and \ref{fig:boxplot metal nut Number of aligned}. Third, we present the ratio of aligned concepts, in Fig. \ref{fig:boxplot concrete crack aligned ratio}, and \ref{fig:boxplot metal nut aligned ratio}. In both cases, SPACE extracts more relevant concepts through a variety of hyperparameters, achieving a higher alignment ratio.}
\label{fig:boxplots}
\end{figure}

Our experimental setup focuses on the number of patches used by each method to extract patches, as this is the most sensitive parameter when dealing with features of different scales.
For ACE, we test multiple values for $n_{\mathrm{SLIC}}$, which directly refers to the number of superpixels/patches obtained using the segmentation technique SLIC \citep{6205760}. Thus, in Figure \ref{fig:boxplots} ACE-sl15, ACE-sl50, ACE-sl80, ACE-sl200, describe ACE runs where $n_{\mathrm{SLIC}}$ is equal to 15, 50, 80, and 200 respectively. The run ACE-full represent a single run as described in by \citep{NEURIPS2019_77d2afcb}, where the patches of multiple SLIC executions (to extract 15, 50, and 80 segments) are extracted and merged.
Other ACE parameters were fixed to the values recommended by \citep{6205760}, $s_{\mathrm{SLIC}}=1.0$ (sigma), $c_{\mathrm{SLIC}}=20.0$ (compactness), $n_{\mathrm{k}}=25$ (clusters), and a gray padding of 128. 

For SPACE, we test multiple values for $n_{\mathrm{s}}$, which refers to the number of vertical and horizontal windows $n_s \times n_s$ the images will be divided on, and thus, defines a total of $n_{\mathrm{s}}^2$ patches. In Figure \ref{fig:boxplots}, the runs SPACE-s4 to SPACE-s8, denote SPACE runs with $n_{\mathrm{s}}=4$ to $n_{\mathrm{s}}=8$ respectively. The percentage of selected patches $n_{\mathrm{p}}$ and the dimensionality reduction before clustering $n_{\mathrm{pca}}$, were fixed to 10\% and 30 respectively. Additionally, the last convolutional layer of the model was used for $l_{\mathrm{gradcam}}$, since this layer is expected to have the best compromise between high-level semantics and detailed spatial information \citep{8237336}.

To obtain comparable results, some parameters were set to the same values for the two methods. The layer selected for performing the TCAV analysis as well as the clustering step ($l_{\mathrm{activ}}$), was the closest to the top of the network, namely the dense layer before the softmax layer. It was chosen to capture the concepts which are most important for the final classifications of the model. In contrast with the original implementation of ACE, we performed a PCA with 30 components before the clustering (k-means or OPTICS) in each method. Similarly, during each run, folders of random concepts were automatically created to obtain a compromise between computational cost, and stable and reliable results.

For each run, we visually inspected the example images from each concept, labelling it as aligned or not, based on the intended semantic meaning of the classification task.
In the case of concrete crack detection, the task is visually straightforward; cracks are easy to identify in images or concept example images without major confounding factors that could result in mislabel concept alignments. 
Similarly, for the metal nut dataset, we had access to binary ground truth masks, providing clear indicators of key visual features.
The visual cues related to defects are unambiguous, making the visual inspection of concept examples feasible without significant exposure to confounding factors. More so, in the selected datasets, there is no major difference in prior knowledge between experts and non-experts (e.g. the visual cues related to a crack are easy to identify).

We aggregated the results of the runs for the ten random seeds on each parameter configuration. 
The aggregation was performed through visual inspection on the results of each run, labeling concepts as aligned or not, based on the intended semantic meaning of the analyzed class.
Then, we aggregate the results, by computing the average ratio of aligned concepts across all runs of a model and method configuration.
We use this aggregation method in response to the question, ``are the right concepts being extracted for the right classes?''.
The summarized results are presented in Figure \ref{fig:boxplots}, and detailed in Appendix \ref{apd:Extended concrete cracks}, and\ref{apd:Extended metal nut}.

The main points that can be highlighted from the empirical results concern the amount and type of extracted concepts. First, \textbf{SPACE extracts fewer concepts, yet, these concepts are more aligned with the semantic meaning of the analyzed classes}. An example can be observed in Figure \ref{fig:boxplot metal nut extracted} and \ref{fig:boxplot metal nut aligned ratio}, where the runs SPACE-s5 and SPACE-s6, extracted a single yet aligned concept, in comparison with the best run of ACE (ACE-full), which extracted on average ten valid concepts, yet only half were meaningful ones. A similar case can be observed for the concrete crack dataset in Figures \ref{fig:boxplot concrete crack extracted}, and \ref{fig:boxplot concrete crack aligned ratio}. 
Second, \textbf{when extracting concepts SPACE is able to focus on the meaningful parts of the images, reducing the effect of outliers and irrelevant regions of the images}. This is the result of the patch extraction, concept clustering steps, which make SPACE able to differentiate outliers, detect non spherical clusters, and focus on the most important patches obtained from images. In contrast, outliers and irrelevant background can influence the k-means process of ACE, diminishing the number of valid concepts (with more than 3 samples), as well as the ratio of aligned concepts (as seen in Figure \ref{fig:boxplots}).
Third, \textbf{the nature of the extracted patches differs for both methods, which impacts the features that are extracted, their encoding and testing}. This leads to better performance of both methods for bigger features, and a better performance of SPACE for smaller features. This second point will be better visualized in the example cases below.

To better visualize our findings, we further discuss two example runs of SPACE and three runs of ACE for each dataset. The parameters of each method were modified to better observe their mechanisms at different scales. In addition, we also show the results of both methods for a real-world use case on quality control on metal casting\footnote{In collaboration with Deevio GmbH.}. Afterward, the results with the highest importance scores are presented and compared.

The example runs of SPACE for the three datasets are described in Table \ref{table:runs}. The number of slices $n_{\mathrm{s}}$ was varied according to each dataset and the scale of its features. The percentage of filtered patches $n_{\mathrm{p}}$ and the number of components extracted through PCA, $n_{\mathrm{pca}}$ were fixed as before. Similarly the Table \ref{table:runs} also summarizes all the example runs of ACE, as well as the sets of parameters used for each run. Where $s_{\mathrm{SLIC}}$, $n_{\mathrm{SLIC}}$, and $c_{\mathrm{SLIC}}$ refer to sigma, number of patches and compactness of the SLIC segmentation. For the ACE runs, the padding of the patches was fixed to 128 as before. In addition, the number of clusters extracted through k-means $n_{\mathrm{k}}$, was varied to account for the lower number of valid concepts usually extracted when using a fewer number of patches (as seen in Figure \ref{fig:boxplot concrete crack extracted}).

\begin{table}[htb]
\caption{Specification of the experiments with SPACE and ACE.} \label{table:runs}
\begin{tabular}{|l|l|l|l|}

\hline

\rowcolor[HTML]{C0C0C0} 

run     & dataset        & class     & parameters                                                                              \\ \hline

SPACE-A & Concrete crack & cracked   & $n_{\mathrm{s}}=3, n_{\mathrm{p}}=10\%,  n_{\mathrm{pca}}=30$                           \\ \hline

SPACE-B & Concrete crack & cracked   & $n_{\mathrm{s}}=6, n_{\mathrm{p}}=10\%,    n_{\mathrm{pca}}=30$                         \\ \hline

SPACE-C & Metal nut      & color     & $n_{\mathrm{s}}=7, n_{\mathrm{p}}=10\%,    n_{\mathrm{pca}}=30$                         \\ \hline

SPACE-D & Metal nut      & color     & $n_{\mathrm{s}}=14, n_{\mathrm{p}}=10\%,    n_{\mathrm{pca}}=30$                        \\ \hline

SPACE-E & Metal casting  & defective & $n_{\mathrm{s}}=9, n_{\mathrm{p}}=10\%,    n_{\mathrm{pca}}=30$                         \\ \hline

SPACE-F & Metal casting  & defective & $n_{\mathrm{s}}=18, n_{\mathrm{p}}=10\%,    n_{\mathrm{pca}}=30$                        \\ \hline

ACE-A   & Concrete crack & cracked   & $s_{\mathrm{SLIC}}=3, n_{\mathrm{SLIC}}=15,  c_{\mathrm{SLIC}}=60, n_{\mathrm{k}}=10$   \\ \hline

ACE-B   & Concrete crack & cracked   & $s_{\mathrm{SLIC}}=3, n_{\mathrm{SLIC}}=50,  c_{\mathrm{SLIC}}=60, n_{\mathrm{k}}=25$   \\ \hline

ACE-C   & Concrete crack & cracked   & $s_{\mathrm{SLIC}}=3, n_{\mathrm{SLIC}}=200,  c_{\mathrm{SLIC}}=300, n_{\mathrm{k}}=25$ \\ \hline

ACE-D   & Metal nut      & color     & $s_{\mathrm{SLIC}}=3, n_{\mathrm{SLIC}}=15,  c_{\mathrm{SLIC}}=60, n_{\mathrm{k}}=10$   \\ \hline

ACE-E   & Metal nut      & color     & $s_{\mathrm{SLIC}}=3, n_{\mathrm{SLIC}}=50,  c_{\mathrm{SLIC}}=60, n_{\mathrm{k}}=25$   \\ \hline

ACE-F   & Metal nut      & color     & $s_{\mathrm{SLIC}}=3, n_{\mathrm{SLIC}}=200,  c_{\mathrm{SLIC}}=300, n_{\mathrm{k}}=25$ \\ \hline

ACE-G   & Metal casting  & defective & $s_{\mathrm{SLIC}}=3, n_{\mathrm{SLIC}}=15,  c_{\mathrm{SLIC}}=60, n_{\mathrm{k}}=10$   \\ \hline

ACE-H   & Metal casting  & defective & $s_{\mathrm{SLIC}}=3, n_{\mathrm{SLIC}}=50,  c_{\mathrm{SLIC}}=60, n_{\mathrm{k}}=25$   \\ \hline

ACE-I   & Metal casting  & defective & $s_{\mathrm{SLIC}}=3, n_{\mathrm{SLIC}}=200,  c_{\mathrm{SLIC}}=300, n_{\mathrm{k}}=25$ \\ \hline

\end{tabular}
\end{table}

\subsection{Concrete crack dataset}
\label{sec:results-crack}

\begin{figure}[h]
\centering
\includegraphics[width=0.93\linewidth]{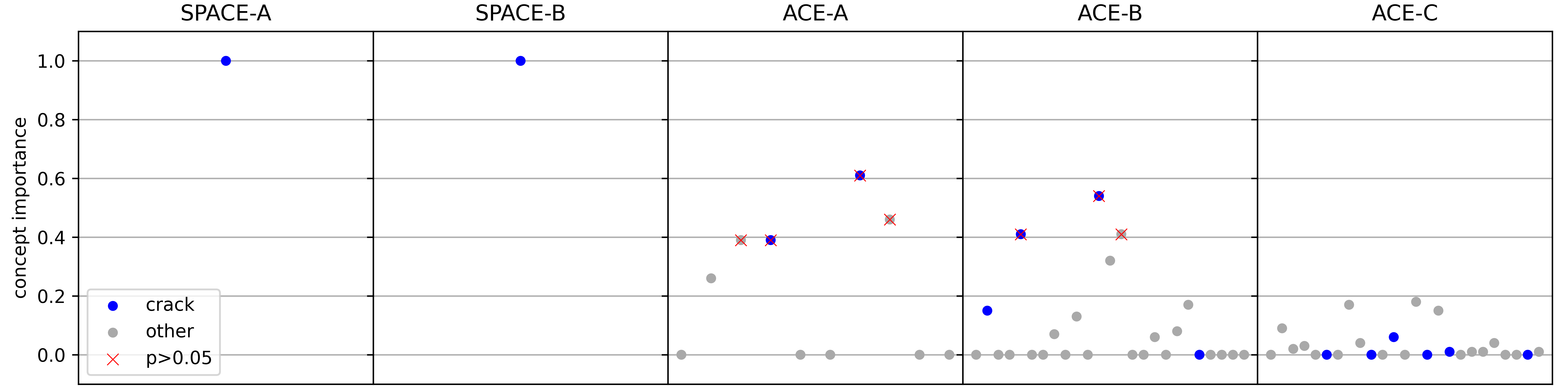}
\caption{Importance scores of concepts for the concrete crack dataset. Markers indicate the interpreted content of the concept, and its statistical significance. SPACE extracted meaningful concepts and scored them consistently. In contrast, ACE runs extracted mixed concepts, scoring them inconsistently.}
\label{fig:concrete crack importance scores}
\end{figure}

\begin{figure}[h]
\centering
\begin{minipage}{1.0\textwidth}
    \centering
    \begin{subfigure}[t]{0.98\linewidth}
        \centering
        \begin{subfigure}[t]{0.48\linewidth}
            \centering
            \includegraphics[width=0.24\linewidth]{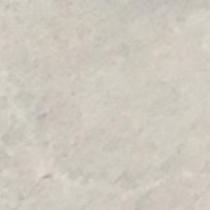}
            \includegraphics[width=0.24\linewidth]{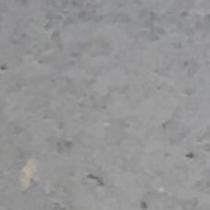}
            \caption{ok}\label{fig:crack-class0}
        \end{subfigure}
        \begin{subfigure}[t]{0.48\linewidth}
            \centering
            \includegraphics[width=0.24\linewidth]{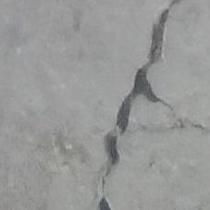}
            \includegraphics[width=0.24\linewidth]{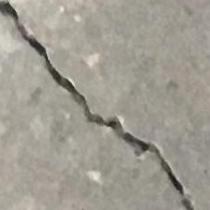}
            \caption{cracked}\label{fig:crack-class1}
        \end{subfigure}
    \end{subfigure}
    
    \begin{subfigure}[t]{0.32\linewidth}
        \centering
        \includegraphics[width=0.24\linewidth]{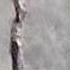}
        \includegraphics[width=0.24\linewidth]{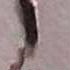}
        \includegraphics[width=0.24\linewidth]{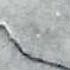}
        \caption{SPACE-A: $c_{0}$ $\overline{S}_0=1.0$}\label{fig:SPACE-A-0}
    \end{subfigure}
    \begin{subfigure}[t]{0.32\linewidth}
        \centering
        \includegraphics[width=0.24\linewidth]{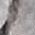}
        \includegraphics[width=0.24\linewidth]{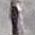}
        \includegraphics[width=0.24\linewidth]{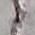}
        \caption{SPACE-B: $c_{0}$ $\overline{S}_0=1.0$}\label{fig:SPACE-B-0}
    \end{subfigure}
    \begin{subfigure}[t]{0.32\linewidth}
        \centering
        \includegraphics[width=0.24\linewidth]{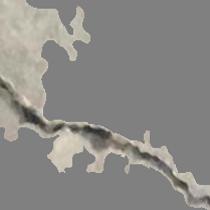}
        \includegraphics[width=0.24\linewidth]{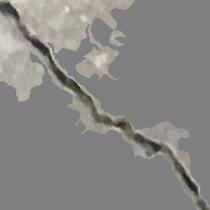}
        \includegraphics[width=0.24\linewidth]{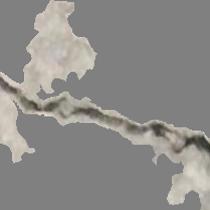}
        \caption{ACE-A: $c_{6}$ $\overline{S}_6=0.61$}\label{fig:ACE-A-6}
    \end{subfigure}
    \begin{subfigure}[t]{0.32\linewidth}
        \centering
        \includegraphics[width=0.24\linewidth]{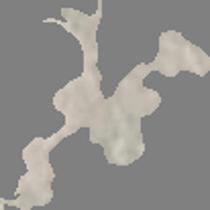}
        \includegraphics[width=0.24\linewidth]{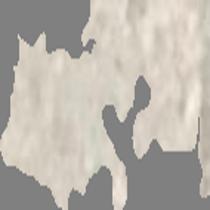}
        \includegraphics[width=0.24\linewidth]{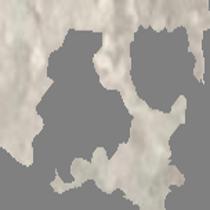}
        \caption{ACE-A: $c_{7}$ $\overline{S}_7=0.46$}\label{fig:ACE-A-7}
    \end{subfigure}
    \begin{subfigure}[t]{0.32\linewidth}
        \centering
        \includegraphics[width=0.24\linewidth]{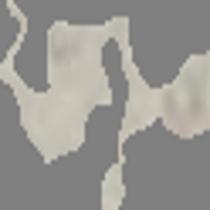}
        \includegraphics[width=0.24\linewidth]{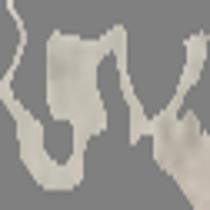}
        \includegraphics[width=0.24\linewidth]{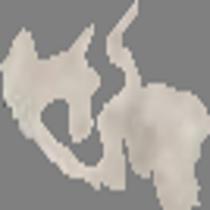}
        \caption{ACE-B: $c_{13}$ $\overline{S}_{13}=0.46$}\label{fig:ACE-B-13}
    \end{subfigure}
    \begin{subfigure}[t]{0.32\linewidth}
        \centering
        \includegraphics[width=0.24\linewidth]{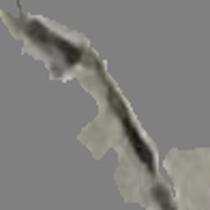}
        \includegraphics[width=0.24\linewidth]{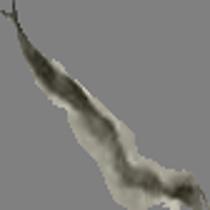}
        \includegraphics[width=0.24\linewidth]{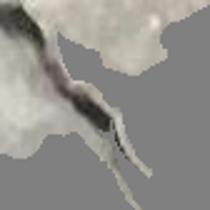}
        \caption{ACE-B: $c_{11}$ $\overline{S}_{11}=0.45$}\label{fig:ACE-B-11}
    \end{subfigure}
    \begin{subfigure}[t]{0.32\linewidth}
        \centering
        \includegraphics[width=0.24\linewidth]{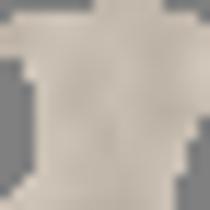}
        \includegraphics[width=0.24\linewidth]{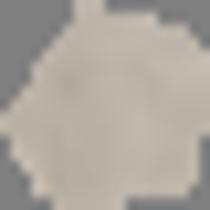}
        \includegraphics[width=0.24\linewidth]{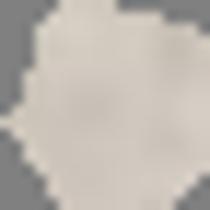}
        \caption{ACE-C: $c_{13}$ $\overline{S}_{13}=0.18$}\label{fig:ACE-C-13}
    \end{subfigure}
    \begin{subfigure}[t]{0.32\linewidth}
        \centering
        \includegraphics[width=0.24\linewidth]{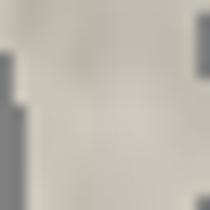}
        \includegraphics[width=0.24\linewidth]{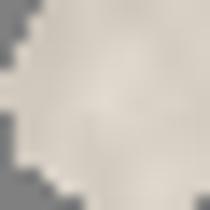}
        \includegraphics[width=0.24\linewidth]{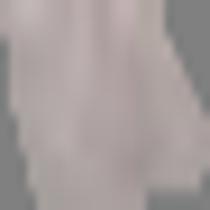}
        \caption{ACE-C: $c_{7}$ $\overline{S}_7=0.17$}\label{fig:ACE-C-7}
    \end{subfigure}
\end{minipage}
\caption{Concrete crack dataset results for class ``cracked'' (k=1). Examples of classes in (a) and (b). Top concepts from two SPACE runs and three ACE runs in (c) to (j). Top concepts with high importance from SPACE-A and SPACE-B contain cracks, which are the most relevant feature of the class. Top concepts from ACE-A and ACE-B contain segmented cracks but are of lower importance (e), (h). In contrast, top concepts from ACE-C do not contain cracks at all (i), (j).}
\label{fig:concrete crack}
\end{figure}

The concrete crack dataset \citep{conc} consists of two classes, each with \num{20000} concrete images of 227x227 pixels resized to 210x210. Class 0 and 1 correspond to good and cracked samples of concrete, respectively (Figure \ref{fig:crack-class0} and \ref{fig:crack-class1}). The main (and only) concept used for the labeling of the classes is well known, as the cracks span through complete images and there are no other significant features present in the dataset. After training, the VGG-16 model obtained a test accuracy of 99.9\%. The importance of the extracted concepts is shown in Figure \ref{fig:concrete crack importance scores}, and examples of the highest importance score concepts are shown in Figure \ref{fig:concrete crack}.

%

\textbf{SPACE} extracted a single concept per run, shown in Figures \ref{fig:SPACE-A-0} and \ref{fig:SPACE-B-0}. In both runs, the extracted concepts were statistically significant with the highest possible importance score (1.0). The background-related patches were filtered due to a low aggregated importance. The number of concepts extracted in each \textbf{ACE} run depends on $n_{\mathrm{SLIC}}$. The most notable finding for the ACE runs is the significantly lower importance score of the concepts. Over 48\% of concepts had an importance score of 0.0, shown in Figure \ref{fig:concrete crack importance scores}. Similarly, the highest importance scores for the runs were lower than 0.61. 


\textbf{Patches and importance scores}. The concepts extracted with SPACE were square patches directly containing cracks and scored with high importance. In contrast, ACE extracted 80\% of concepts containing normal concrete patches and 20\% of concepts containing cracks. These concepts were scored significantly lower, with 70\% scored with less than 0.1. Even the concepts with the highest importance were scored a maximum of 0.61, as can be seen in Figure \ref{fig:ACE-A-6}.

\textbf{Alignment}. The discriminative feature of the datasets are the cracks, which is aligned with the results obtained through the two SPACE runs, as seen in Figures \ref{fig:SPACE-A-0} and \ref{fig:SPACE-B-0}. In comparison, ACE concepts containing cracks were not always scored with high importance. As seen in Figure \ref{fig:concrete crack importance scores}, most concepts containing cracks were scored close to 0.0, meaning that they adversely affected the prediction of the cracked class. This points towards issues when using TCAV in the ACE runs, which is caused by how the patches were segmented, the padding of the patches, and the interpolation used for resizing.

\subsection{MVTec Metal Nuts dataset}
\label{sec:results-metalnut}

\begin{figure}[h]
\centering
\includegraphics[width=0.93\linewidth]{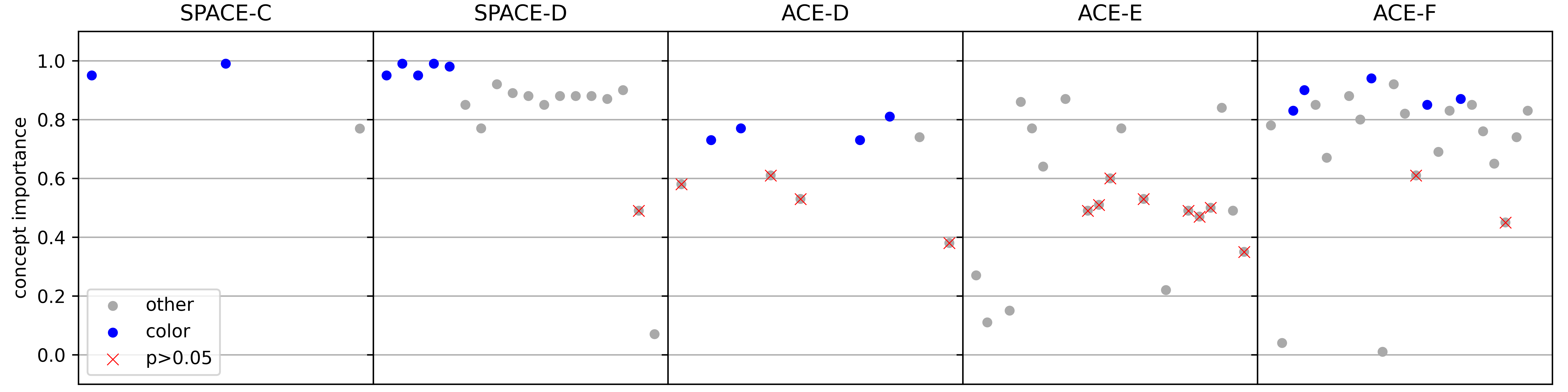}
\caption{Importance scores of concepts for the metal nut dataset. Markers indicate the interpreted content of the concept, and its statistical significance. SPACE consistently extracted and scored meaningful concepts. In contrast, the extraction and scoring of meaningful concepts differed in the ACE runs.}
\label{fig:metal nut importance scores}
\end{figure}

\begin{figure}[h]

\centering
\begin{minipage}{1.0\textwidth}
    \centering
    \begin{subfigure}[t]{0.98\linewidth}
        \centering
        \begin{subfigure}[t]{0.18\linewidth}
            \centering
            \includegraphics[width=1.0\linewidth]{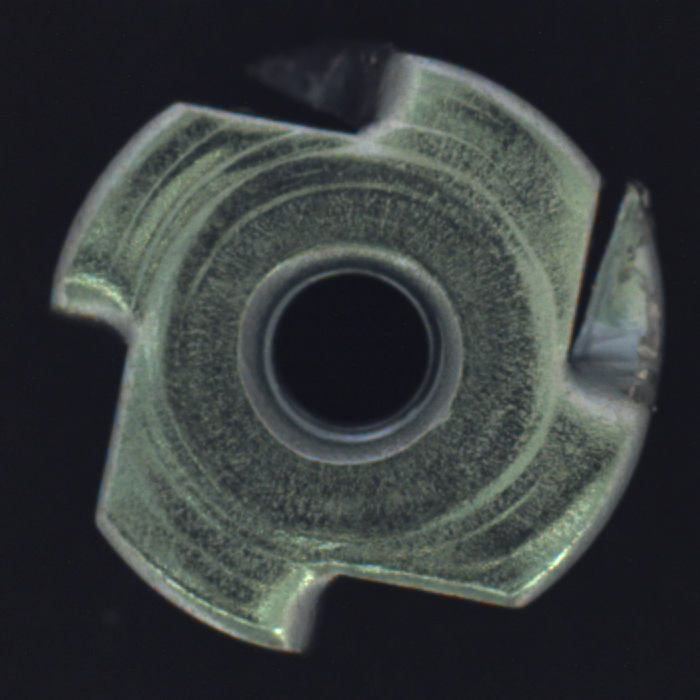}
            \caption{bent}\label{fig:metalnut-class0}
        \end{subfigure}
        \begin{subfigure}[t]{0.18\linewidth}
            \centering
            \includegraphics[width=1.0\linewidth]{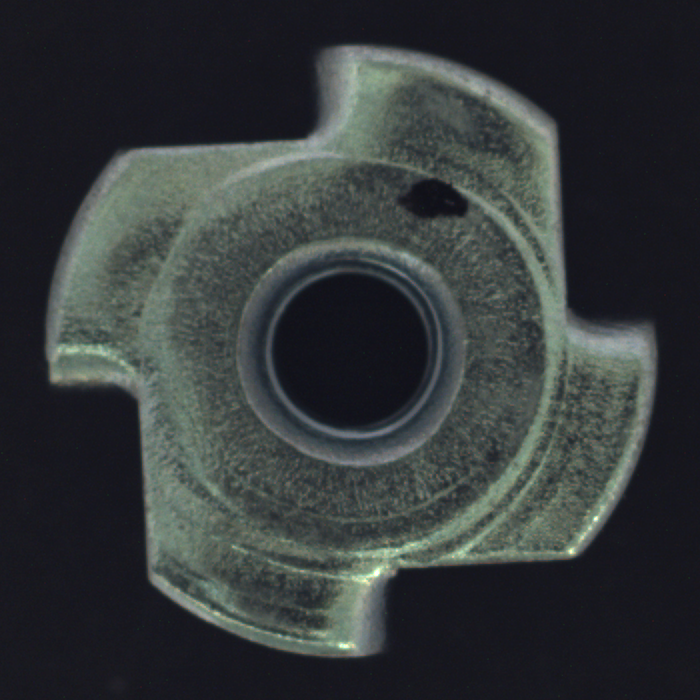}
            \caption{color}\label{fig:metalnut-class1}
        \end{subfigure}
        \begin{subfigure}[t]{0.18\linewidth}
            \centering
            \includegraphics[width=1.0\linewidth]{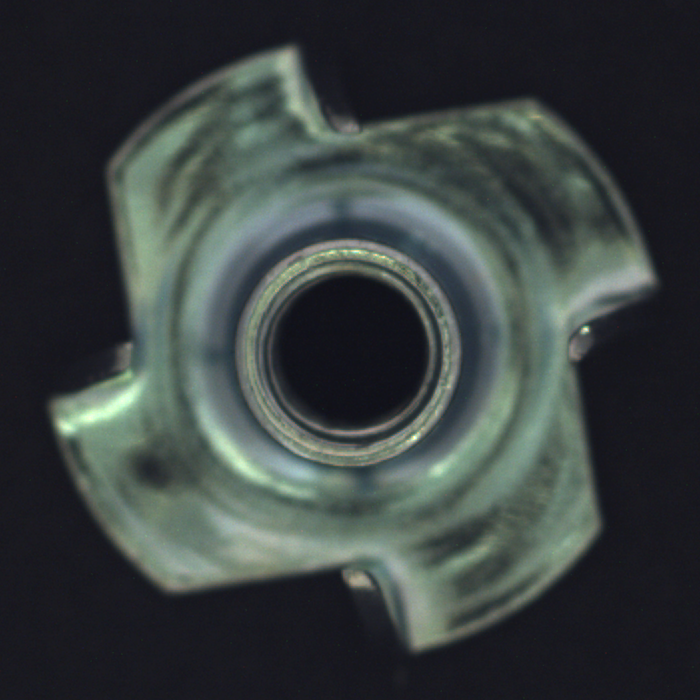}
            \caption{flipped}\label{fig:metalnut-class2}
        \end{subfigure}
        \begin{subfigure}[t]{0.18\linewidth}
            \centering
            \includegraphics[width=1.0\linewidth]{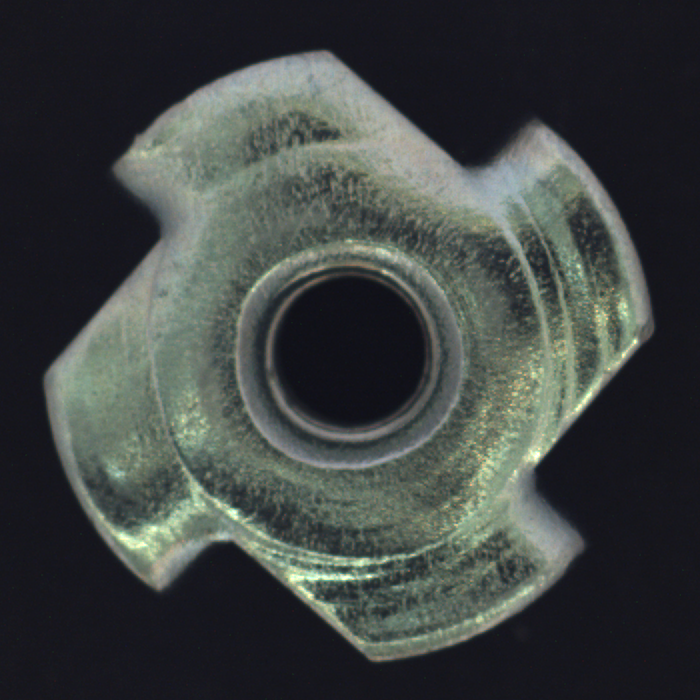}
            \caption{ok}\label{fig:metalnut-class3}
        \end{subfigure}
        \begin{subfigure}[t]{0.18\linewidth}
            \centering
            \includegraphics[width=1.0\linewidth]{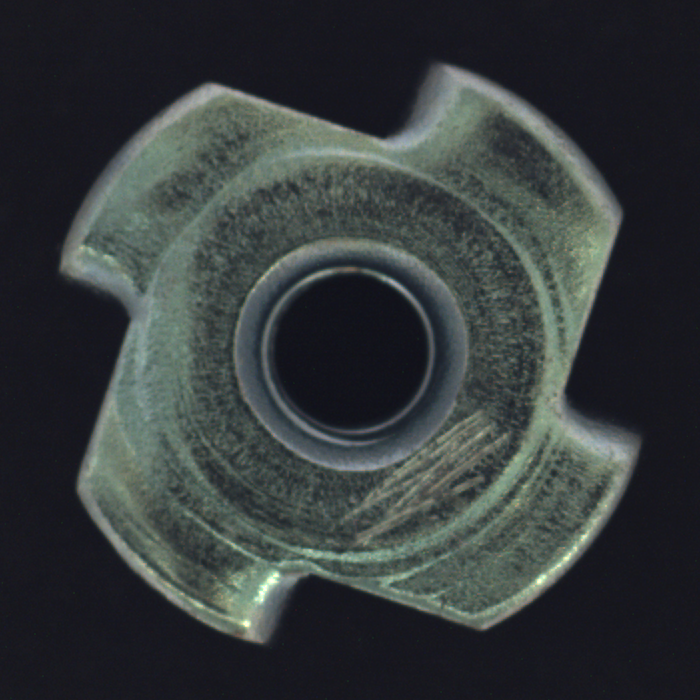}
            \caption{scratch}\label{fig:metalnut-class4}
        \end{subfigure}
    \end{subfigure}
    
    \begin{subfigure}[t]{0.32\linewidth}
        \centering
        \includegraphics[width=0.24\linewidth]{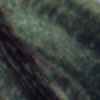}
        \includegraphics[width=0.24\linewidth]{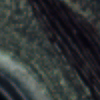}
        \includegraphics[width=0.24\linewidth]{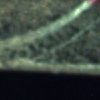}
        \caption{SPACE-C: $c_{1}$ $\overline{S}_1=0.99$}\label{fig:SPACE-C-1}
    \end{subfigure}
    \begin{subfigure}[t]{0.32\linewidth}
        \centering
        \includegraphics[width=0.24\linewidth]{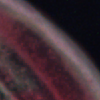}
        \includegraphics[width=0.24\linewidth]{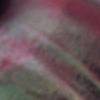}
        \includegraphics[width=0.24\linewidth]{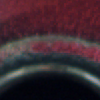}
        \caption{SPACE-C: $c_{0}$ $\overline{S}_0=0.95$}\label{fig:SPACE-C-0}
    \end{subfigure}
    \begin{subfigure}[t]{0.32\linewidth}
        \centering
        \includegraphics[width=0.24\linewidth]{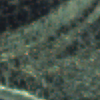}
        \includegraphics[width=0.24\linewidth]{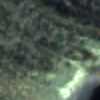}
        \includegraphics[width=0.24\linewidth]{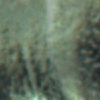}
        \caption{SPACE-C: $c_{2}$ $\overline{S}_2=0.77$}\label{fig:SPACE-C-2}
    \end{subfigure}
    
    \begin{subfigure}[t]{0.32\linewidth}
        \centering
        \includegraphics[width=0.24\linewidth]{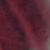}
        \includegraphics[width=0.24\linewidth]{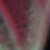}
        \includegraphics[width=0.24\linewidth]{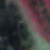}
        \caption{SPACE-D: $c_{1}$ $\overline{S}_1=0.99$}\label{fig:SPACE-D-1}
    \end{subfigure}
    \begin{subfigure}[t]{0.32\linewidth}
        \centering
        \includegraphics[width=0.24\linewidth]{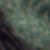}
        \includegraphics[width=0.24\linewidth]{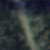}
        \includegraphics[width=0.24\linewidth]{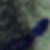}
        \caption{SPACE-D: $c_{3}$ $\overline{S}_3=0.99$}\label{fig:SPACE-D-3}
    \end{subfigure}
    \begin{subfigure}[t]{0.32\linewidth}
        \centering
        \includegraphics[width=0.24\linewidth]{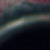}
        \includegraphics[width=0.24\linewidth]{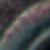}
        \includegraphics[width=0.24\linewidth]{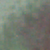}
        \caption{SPACE-D: $c_{4}$ $\overline{S}_4=0.98$}\label{fig:SPACE-D-4}
    \end{subfigure}
    
    \begin{subfigure}[t]{0.32\linewidth}
        \centering
        \includegraphics[width=0.24\linewidth]{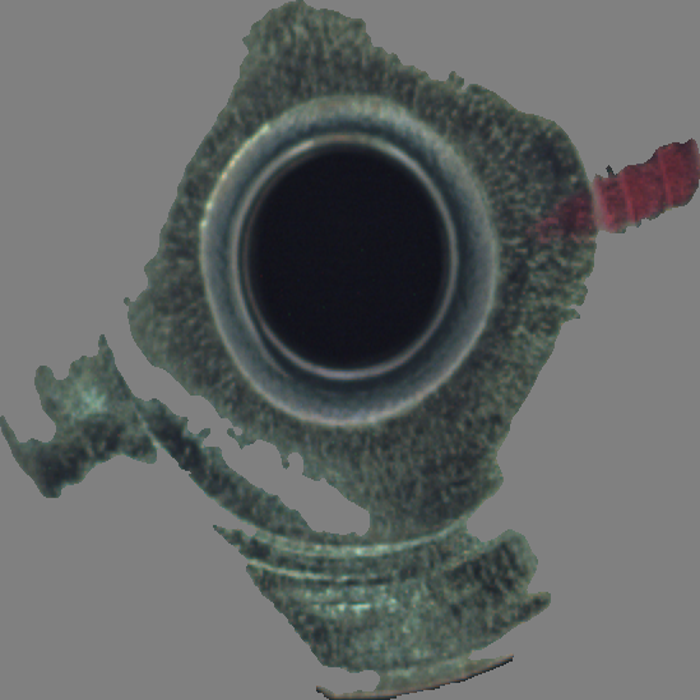}
        \includegraphics[width=0.24\linewidth]{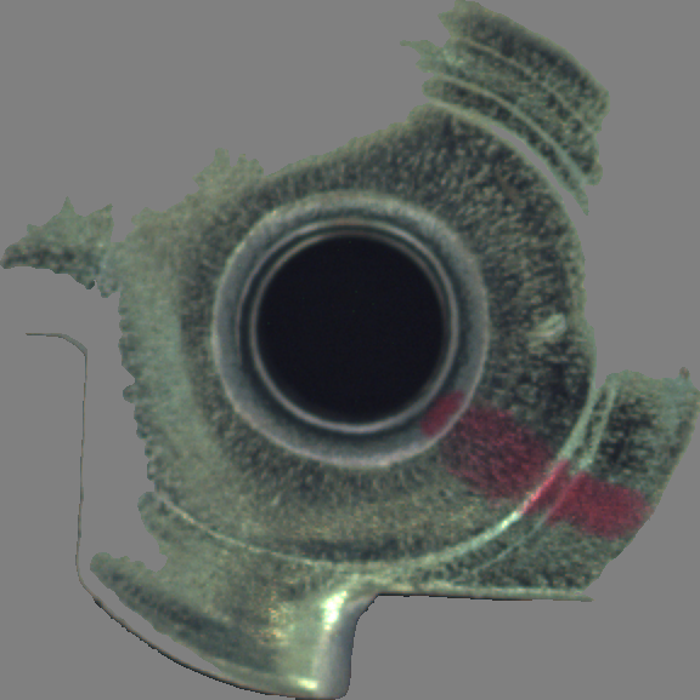}
        \caption{ACE-D: $c_{7}$ $\overline{S}_7=0.81$}\label{fig:ACE-D-7}
    \end{subfigure}
    \begin{subfigure}[t]{0.32\linewidth}
        \centering
        \includegraphics[width=0.24\linewidth]{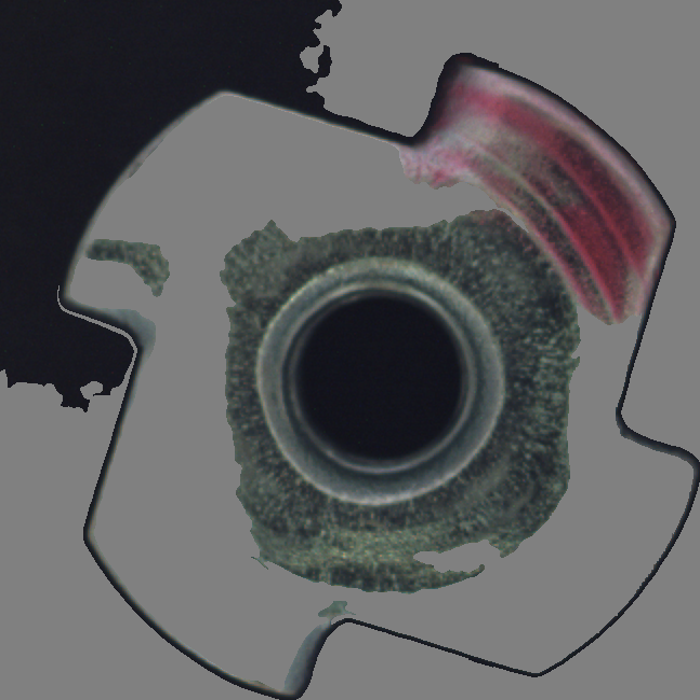}
        \includegraphics[width=0.24\linewidth]{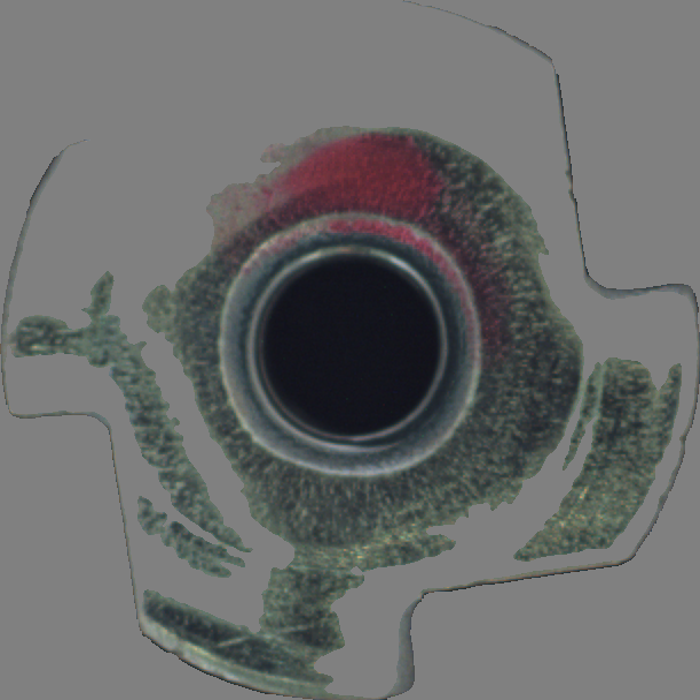}
        \caption{ACE-D: $c_{2}$ $\overline{S}_2=0.77$}\label{fig:ACE-D-2}
    \end{subfigure}
    \begin{subfigure}[t]{0.32\linewidth}
        \centering
        \includegraphics[width=0.24\linewidth]{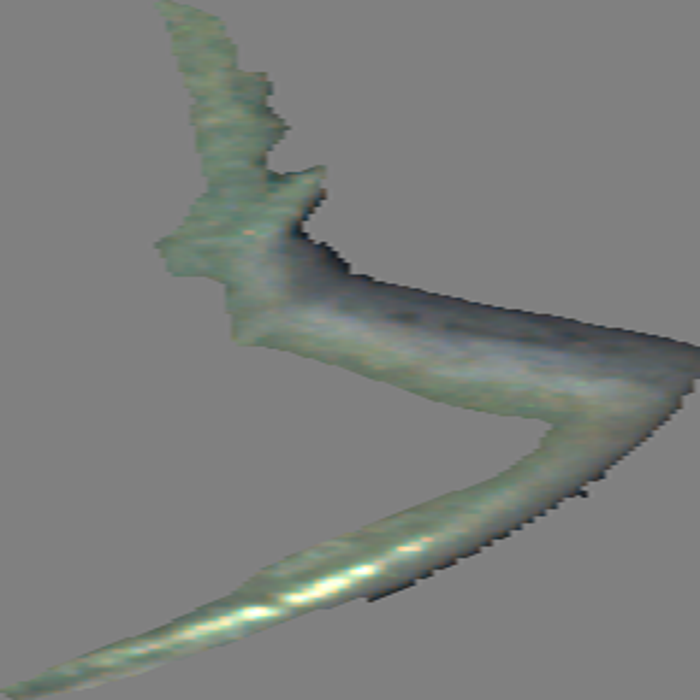}
        \includegraphics[width=0.24\linewidth]{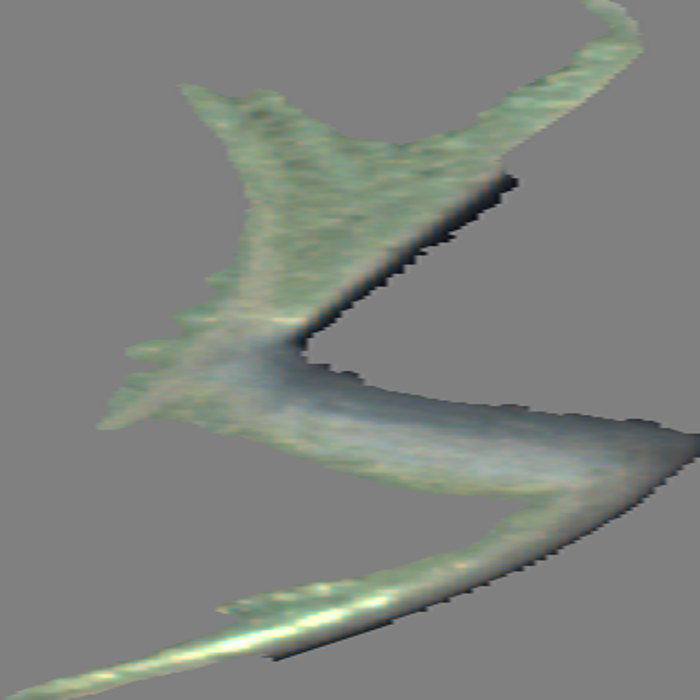}
        \includegraphics[width=0.24\linewidth]{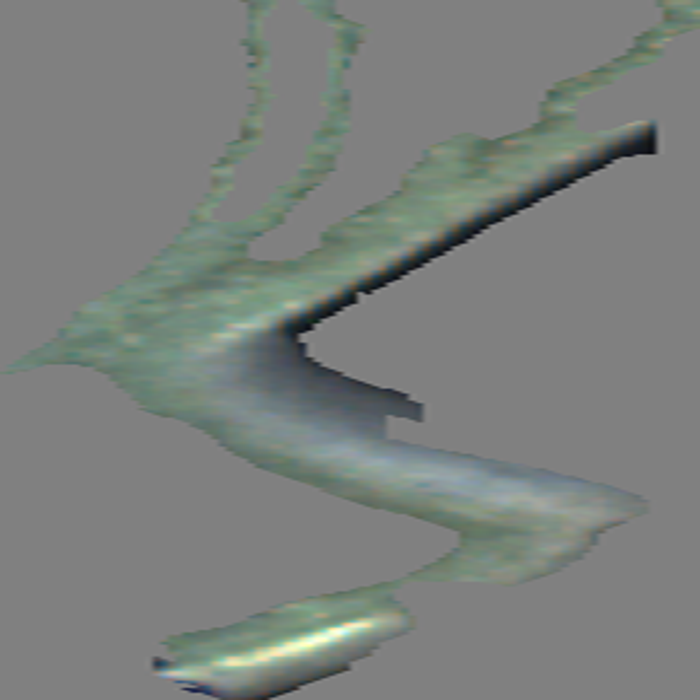}
        \caption{ACE-E: $c_{8}$ $\overline{S}_8=0.87$}\label{fig:ACE-E-8}
    \end{subfigure}
    \begin{subfigure}[t]{0.32\linewidth}
        \centering
        \includegraphics[width=0.24\linewidth]{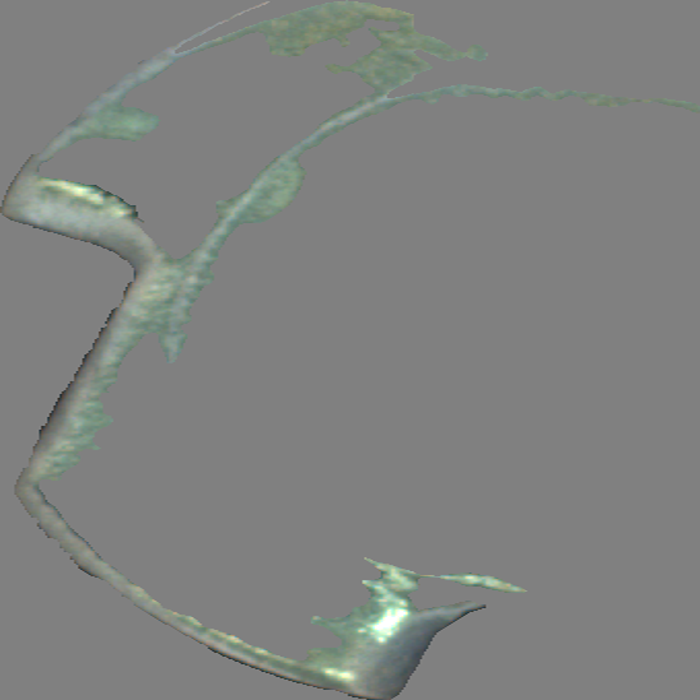}
        \includegraphics[width=0.24\linewidth]{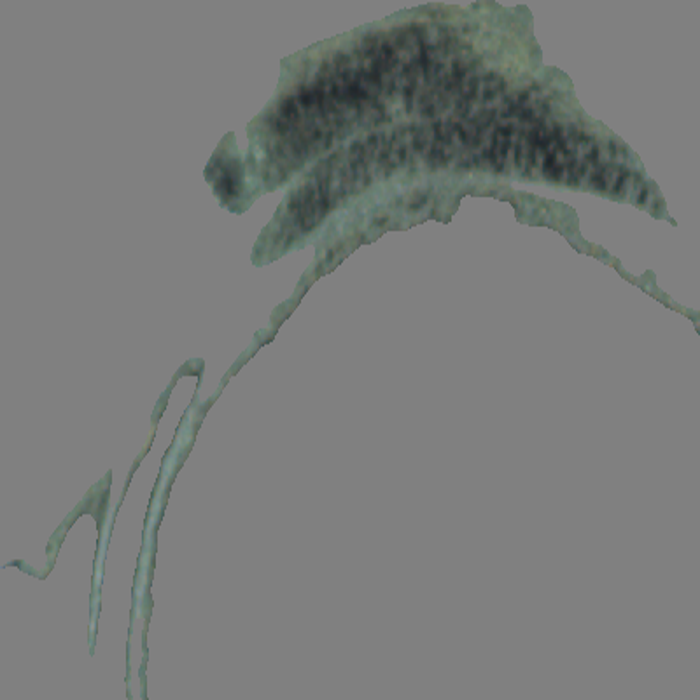}
        \includegraphics[width=0.24\linewidth]{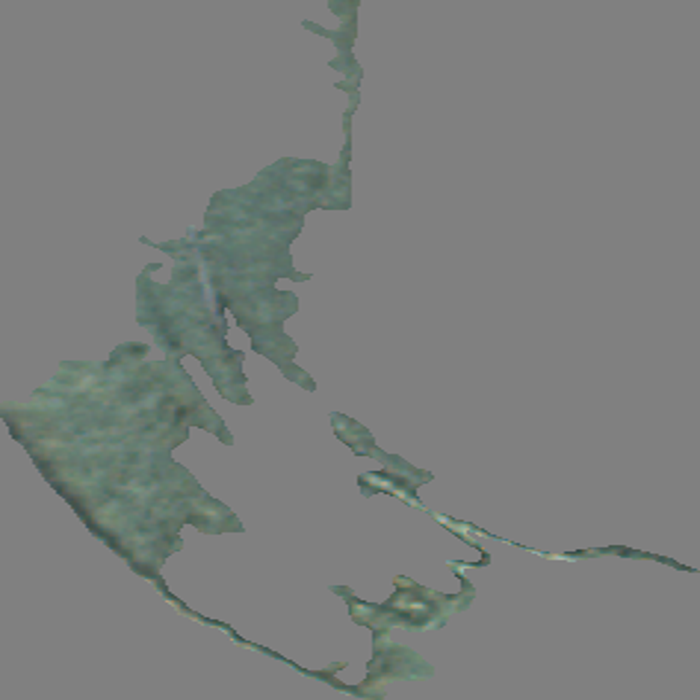}
        \caption{ACE-E: $c_{4}$ $\overline{S}_4=0.86$}\label{fig:ACE-E-4}
    \end{subfigure}
    \begin{subfigure}[t]{0.32\linewidth}
        \centering
        \includegraphics[width=0.24\linewidth]{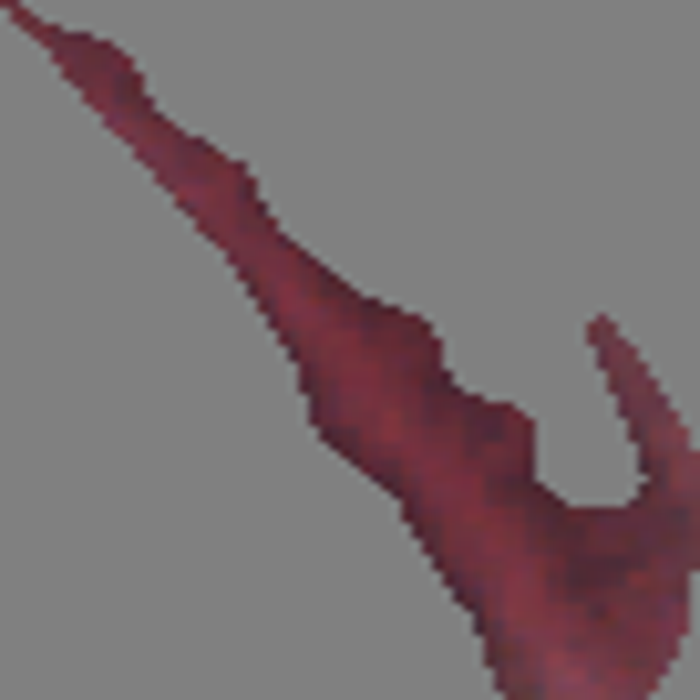}
        \includegraphics[width=0.24\linewidth]{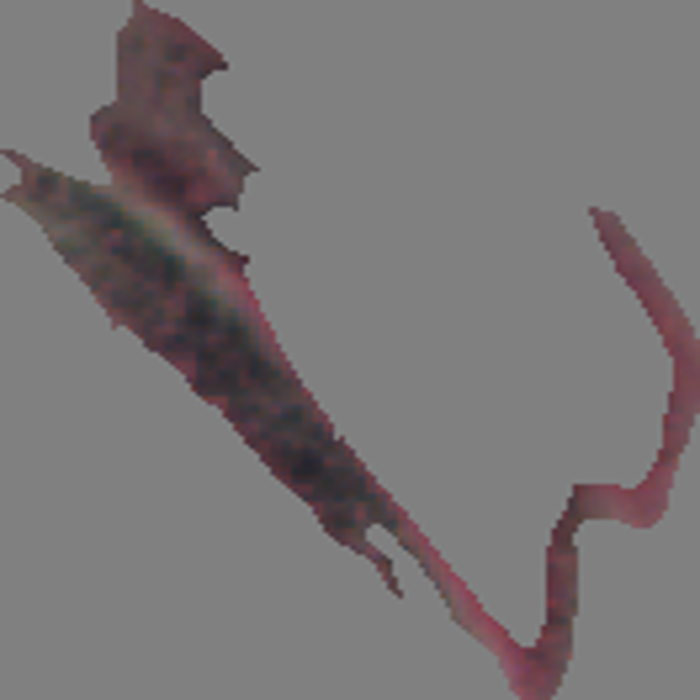}
        \includegraphics[width=0.24\linewidth]{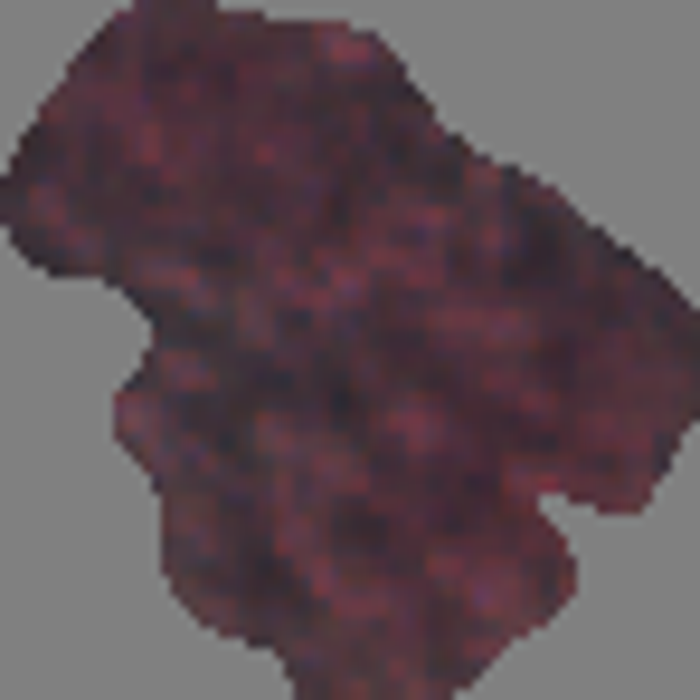}
        \caption{ACE-F: $c_{9}$ $\overline{S}_9=0.94$}\label{fig:ACE-F-9}
    \end{subfigure}
    \begin{subfigure}[t]{0.32\linewidth}
        \centering
        \includegraphics[width=0.24\linewidth]{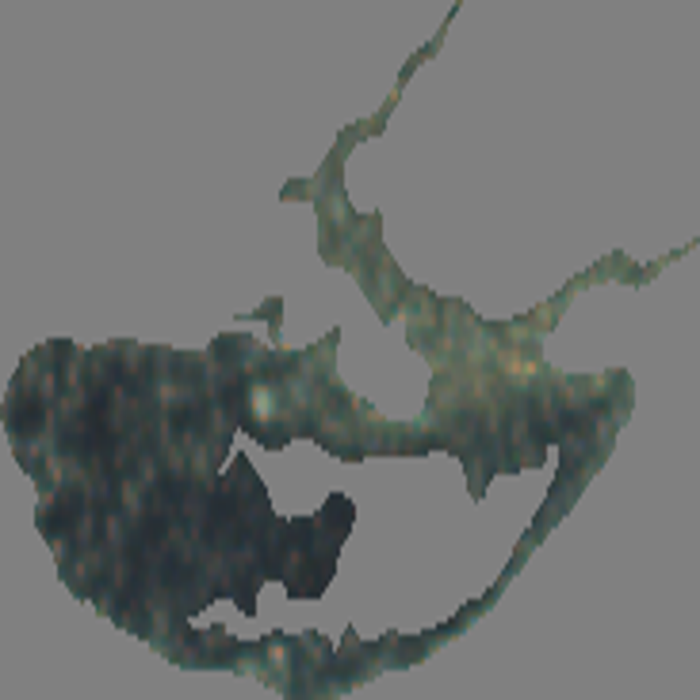}
        \includegraphics[width=0.24\linewidth]{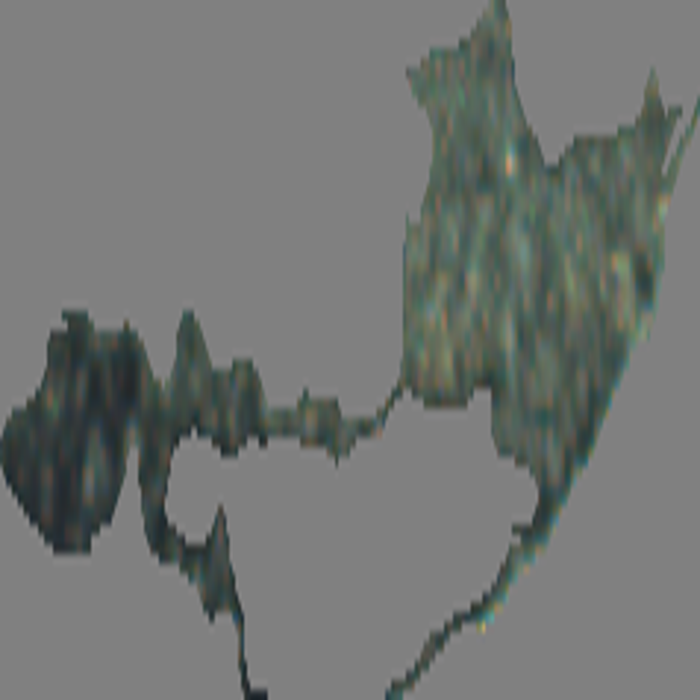}
        \includegraphics[width=0.24\linewidth]{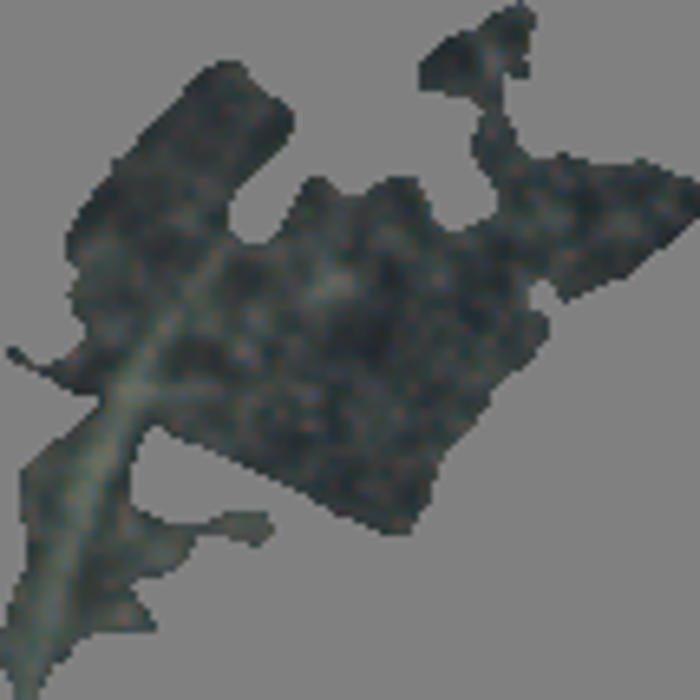}
        \caption{ACE-F: $c_{11}$ $\overline{S}_{11}=0.92$}\label{fig:ACE-F-11}
    \end{subfigure}
\end{minipage}
\caption{MVTec metal nut dataset results for class ``color'' (k=1). Examples of classes in (a) to (e). Top concepts from two SPACE runs and three ACE runs in (f) to (q). Top concepts with high importance from SPACE-C and SPACE-D contain red marks (g), (i), (k), and black or blue marks (f), (j), which are important for the class (as opposed to normal metallic). Concepts from ACE-D contain large segments including red marks (l), (m). In contrast, concepts from ACE-E contain metal pieces (n), (o). Finally, concepts from ACE-F contain red marks (p) and dark metal patches (q).}
\label{fig:metal nut}
\end{figure}

The MVTec metal nuts dataset \citep{8954181} is originally an anomaly detection dataset, which was reframed as a classification task. It consists of five classes, illustrated in Figures \ref{fig:metalnut-class0} to \ref{fig:metalnut-class4}. Each image is of size 700x700 pixels and contains a metal nut in front of a black background. Classes 0 to 4 are bent, color, flipped, ok, and scratched metal nuts, respectively. The numbers of images for each of the classes are imbalanced (e.g., class 3 has 242 images, whereas class 1 has only 22 images in total), which is why resampling and data augmentation were used for the training data. After training, the obtained test accuracy was 100\%. From this dataset, a single class was analyzed with SPACE and ACE, and the importance of the extracted concepts is shown in Figure \ref{fig:metal nut importance scores}. Examples of the concepts with the highest importance scores are shown in Figure \ref{fig:metal nut}.

\textbf{SPACE} extracted three and eighteen concepts for the runs SPACE-C and SPACE-D. The two concepts with the highest importance for each SPACE run are shown in Figures \ref{fig:SPACE-C-1} and \ref{fig:SPACE-D-1}/\ref{fig:SPACE-D-3}. All concepts from SPACE-C were statistically significant with a high importance score, as shown in Figure \ref{fig:metal nut importance scores}. SPACE-D extracted mostly (>80\%) high important concepts, except for $c_{16}$ which was not statistically significant, and $c_{17}$ which had an adverse effect in the predictions of this class.
On the \textbf{ACE} runs, the extracted concepts were of mixed importance. On the ACE-D run, half of the extracted concepts were either not statistically significant or didn't have enough samples for the testing, as shown in Figure \ref{fig:metal nut importance scores}. A similar phenomenon was observed in ACE-E. Finally, ACE-F extracted small patches which were significant more than 80\% of the times.

\textbf{Patches and importance scores}. The number of patches used for each analysis had a significant impact on the results. As a clear difference, SPACE-D extracted smaller and more numerous patches than SPACE-C, as a consequence other less important or significant concepts were extracted. Nonetheless, the concepts with the highest importance contained the expected features. In contrast, the ACE runs extracted superpixels following color and intensity boundaries, which were significantly different for each scale, (see Figures \ref{fig:ACE-D-7}, \ref{fig:ACE-E-4}, and \ref{fig:ACE-F-9}). The scale of the ACE patches did not adversely affect the scale of the importance scores.

\textbf{Alignment}. The features used for the labelling of this class were the colored marks on the metal nuts. At different scales, SPACE performed more consistently, and was able to extract clearly and with high importance the discriminative features of the class. Other concepts were also extracted containing dark metallic regions, which could indicate a bias. In comparison, ACE runs varied significantly, where ACE-D and ACE-F had high importance concepts containing the red marks, but ACE-E did not extract a single concept containing this feature.

\subsection{Metal casting}
\label{sec:results-metal casting}

\begin{figure}[h]
\centering
\includegraphics[width=0.93\linewidth]{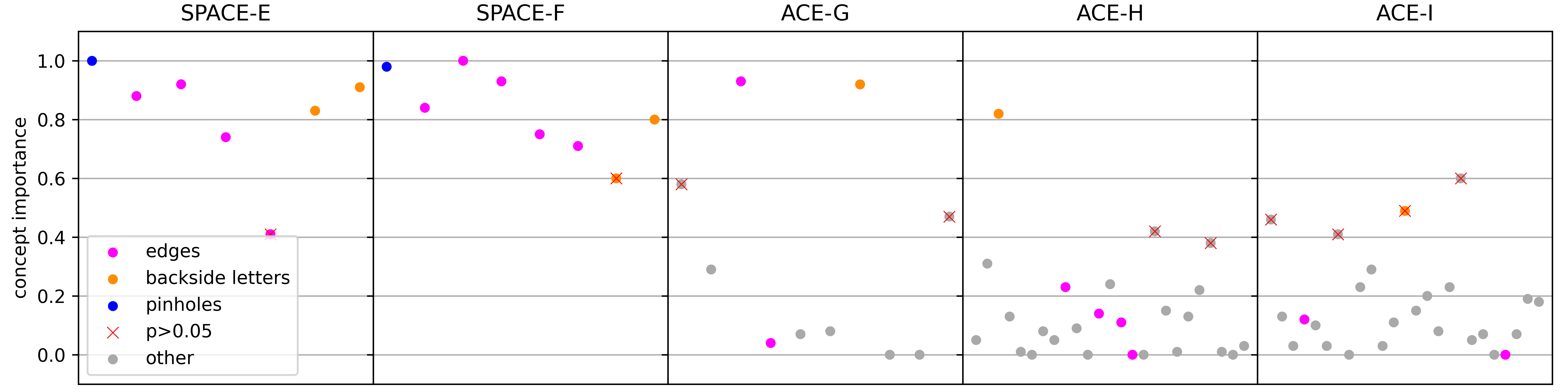}
\caption{Importance scores of concepts for the metal casting dataset. Markers indicate the interpreted content of the concept, and its statistical significance. SPACE consistently extracted and scored similar concepts. ACE failed to extract small concepts, as well as to provide a consistent scoring.}
\label{fig:metal casting importance scores}
\end{figure}

\begin{figure}[h]

\centering
\begin{minipage}{1.0\textwidth}
    \centering
    \begin{subfigure}[t]{0.98\linewidth}
        \centering
        \begin{subfigure}[t]{0.48\linewidth}
            \centering
            \includegraphics[width=0.3\linewidth]{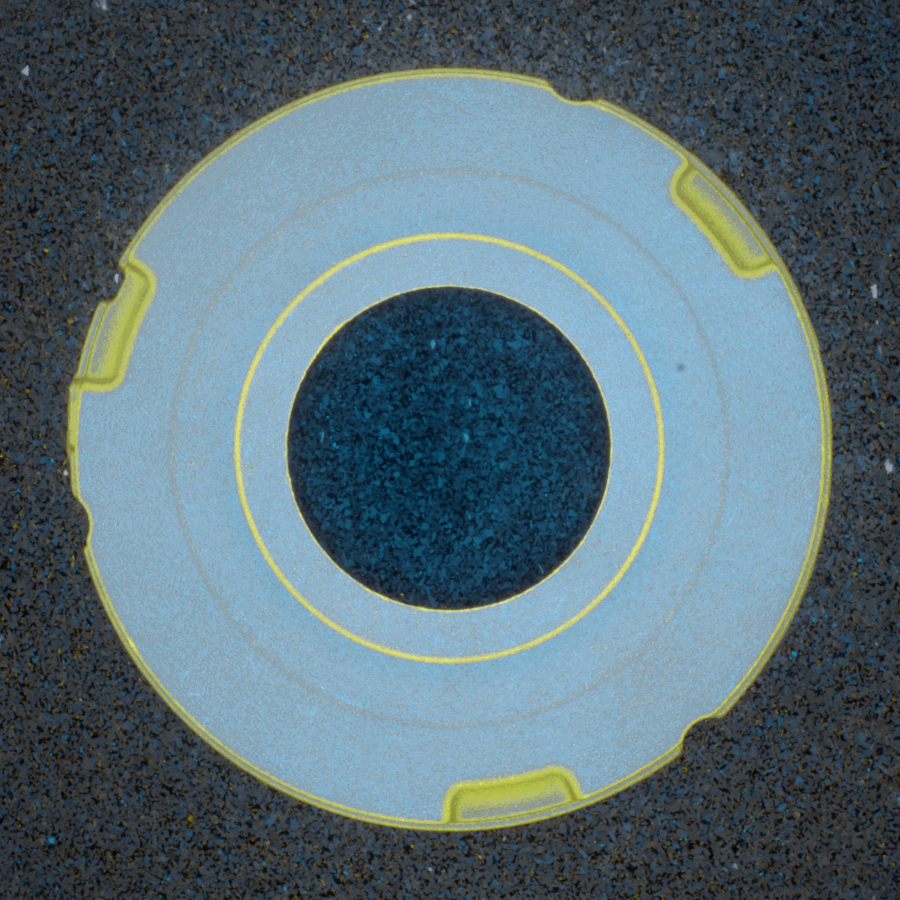}
            \includegraphics[width=0.3\linewidth]{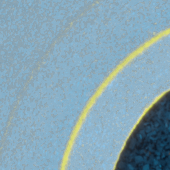}
            \caption{ok and upper left zoom}\label{fig:fg-class0}
        \end{subfigure}
        \begin{subfigure}[t]{0.48\linewidth}
            \centering
            \includegraphics[width=0.3\linewidth]{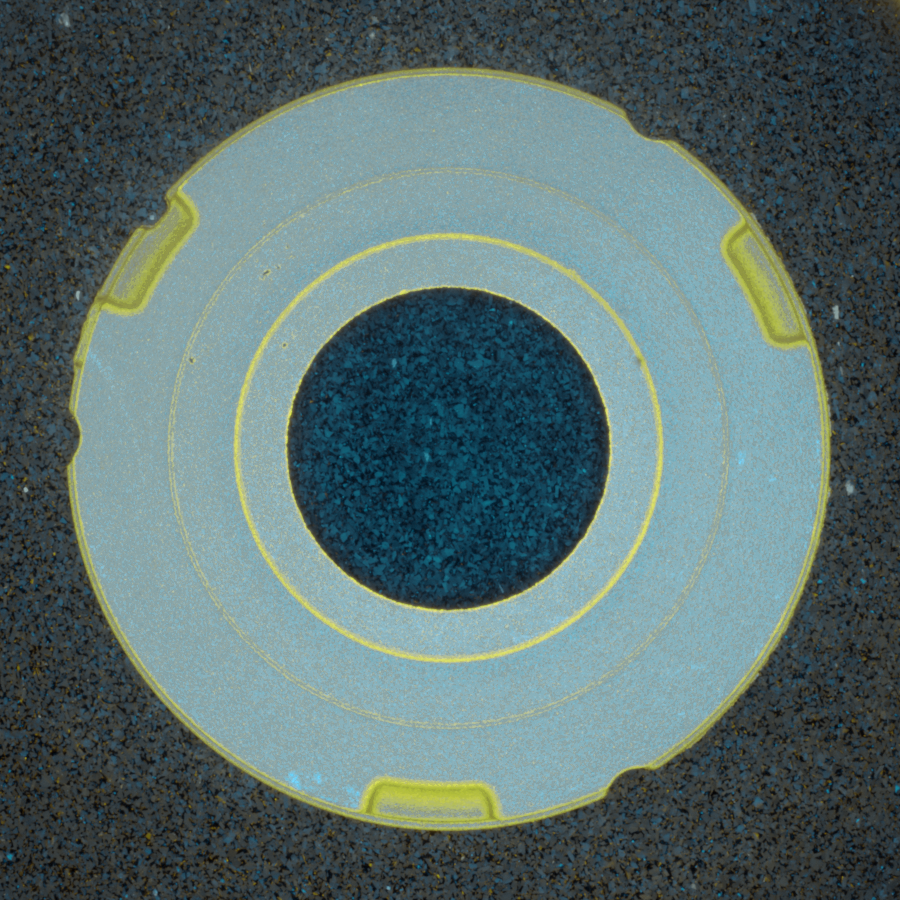}
            \includegraphics[width=0.3\linewidth]{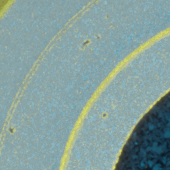}
            \caption{defect and upper left zoom}\label{fig:fg-class1}
        \end{subfigure}
    \end{subfigure}
    
    \begin{subfigure}[t]{0.32\linewidth}
        \centering
        \includegraphics[width=0.24\linewidth]{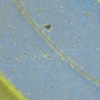}
        \includegraphics[width=0.24\linewidth]{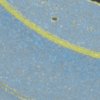}
        \includegraphics[width=0.24\linewidth]{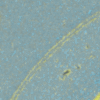}
        \caption{SPACE-E: $c_{0}$ $\overline{S}_0=1.0$}\label{fig:SPACE-E-0}
    \end{subfigure}
    \begin{subfigure}[t]{0.32\linewidth}
        \centering
        \includegraphics[width=0.24\linewidth]{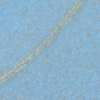}
        \includegraphics[width=0.24\linewidth]{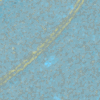}
        \includegraphics[width=0.24\linewidth]{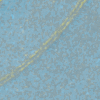}
        \caption{SPACE-E: $c_{2}$ $\overline{S}_2=0.92$}\label{fig:SPACE-E-2}
    \end{subfigure}
    
    \begin{subfigure}[t]{0.32\linewidth}
        \centering
        \includegraphics[width=0.24\linewidth]{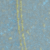}
        \includegraphics[width=0.24\linewidth]{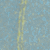}
        \includegraphics[width=0.24\linewidth]{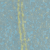}
        \caption{SPACE-F: $c_{2}$ $\overline{S}_1=1.0$}\label{fig:SPACE-F-2}
    \end{subfigure}
    \begin{subfigure}[t]{0.32\linewidth}
        \centering
        \includegraphics[width=0.24\linewidth]{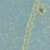}
        \includegraphics[width=0.24\linewidth]{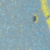}
        \includegraphics[width=0.24\linewidth]{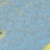}
        \caption{SPACE-F: $c_{0}$ $\overline{S}_3=0.98$}\label{fig:SPACE-F-0}
    \end{subfigure}
    
    \begin{subfigure}[t]{0.32\linewidth}
        \centering
        \includegraphics[width=0.24\linewidth]{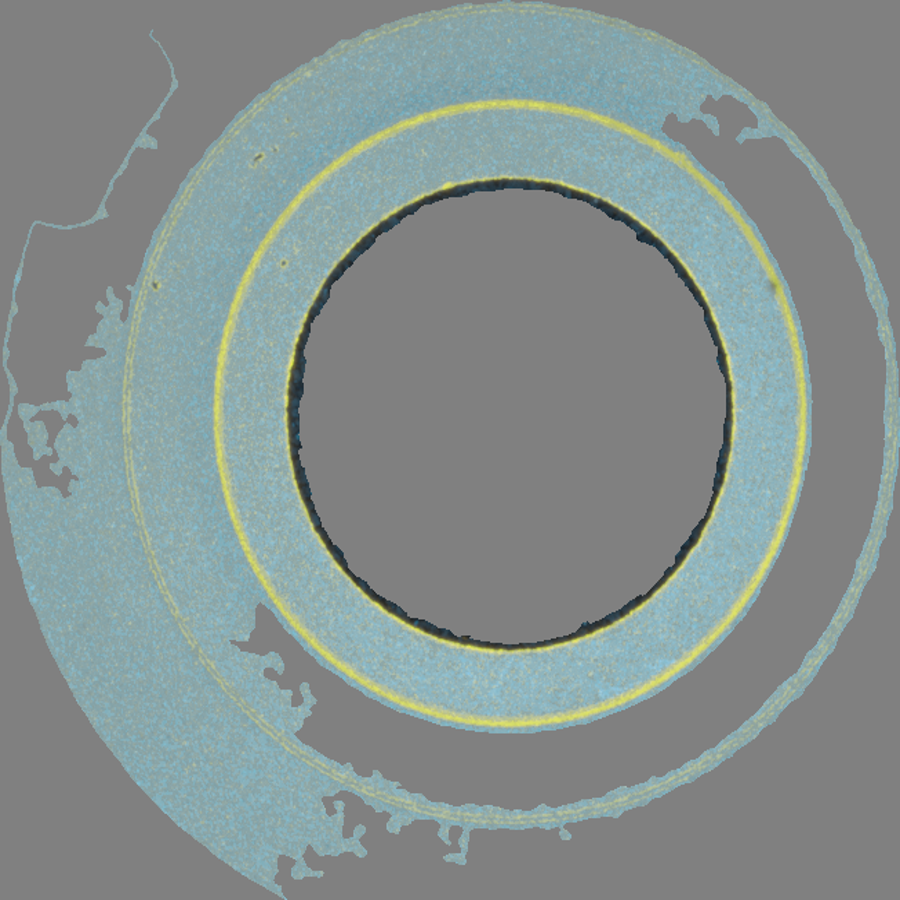}
        \includegraphics[width=0.24\linewidth]{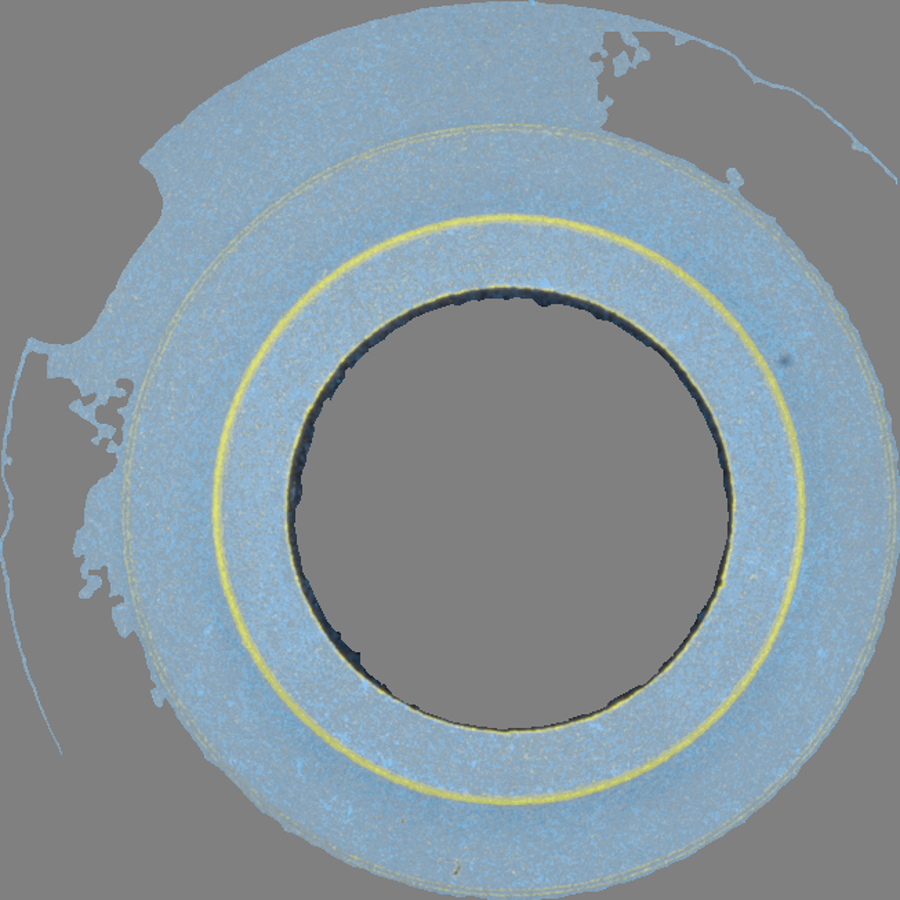}
        \includegraphics[width=0.24\linewidth]{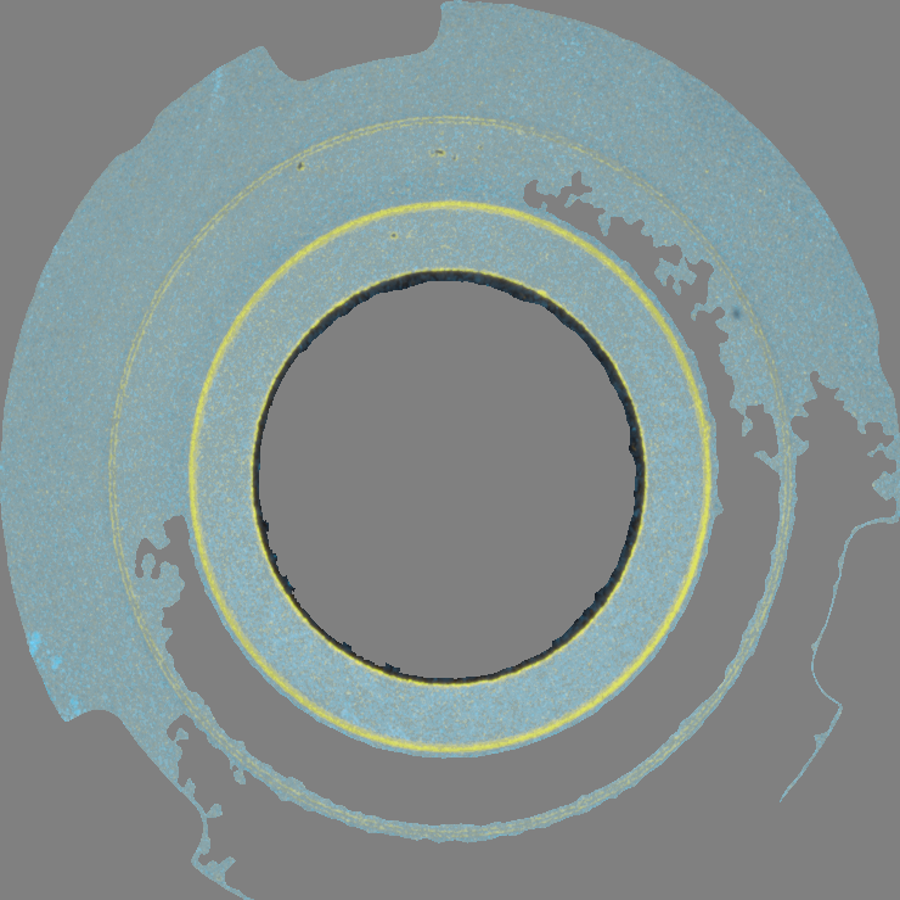}
        \caption{ACE-G: $c_{2}$ $\overline{S}_2=0.93$}\label{fig:ACE-G-2}
    \end{subfigure}
    \begin{subfigure}[t]{0.32\linewidth}
        \centering
        \includegraphics[width=0.24\linewidth]{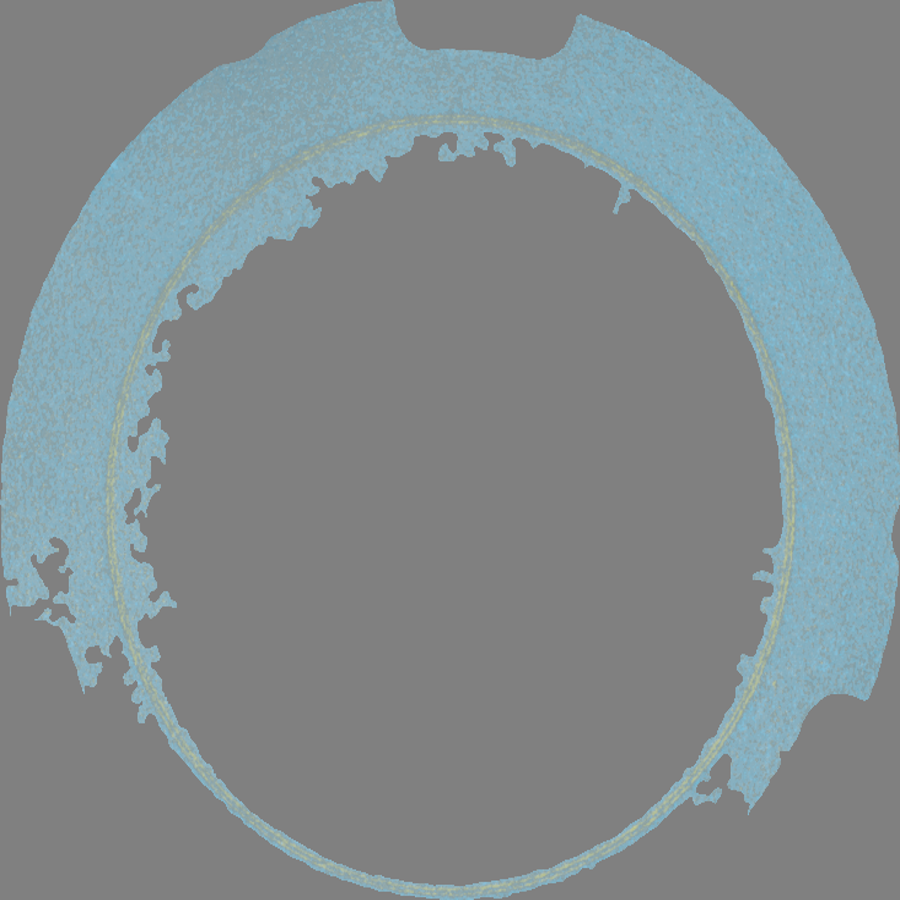}
        \includegraphics[width=0.24\linewidth]{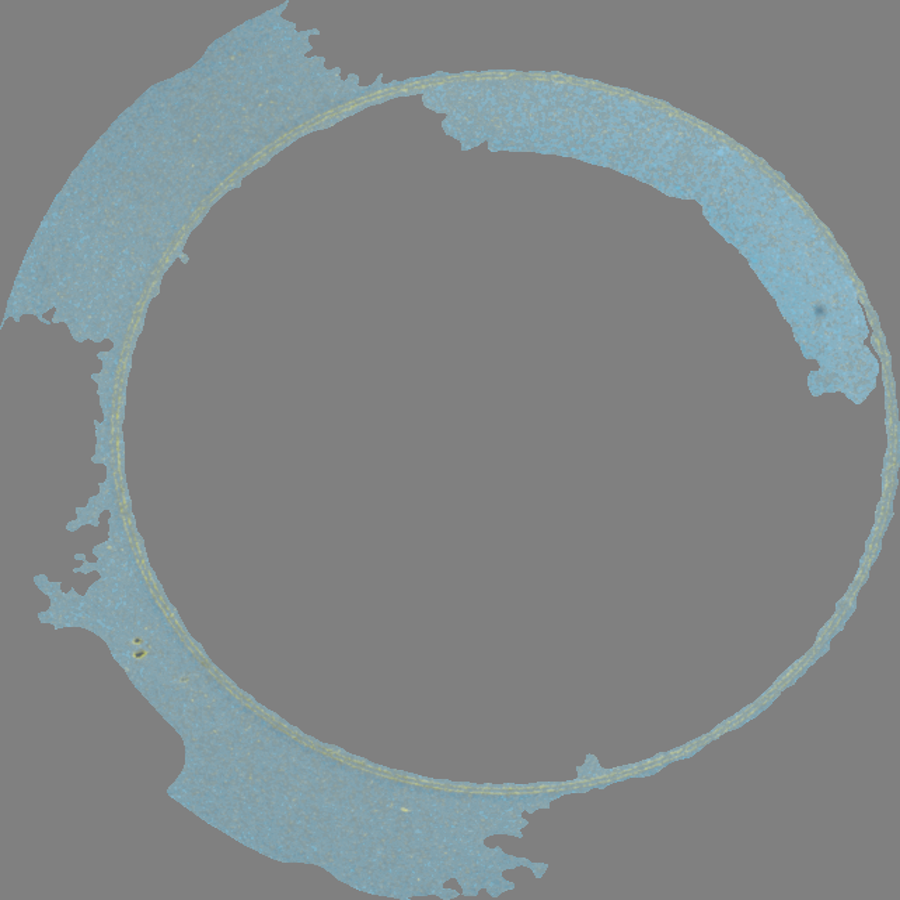}
        \includegraphics[width=0.24\linewidth]{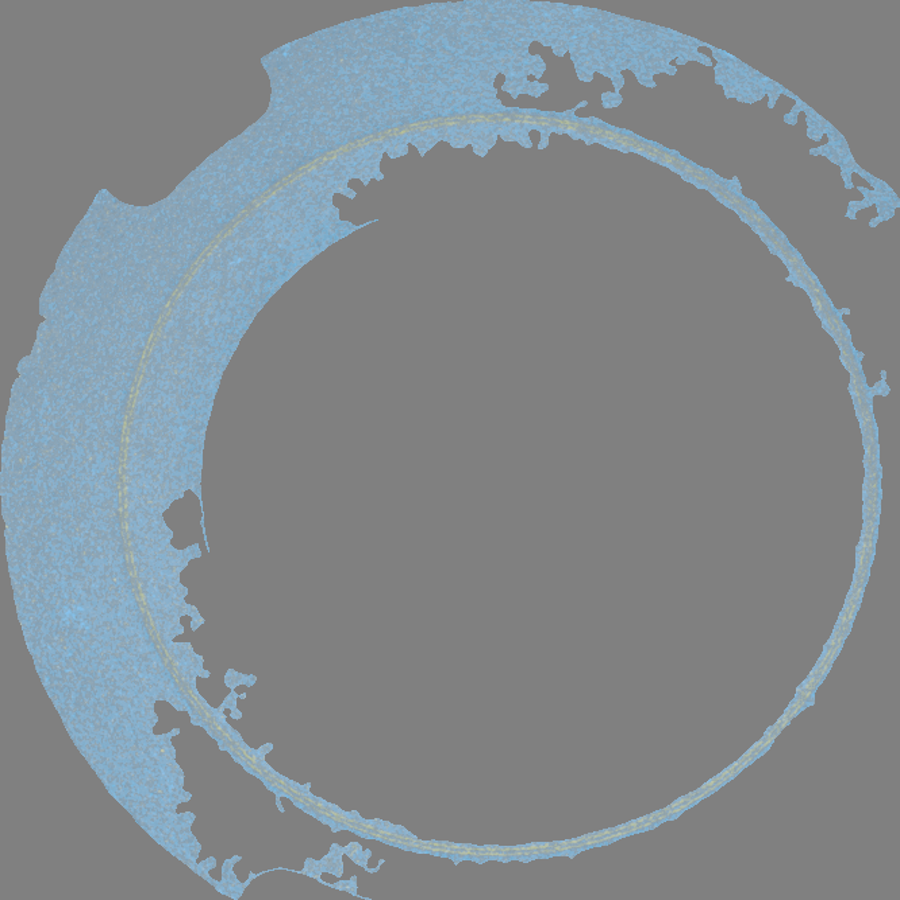}
        \caption{ACE-G: $c_{0}$ $\overline{S}_0=0.58$}\label{fig:ACE-G-0}
    \end{subfigure}
    \begin{subfigure}[t]{0.32\linewidth}
        \centering
        \includegraphics[width=0.24\linewidth]{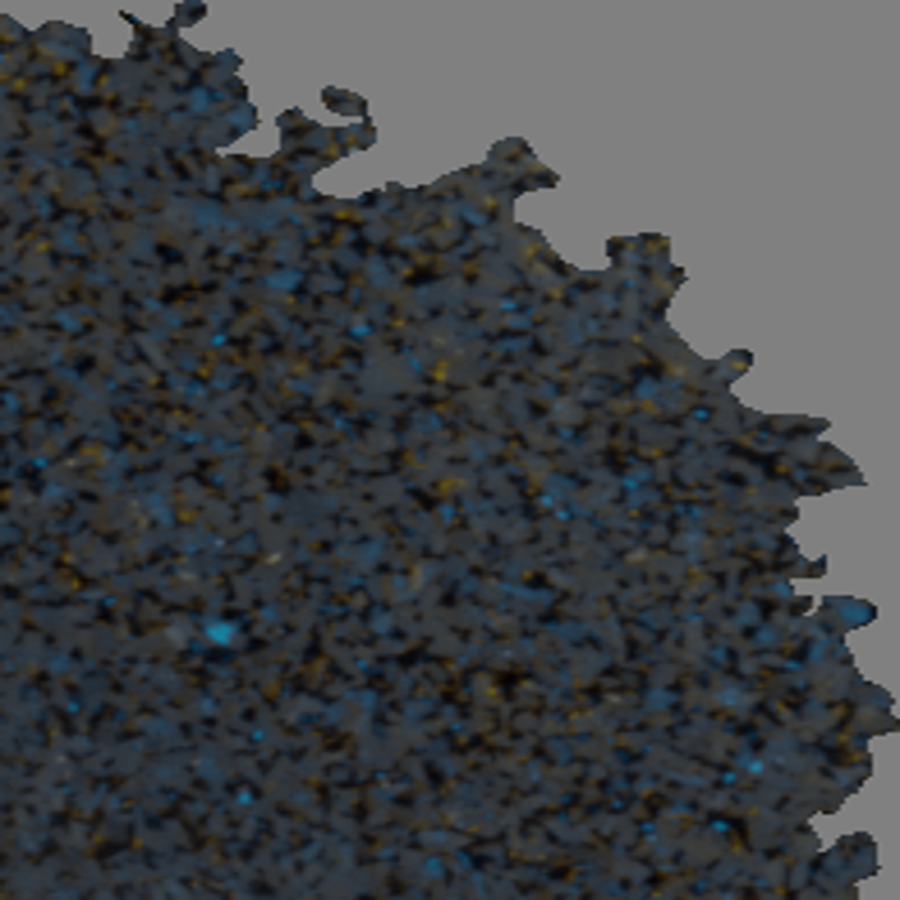}
        \includegraphics[width=0.24\linewidth]{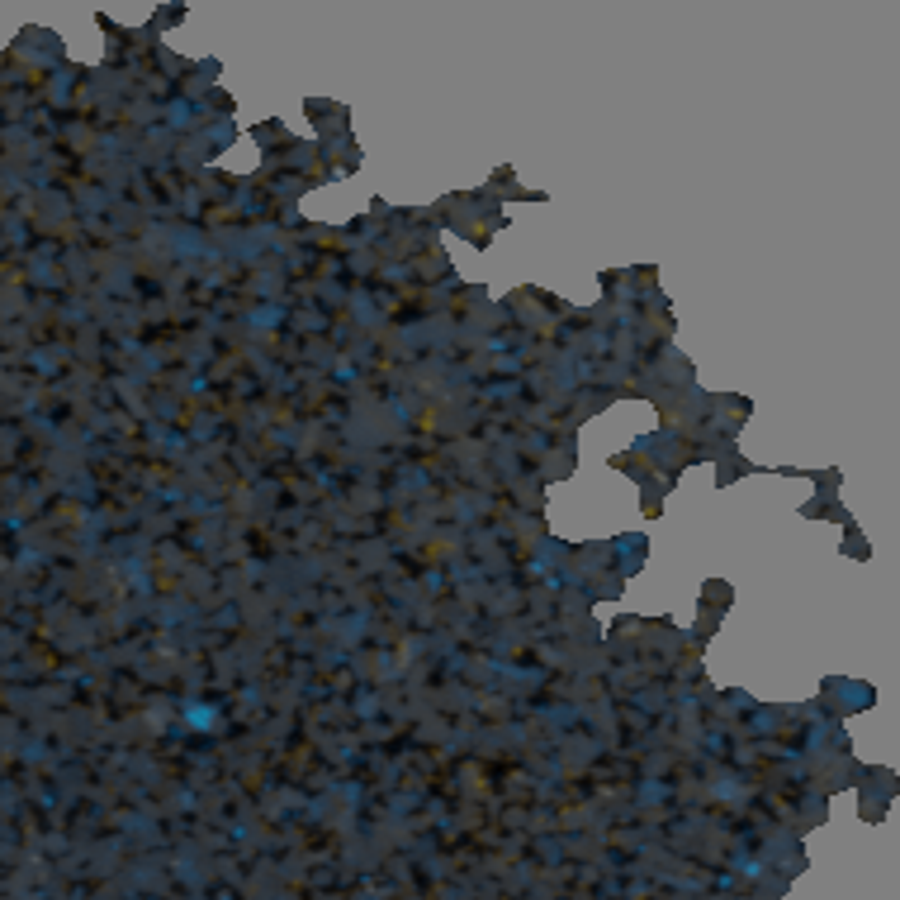}
        \includegraphics[width=0.24\linewidth]{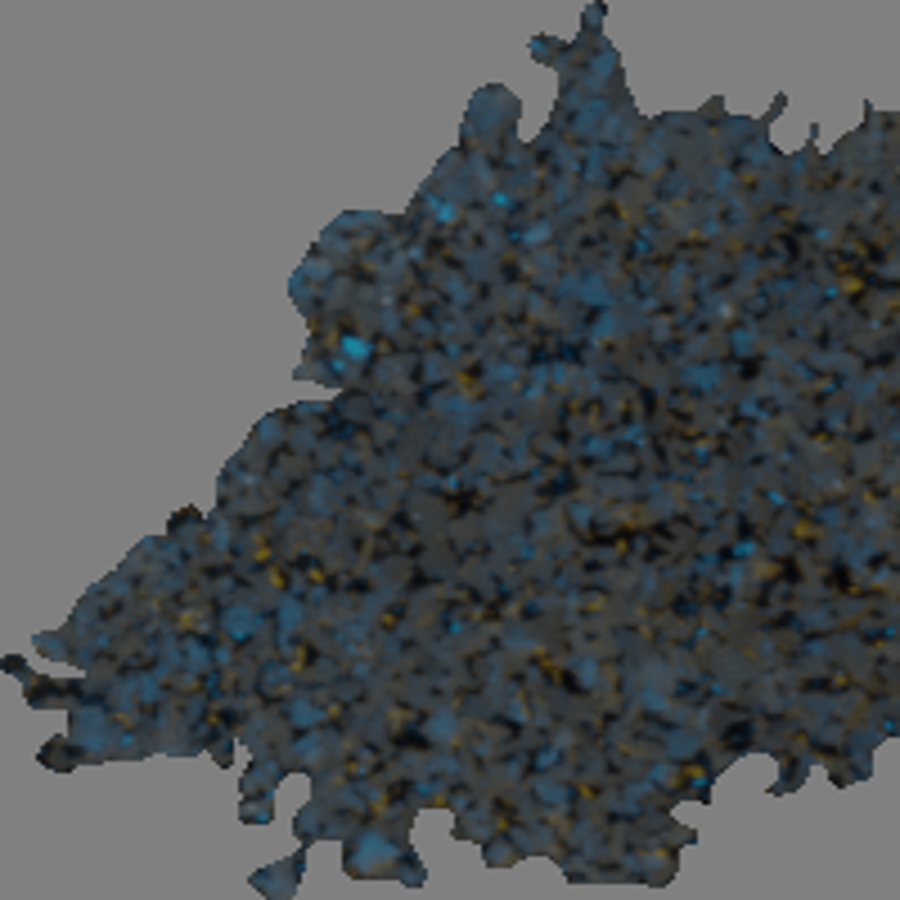}
        \caption{ACE-H: $c_{16}$ $\overline{S}_{16}=0.42$}\label{fig:ACE-H-16}
    \end{subfigure}
    \begin{subfigure}[t]{0.32\linewidth}
        \centering
        \includegraphics[width=0.24\linewidth]{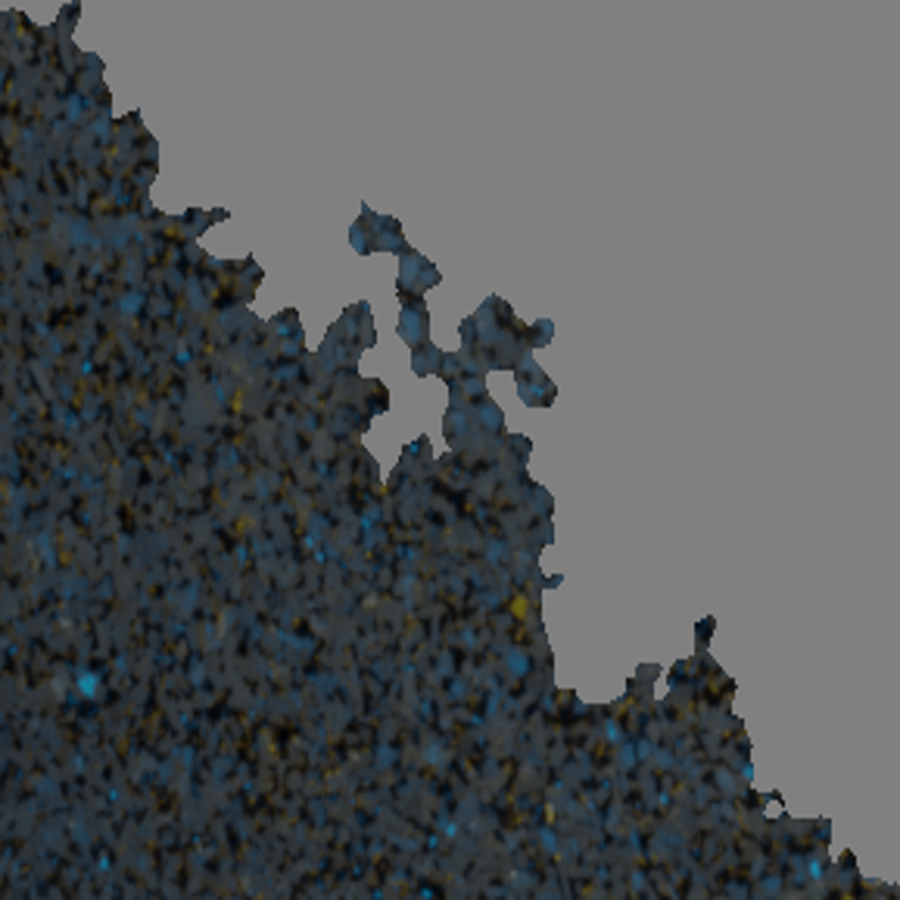}
        \includegraphics[width=0.24\linewidth]{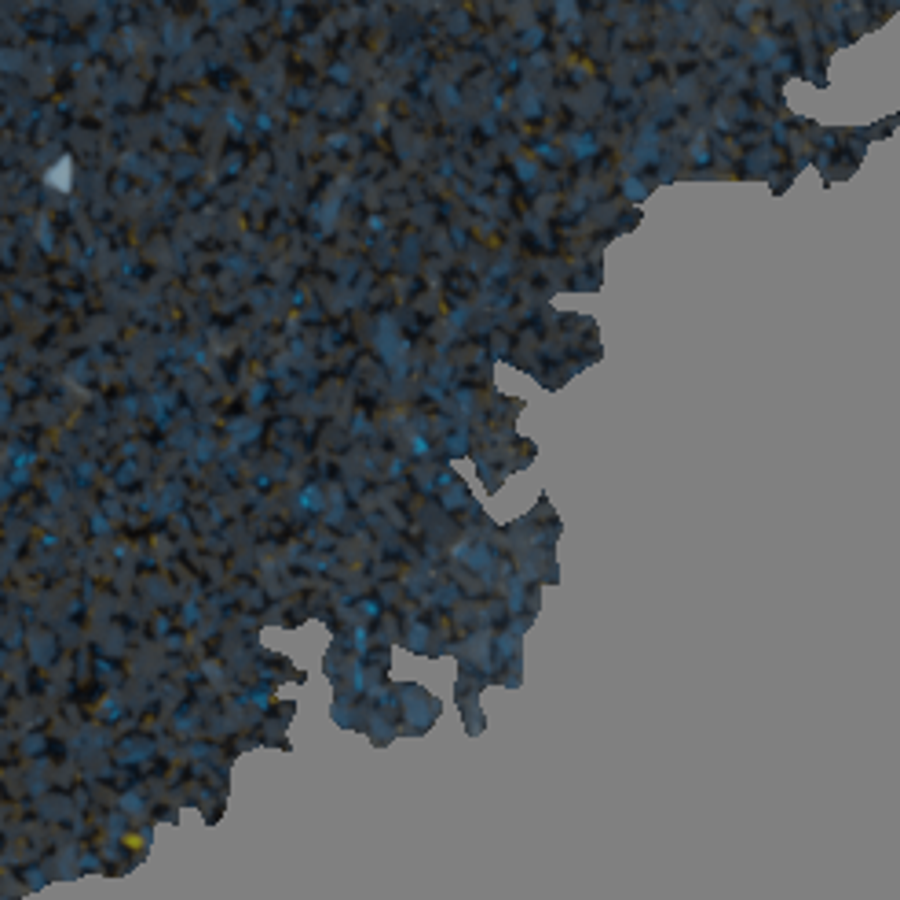}
        \includegraphics[width=0.24\linewidth]{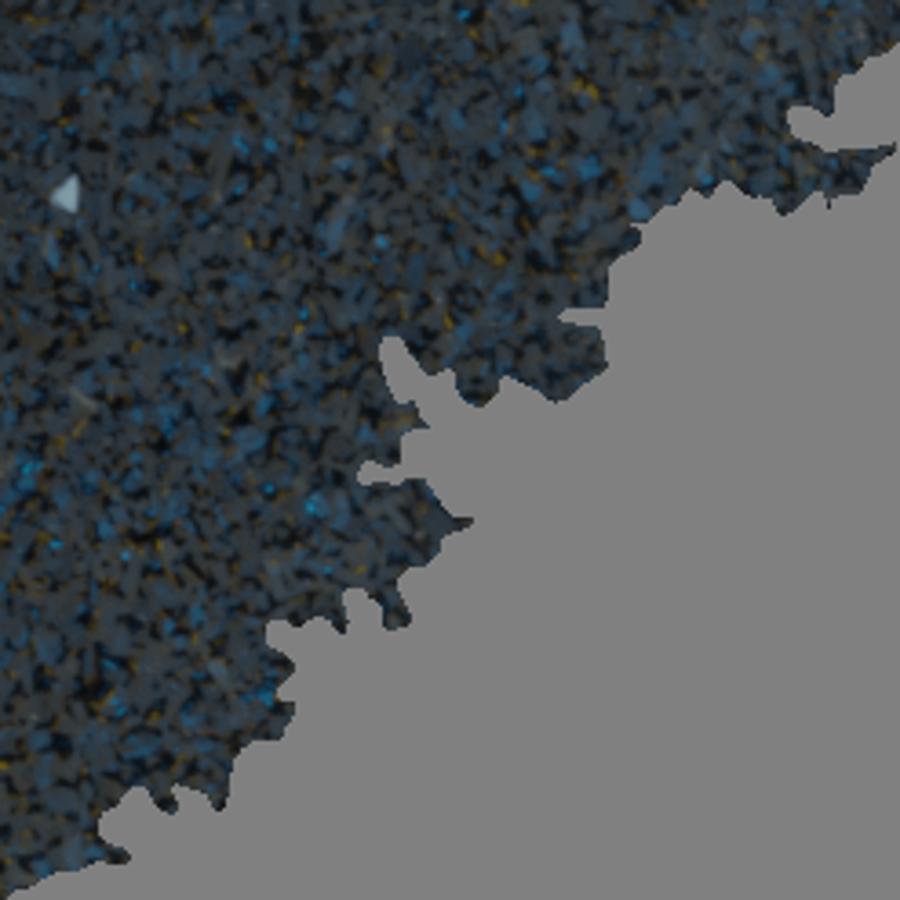}
        \caption{ACE-H: $c_{21}$ $\overline{S}_{21}=0.38$}\label{fig:ACE-H-21}
    \end{subfigure}
    \begin{subfigure}[t]{0.32\linewidth}
        \centering
        \includegraphics[width=0.24\linewidth]{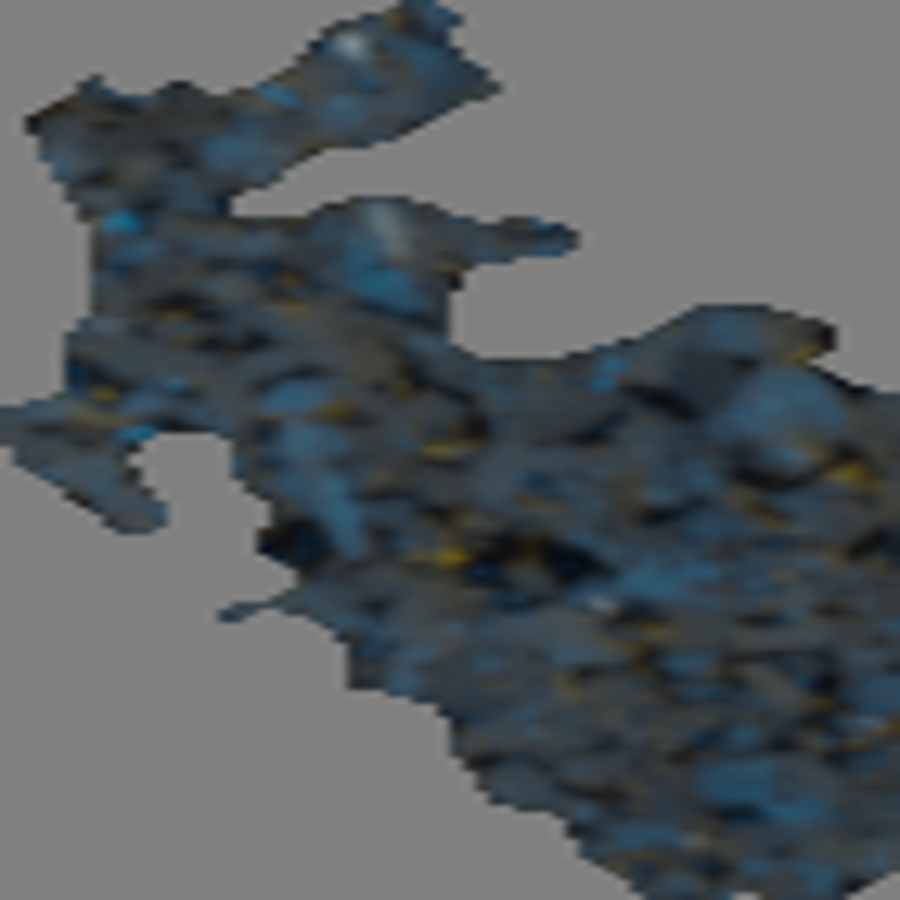}
        \includegraphics[width=0.24\linewidth]{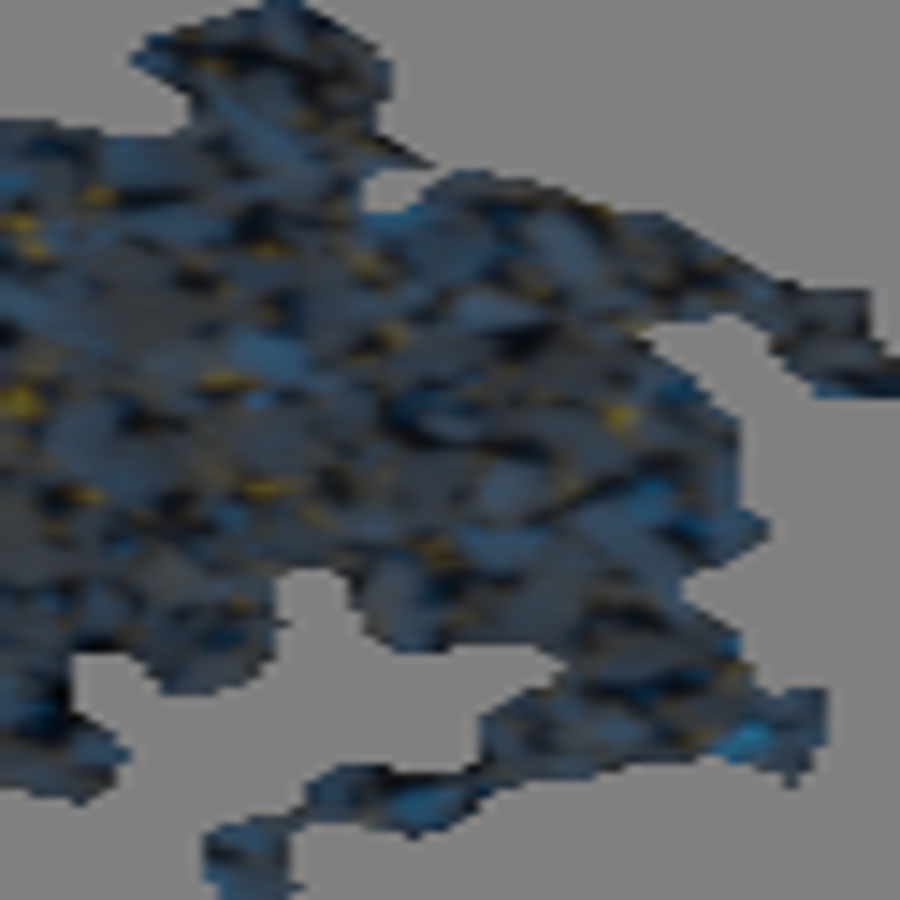}
        \includegraphics[width=0.24\linewidth]{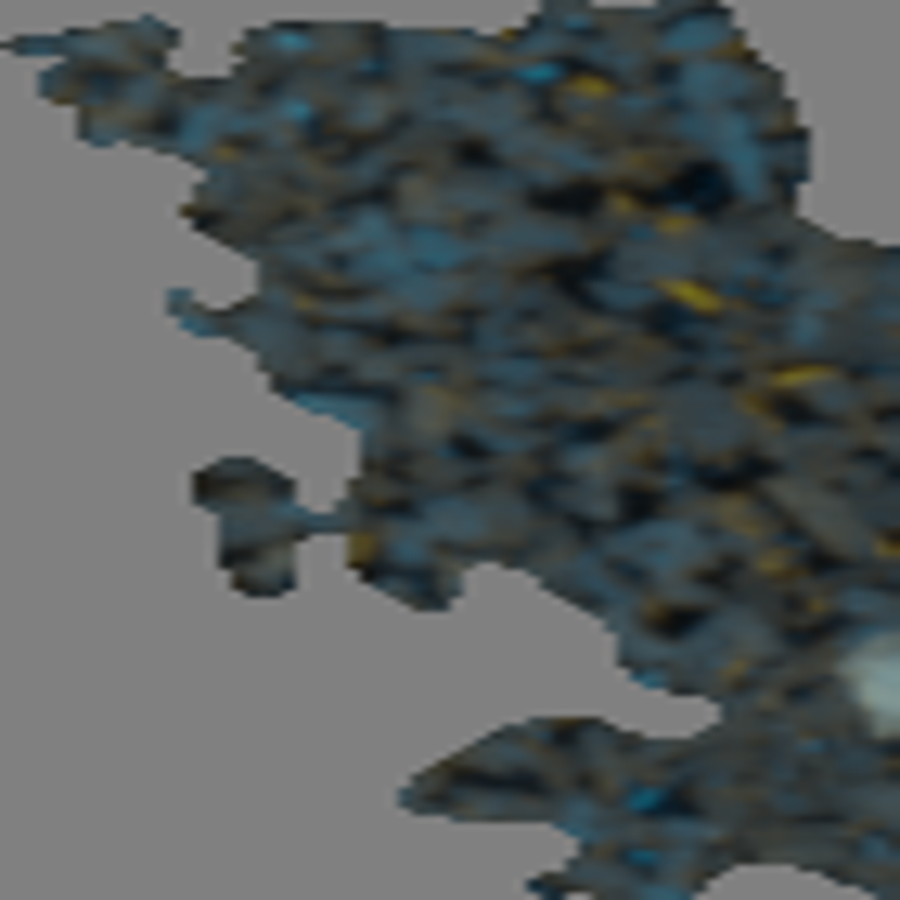}
        \caption{ACE-I: $c_{17}$ $\overline{S}_{17}=0.60$}\label{fig:ACE-I-17}
    \end{subfigure}
    \begin{subfigure}[t]{0.32\linewidth}
        \centering
        \includegraphics[width=0.24\linewidth]{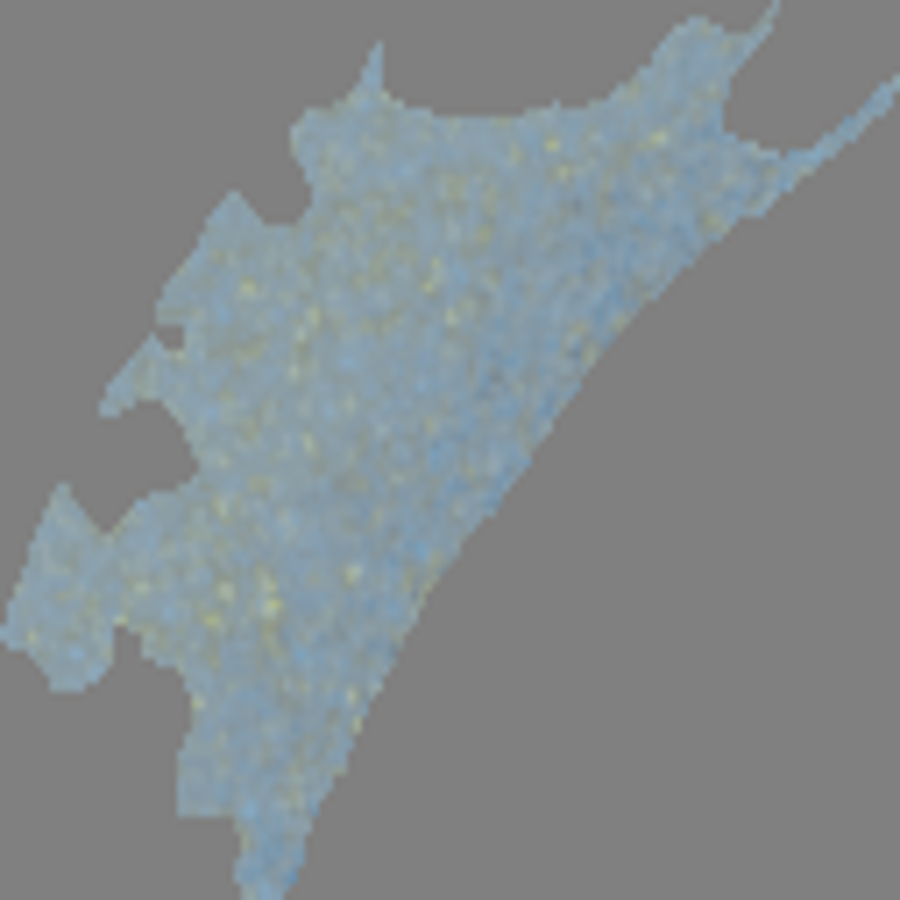}
        \includegraphics[width=0.24\linewidth]{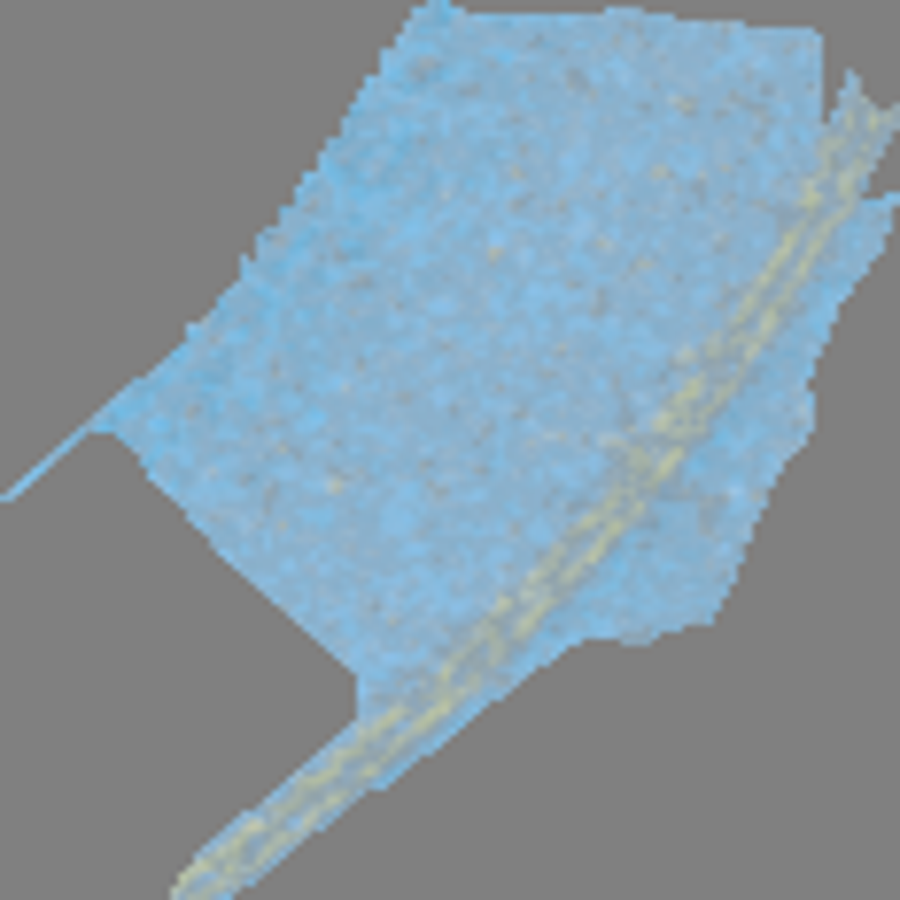}
        \includegraphics[width=0.24\linewidth]{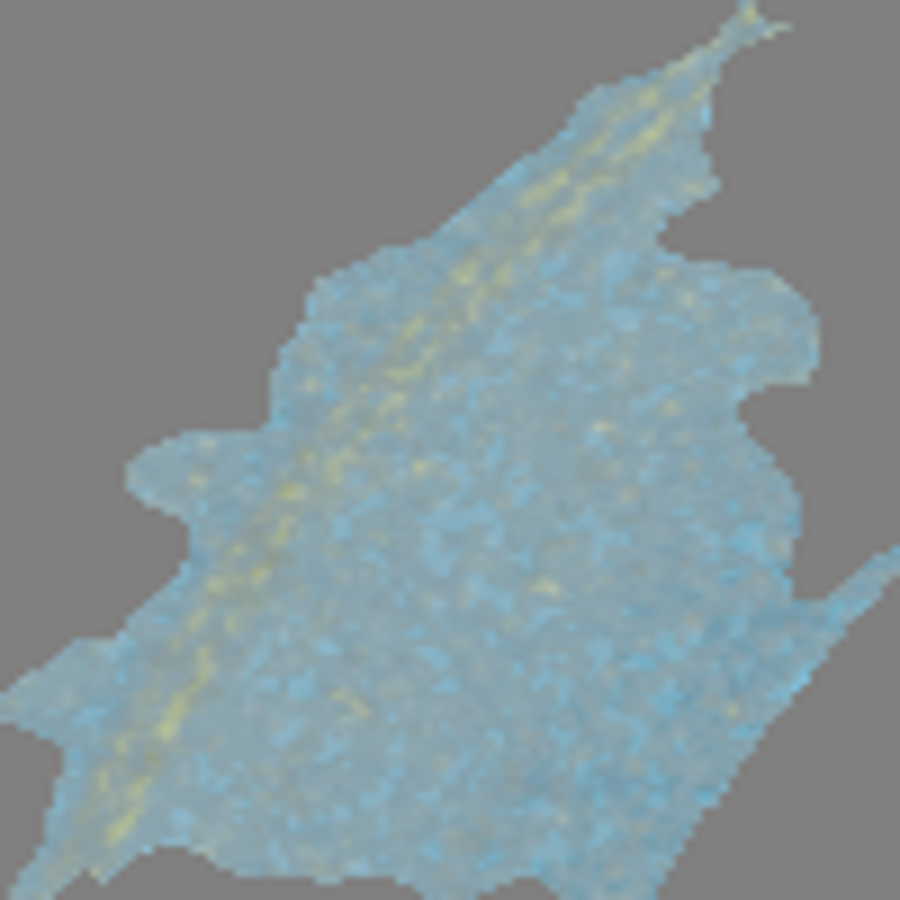}
        \caption{ACE-I: $c_{0}$ $\overline{S}_0=0.46$}\label{fig:ACE-I-0}
    \end{subfigure}
\end{minipage}
\caption{Casting dataset results for class ``defect'' (k=1). Examples of classes in (a) and (b). Two of the top three concepts from two SPACE runs and three ACE runs in (c) to (l). Top concepts of SPACE-A and SPACE-B contain pinholes (c), (f), which is the feature used for labelling the class; They also contain yellow lines (d), (e), which is a detected bias. Top concepts from ACE-G contain large segments with yellow lines (g) and large segments of metal (h). Similarly, concepts from ACE-H and ACE-I contain segments of the background (i), (j), (k), and patches of metal (l). One concept per ACE run was excluded due to confidentiality.}
\label{fig:metal casting}
\end{figure}

The last dataset was obtained during an actual application of an automated quality control process with an industrial partner (Deevio GmbH). It consists of images of a part manufactured through metal casting. The two labeled classes were class 0 consisting of 139 images without defects (e.g., \ref{fig:fg-class0}) and class 1 consisting of 141 images containing defects (e.g., \ref{fig:fg-class1}). Each image has a size of 900x900 pixels, and the dataset is composed of images from the front and back of the casting part. The predominant defect that was labeled during the experiment were pinholes, which are common small casting defects that generate a small yet visible porosity in the part. After training the mentioned classification model, the obtained test accuracy was 100\%. The importance of the extracted concepts for all runs are shown in Figure \ref{fig:metal casting importance scores} and examples of the concepts with the highest importance scores are shown in Figure \ref{fig:metal casting}.

\textbf{SPACE} extracted seven and eight concepts for runs SPACE-E and SPACE-F. For each SPACE run, the extracted concepts had an importance score between 0.71 and 1.0, except for one concept per run that was not statistically significant. The content of the concepts contained either pinholes, a yellow line from an edge, or letters from the backside of the parts, as highlighted in Figure \ref{fig:metal casting importance scores}. Respecting confidentiality, the images of the concepts containing the letters from the backside of the parts are omitted in the results. In contrast, the concepts extracted from the \textbf{ACE} runs either contained a big segment of the image (e.g., Figure \ref{fig:ACE-G-2}), or had a significantly lower importance score (e.g., Figure \ref{fig:ACE-I-0} and \ref{fig:metal casting importance scores}). 

\textbf{Patches and importance scores}. The two SPACE runs obtained similar results, locating semantically equivalent areas (pinholes, yellow lines, and backside letter). Similarly, the importance scores obtained in the two runs were comparable. In both SPACE runs, the concepts containing pinholes were within the two concepts with the highest important scores. ACE runs achieved different results, in ACE-G large portions of the parts were extracted and labeled with high importance. These concepts (see Figure \ref{fig:ACE-G-2}) did not contain specifically pinholes, but they did contain the yellow lines mentioned before. No concept from the ACE runs contained pinholes specifically. This can be due to the severe interpolation required for the small patches and the lack of specificity of the bigger ones. This is also a possible cause for a predominantly low importance score for the ACE runs.

\textbf{Alignment}. The features used for the labeling of this dataset were the pinholes. On the SPACE runs, it was clear that the pinholes were important for the trained models, but the yellow lines and backside letters were also important. On the ACE runs, only bigger features such as the yellow lines were extracted and important scores were predominantly low. This erroneously hinted towards the model only using the yellow lines for the prediction.

\textbf{Impact}. This dataset was obtained in a real-world scenario, and the current method was used to verify its decision-making process. Through this analysis, it was detected that the model was also using the yellow lines and backside letters for its predictions, which in this context represents a \textit{target leakage}. In general, target leakage occurs when the training data contains information about the target variable that will not be available at prediction time \citep{10.1145/2382577.2382579}. In this case, the target leakage not only leads to learning concepts which are not directly helpful to solve the task, but can also become an unexpected bias and diminish robustness during the deployment of the models. After consulting with the domain expert, we learned that defective parts had been created from a different mold than non-defective parts. The different molds had subtle but significant differences in shape and identifying letters, which were learned by the model. Thanks to the detection of data leakage, mitigation was possible with further actions.

\subsection{Conclusion}

Two important phenomena were observed through all experiments. First, the encoding proposed by SPACE proves to be superior when analyzing scale variant models, introducing fewer artifacts when composing bigger images from the extracted patches. This impacts the concept clustering and importance testing as the encoding are closer to what the networks have learned. In contrast, using interpolation techniques for resizing patches modifies the scale-dependent features of the model. This had a significant impact when testing concepts containing pinholes. Second, SPACE improves concept extraction when analyzing features related to edges, or boundaries (e.g., cracks). The main reason for this is that the square slicing and tiling proposed by SPACE is not biased by color or intensity boundaries in images (as opposed to superpixels). Third, SPACE extracts fewer concepts than ACE, yet, the extracted concepts are more aligned to the semantic meaning of the analyzed classes. This is a consequence of guiding the patch extraction step by using the aggregated local importance of the patches.


\section{Discussion and concluding remarks}
\label{sec:conclusions}

The current work proposes the algorithm SPACE, which is specifically designed to perform concept extraction in industrial settings. SPACE was then tested over three datasets representative of real-world industrial quality control problems. These datasets contained relevant features of different scales captured from a single perspective. In these scenarios, the results of SPACE outperform ACE.

The techniques introduced by SPACE for patch extraction, image composition and concept testing enable a better concept extraction in quality control. Specifically, SPACE should be preferred, when analyzing models sensitive to scale changes and when the intended features of the labeled classes are small or related to intensity and color boundaries.

In real-world scenarios (e.g. metal casting), a model can show a perfect accuracy during training and testing. However, the model may contain biases and be influenced by data leakages, which will reflect in a poor performance on unseen data. During this study, two data leakages were identified (yellow lines and backside letters of the metal casting study). It was SPACE which allowed a clear identification of the unintended leakages and biases of the data. In cases like this one, actions can be taken to obtain more robust models which are better suited for industrial deployment.

Throughout this work, SPACE has shown a reliable performance in multiple use cases. Yet, some points must be considered when applying it. First, the number of slices $n_s$ must respect the general size of the features that are analyzed. If the selected number of slices yields patches smaller than the general features of the dataset, suboptimal results will be obtained. Second, SPACE should not be used in cases where datasets already contain features similar to the discontinuities introduced during the tiling process of SPACE.

While SPACE has shown significant improvement over the state-of-the-art when it comes to scale-sensitive problems, there are several aspects worth of consideration in future work. First, on a practical side, more research is required on concept-based explanations and how it fits compliance frameworks in the industry. Second, the topic of offering guarantees for concept-based explanations can have a huge impact on the adoption of these techniques in practical applications. Finally, two important points of this work are the choice of feature representations (the extracted feature maps) and the explanation mediums (the images presented to the users). Exploring different alternatives of both can allow a more precise extraction of local concepts and a more complete information flow towards the users.
\section{Declarations}
\textbf{Funding} Funded by the Deutsche Forschungsgemeinschaft (DFG, German Research Foundation) under Germany’s Excellence Strategy-EXC-2023 Internet of Production-390621612. 

\noindent\textbf{Authors contributions} 
The conceptualization and study design was performed by AP, LK, and TG; The software implementation was written by LK and AP; The industrial data was provided by TG; The scientific supervision was performed by ST; The first draft of the manuscript was written by AP, later edited together with LK, and ST. All authors
reviewed the results, commented during the writing process, and approved the final version of the manuscript. 

\noindent\textbf{Conflict of interest} The authors declare that they have no conflicts of interest/competing interests.

\noindent\textbf{Availability of data and material} Two of the used datasets are publicly available from previous studies \citep{8954181,conc}.

\noindent\textbf{Code availability} The source code is available at:\\
\repolink

\noindent\textbf{Ethics approval} Not applicable.

\noindent\textbf{Consent to participate} Not applicable.

\noindent\textbf{Consent for publication} Not applicable.


%
%

\bibliography{literature_AP}   

\begin{thebibliography}{36}
\providecommand{\natexlab}[1]{#1}
\providecommand{\url}[1]{\texttt{#1}}
\expandafter\ifx\csname urlstyle\endcsname\relax
  \providecommand{\doi}[1]{doi: #1}\else
  \providecommand{\doi}{doi: \begingroup \urlstyle{rm}\Url}\fi

\bibitem[Achanta et~al.(2012)Achanta, Shaji, Smith, Lucchi, Fua, and
  Süsstrunk]{6205760}
R.~Achanta, A.~Shaji, K.~Smith, A.~Lucchi, P.~Fua, and S.~Süsstrunk.
\newblock Slic superpixels compared to state-of-the-art superpixel methods.
\newblock \emph{IEEE Transactions on Pattern Analysis and Machine
  Intelligence}, 34\penalty0 (11), 2012.

\bibitem[Ankerst et~al.(1999)Ankerst, Breunig, Kriegel, and
  Sander]{10.1145/304182.304187}
M.~Ankerst, M.~M. Breunig, H.-P. Kriegel, and J.~Sander.
\newblock Optics: Ordering points to identify the clustering structure.
\newblock In \emph{Proceedings of the 1999 ACM SIGMOD International Conference
  on Management of Data}, SIGMOD '99, page 49–60, New York, NY, USA, 1999.
  Association for Computing Machinery.
\newblock ISBN 1581130848.

\bibitem[Arrieta et~al.(2020)Arrieta, D\'{i}az-Rodr\'{i}guez, Del~Ser,
  Bennetot, Tabik, Barbado, Garc\'{i}a, Gil-L\'{o}pez, Molina, Benjamins,
  et~al.]{arrieta2020explainable}
A.~B. Arrieta, N.~D\'{i}az-Rodr\'{i}guez, J.~Del~Ser, A.~Bennetot, S.~Tabik,
  A.~Barbado, S.~Garc\'{i}a, S.~Gil-L\'{o}pez, D.~Molina, R.~Benjamins, et~al.
\newblock Explainable artificial intelligence (xai): Concepts, taxonomies,
  opportunities and challenges toward responsible ai.
\newblock \emph{Information Fusion}, 58:\penalty0 82--115, 2020.

\bibitem[Augasta and Kathirvalavakumar(2012)]{Augasta2012}
M.~G. Augasta and T.~Kathirvalavakumar.
\newblock Reverse engineering the neural networks for rule extraction in
  classification problems.
\newblock \emph{Neural Processing Letters}, 35\penalty0 (2):\penalty0 131--150,
  4 2012.
\newblock ISSN 1573-773X.

\bibitem[Bergmann et~al.(2019)Bergmann, Fauser, Sattlegger, and
  Steger]{8954181}
P.~Bergmann, M.~Fauser, D.~Sattlegger, and C.~Steger.
\newblock Mvtec ad — a comprehensive real-world dataset for unsupervised
  anomaly detection.
\newblock In \emph{2019 IEEE/CVF Conference on Computer Vision and Pattern
  Recognition (CVPR)}, pages 9584--9592, 2019.

\bibitem[Burkart and Huber(2021)]{10.1613/jair.1.12228}
N.~Burkart and M.~F. Huber.
\newblock A survey on the explainability of supervised machine learning.
\newblock \emph{J. Artif. Int. Res.}, 70:\penalty0 245–317, May 2021.
\newblock ISSN 1076-9757.

\bibitem[\c{C}a\u{g}lar F\i{}rat~\"{O}zgenel(2019)]{conc}
\c{C}a\u{g}lar F\i{}rat~\"{O}zgenel.
\newblock Concrete crack images for classification.
\newblock 2, 2019.
\newblock \doi{10.17632/5Y9WDSG2ZT.2}.

\bibitem[Chen et~al.(2019)Chen, Li, Tao, Barnett, Rudin, and
  Su]{DBLP:conf/nips/ChenLTBRS19}
C.~Chen, O.~Li, D.~Tao, A.~Barnett, C.~Rudin, and J.~Su.
\newblock This looks like that: Deep learning for interpretable image
  recognition.
\newblock In H.~M. Wallach, H.~Larochelle, A.~Beygelzimer,
  F.~d'Alch{\'{e}}{-}Buc, E.~B. Fox, and R.~Garnett, editors, \emph{Advances in
  Neural Information Processing Systems 32: Annual Conference on Neural
  Information Processing Systems 2019, NeurIPS 2019, December 8-14, 2019,
  Vancouver, BC, Canada}, pages 8928--8939, 2019.

\bibitem[Chen et~al.(2020)Chen, Bei, and
  Rudin]{DBLP:journals/corr/abs-2002-01650}
Z.~Chen, Y.~Bei, and C.~Rudin.
\newblock Concept whitening for interpretable image recognition.
\newblock \emph{CoRR}, abs/2002.01650, 2020.

\bibitem[Das and Rad(2020)]{DBLP:journals/corr/abs-2006-11371}
A.~Das and P.~Rad.
\newblock Opportunities and challenges in explainable artificial intelligence
  {(XAI):} {A} survey.
\newblock \emph{CoRR}, abs/2006.11371, 2020.

\bibitem[Dikmen and Burns(2016)]{10.1145/3003715.3005465}
M.~Dikmen and C.~M. Burns.
\newblock Autonomous driving in the real world: Experiences with tesla
  autopilot and summon.
\newblock In \emph{Proceedings of the 8th International Conference on
  Automotive User Interfaces and Interactive Vehicular Applications},
  Automotive'UI 16, page 225–228, New York, NY, USA, 2016. Association for
  Computing Machinery.
\newblock ISBN 9781450345330.

\bibitem[Ge et~al.(2021)Ge, Xiao, Xu, Zheng, Karanam, Chen, Itti, and
  Wu]{DBLP:journals/corr/abs-2105-00290}
Y.~Ge, Y.~Xiao, Z.~Xu, M.~Zheng, S.~Karanam, T.~Chen, L.~Itti, and Z.~Wu.
\newblock A peek into the reasoning of neural networks: Interpreting with
  structural visual concepts.
\newblock \emph{CoRR}, abs/2105.00290, 2021.

\bibitem[Ghorbani et~al.(2019)Ghorbani, Wexler, Zou, and
  Kim]{NEURIPS2019_77d2afcb}
A.~Ghorbani, J.~Wexler, J.~Y. Zou, and B.~Kim.
\newblock Towards automatic concept-based explanations.
\newblock In H.~Wallach, H.~Larochelle, A.~Beygelzimer, F.~d\textquotesingle
  Alch\'{e}-Buc, E.~Fox, and R.~Garnett, editors, \emph{Advances in Neural
  Information Processing Systems}, volume~32. Curran Associates, Inc., 2019.

\bibitem[Goyal et~al.(2019)Goyal, Feder, Shalit, and Kim]{goyal2019explaining}
Y.~Goyal, A.~Feder, U.~Shalit, and B.~Kim.
\newblock Explaining classifiers with causal concept effect (cace).
\newblock \emph{arXiv preprint arXiv:1907.07165}, 2019.

\bibitem[Gu and Cheng(2020)]{DBLP:journals/corr/abs-2010-04974}
X.~Gu and X.~Cheng.
\newblock Distilling a deep neural network into a takagi-sugeno-kang fuzzy
  inference system.
\newblock \emph{CoRR}, abs/2010.04974, 2020.

\bibitem[He et~al.(2015)He, Zhang, Ren, and Sun]{DBLP:journals/corr/HeZRS15}
K.~He, X.~Zhang, S.~Ren, and J.~Sun.
\newblock Deep residual learning for image recognition.
\newblock \emph{CoRR}, abs/1512.03385, 2015.

\bibitem[Huang et~al.(2016)Huang, Liu, and
  Weinberger]{DBLP:journals/corr/HuangLW16a}
G.~Huang, Z.~Liu, and K.~Q. Weinberger.
\newblock Densely connected convolutional networks.
\newblock \emph{CoRR}, abs/1608.06993, 2016.

\bibitem[Kapishnikov et~al.(2019)Kapishnikov, Bolukbasi, Viegas, and
  Terry]{9008576}
A.~Kapishnikov, T.~Bolukbasi, F.~Viegas, and M.~Terry.
\newblock Xrai: Better attributions through regions.
\newblock In \emph{2019 IEEE/CVF International Conference on Computer Vision
  (ICCV)}, pages 4947--4956, 2019.

\bibitem[Kaufman et~al.(2012)Kaufman, Rosset, Perlich, and
  Stitelman]{10.1145/2382577.2382579}
S.~Kaufman, S.~Rosset, C.~Perlich, and O.~Stitelman.
\newblock Leakage in data mining: Formulation, detection, and avoidance.
\newblock \emph{ACM Trans. Knowl. Discov. Data}, 6\penalty0 (4), Dec. 2012.
\newblock ISSN 1556-4681.

\bibitem[Kim et~al.(2018{\natexlab{a}})Kim, Wattenberg, Gilmer, Cai, Wexler,
  Viegas, et~al.]{kim2018interpretability}
B.~Kim, M.~Wattenberg, J.~Gilmer, C.~Cai, J.~Wexler, F.~Viegas, et~al.
\newblock Interpretability beyond feature attribution: Quantitative testing
  with concept activation vectors (tcav).
\newblock In \emph{International conference on machine learning}, pages
  2668--2677. PMLR, 2018{\natexlab{a}}.

\bibitem[Kim et~al.(2018{\natexlab{b}})Kim, Kook, Sun, Kang, and
  Ko]{kim2018parallel}
S.-W. Kim, H.-K. Kook, J.-Y. Sun, M.-C. Kang, and S.-J. Ko.
\newblock Parallel feature pyramid network for object detection.
\newblock In \emph{Proceedings of the European Conference on Computer Vision
  (ECCV)}, pages 234--250, 2018{\natexlab{b}}.

\bibitem[Koh et~al.(2020)Koh, Nguyen, Tang, Mussmann, Pierson, Kim, and
  Liang]{pmlr-v119-koh20a}
P.~W. Koh, T.~Nguyen, Y.~S. Tang, S.~Mussmann, E.~Pierson, B.~Kim, and
  P.~Liang.
\newblock Concept bottleneck models.
\newblock In H.~D. III and A.~Singh, editors, \emph{Proceedings of the 37th
  International Conference on Machine Learning}, volume 119 of
  \emph{Proceedings of Machine Learning Research}. PMLR, 7 2020.

\bibitem[{Landing AI}(2020)]{LandingAI}
{Landing AI}.
\newblock {2020 State of AI-Based Machine Vision High confidence, growing
  adoption and many challenges-trends and insights based on a survey of 110
  companies}.
\newblock Technical report, 2020.
\newblock URL
  \url{https://landing.ai/wp-content/uploads/2020/11/MachineVisionSurvey.pdf}.

\bibitem[Lin et~al.(2017)Lin, Tegmark, and Rolnick]{Lin2017}
H.~W. Lin, M.~Tegmark, and D.~Rolnick.
\newblock Why does deep and cheap learning work so well?
\newblock \emph{Journal of Statistical Physics}, 168\penalty0 (6):\penalty0
  1223--1247, 9 2017.
\newblock ISSN 1572-9613.

\bibitem[Sato and Tsukimoto(2001)]{938448}
M.~Sato and H.~Tsukimoto.
\newblock Rule extraction from neural networks via decision tree induction.
\newblock In \emph{IJCNN'01. International Joint Conference on Neural Networks.
  Proceedings (Cat. No.01CH37222)}, volume~3, pages 1870--1875 vol.3, 2001.

\bibitem[Schaaf et~al.(2019)Schaaf, Huber, and Maucher]{8999213}
N.~Schaaf, M.~Huber, and J.~Maucher.
\newblock Enhancing decision tree based interpretation of deep neural networks
  through l1-orthogonal regularization.
\newblock In \emph{2019 18th IEEE International Conference On Machine Learning
  And Applications (ICMLA)}, pages 42--49, 2019.

\bibitem[Selvaraju et~al.(2017)Selvaraju, Cogswell, Das, Vedantam, Parikh, and
  Batra]{8237336}
R.~R. Selvaraju, M.~Cogswell, A.~Das, R.~Vedantam, D.~Parikh, and D.~Batra.
\newblock Grad-cam: Visual explanations from deep networks via gradient-based
  localization.
\newblock In \emph{2017 IEEE International Conference on Computer Vision
  (ICCV)}, pages 618--626, 2017.

\bibitem[Simonyan and Zisserman(2015)]{DBLP:journals/corr/SimonyanZ14a}
K.~Simonyan and A.~Zisserman.
\newblock Very deep convolutional networks for large-scale image recognition.
\newblock In Y.~Bengio and Y.~LeCun, editors, \emph{3rd International
  Conference on Learning Representations, {ICLR} 2015, San Diego, CA, USA, May
  7-9, 2015, Conference Track Proceedings}, 2015.

\bibitem[Utkin et~al.(2021)Utkin, Drobintsev, Kovalev, and
  Konstantinov]{DBLP:conf/fruct/UtkinDKK21}
L.~V. Utkin, P.~D. Drobintsev, M.~Kovalev, and A.~V. Konstantinov.
\newblock Combining an autoencoder and a variational autoencoder for explaining
  the machine learning model predictions.
\newblock In \emph{28th Conference of Open Innovations Association, {FRUCT}
  2021, Moscow, Russia, January 27-29, 2021}. {IEEE}, 2021.

\bibitem[Van~Noord and Postma(2017)]{van2017learning}
N.~Van~Noord and E.~Postma.
\newblock Learning scale-variant and scale-invariant features for deep image
  classification.
\newblock \emph{Pattern Recognition}, 61:\penalty0 583--592, 2017.

\bibitem[Wang et~al.(2021)Wang, Kong, Guo, and Zhang]{9405672}
P.~Wang, X.~Kong, W.~Guo, and X.~Zhang.
\newblock Exclusive feature constrained class activation mapping for better
  visual explanation.
\newblock \emph{IEEE Access}, 9, 2021.

\bibitem[Xu et~al.(2014)Xu, Xiao, Zhang, Yang, and Zhang]{xu2014scale}
Y.~Xu, T.~Xiao, J.~Zhang, K.~Yang, and Z.~Zhang.
\newblock Scale-invariant convolutional neural networks.
\newblock \emph{arXiv preprint arXiv:1411.6369}, 2014.

\bibitem[Yeh et~al.(2020)Yeh, Kim, Arik, Li, Pfister, and
  Ravikumar]{NEURIPS2020_ecb287ff}
C.-K. Yeh, B.~Kim, S.~Arik, C.-L. Li, T.~Pfister, and P.~Ravikumar.
\newblock On completeness-aware concept-based explanations in deep neural
  networks.
\newblock In H.~Larochelle, M.~Ranzato, R.~Hadsell, M.~F. Balcan, and H.~Lin,
  editors, \emph{Advances in Neural Information Processing Systems}, volume~33.
  Curran Associates, Inc., 2020.

\bibitem[Zhang et~al.(2019)Zhang, Yang, Ma, and Wu]{8953917}
Q.~Zhang, Y.~Yang, H.~Ma, and Y.~N. Wu.
\newblock Interpreting cnns via decision trees.
\newblock In \emph{2019 IEEE/CVF Conference on Computer Vision and Pattern
  Recognition (CVPR)}, pages 6254--6263, 2019.

\bibitem[Zhou et~al.(2003)Zhou, Jiang, and Chen]{Zhou2003}
Z.~Zhou, Y.~Jiang, and S.~Chen.
\newblock Extracting symbolic rules from trained neural network ensembles.
\newblock \emph{AI Communications}, 16:\penalty0 3--15, 2003.

\bibitem[Zilke et~al.(2016)Zilke, Loza~Menc{\'i}a, and
  Janssen]{10.1007/978-3-319-46307-0_29}
J.~R. Zilke, E.~Loza~Menc{\'i}a, and F.~Janssen.
\newblock Deepred -- rule extraction from deep neural networks.
\newblock In T.~Calders, M.~Ceci, and D.~Malerba, editors, \emph{Discovery
  Science}, pages 457--473, Cham, 2016. Springer International Publishing.
\newblock ISBN 978-3-319-46307-0.

\end{thebibliography}

\begin{appendices}
\section{Pseudocode}

This appendix presents a more detailed pseudocode on how SPACE works. The pseudocode is the result of the description in Section \ref{sec:SPACE}, and was used as reference for the implementation of our method.

\newlength{\commentWidth}
\setlength{\commentWidth}{6cm}
\newcommand{\atcp}[1]{\tcp*[r]{\makebox[\commentWidth]{#1\hfill}}}

\begin{algorithm}[H]
    \caption{Scale-Preserving Automatic Concept Extraction, \texttt{SPACE}}
    \SetAlgoLined
    \SetKwInOut{Input}{input}
    \SetKwInOut{Output}{output}
    \DontPrintSemicolon
    \Input{
    $E:\left\{ (x_{1},y_{1}),...,(x_{N},y_{N}) \right\}$,
    $f_\mathrm{M}$,
    $k$,
    $n_\mathrm{s}$,
    $n_\mathrm{p}$,
    $n_\mathrm{pca}$,
    $l_\mathrm{gradcam}$,
    $l_\mathrm{activ}$
    }
    
    \Output{
    Set of concepts $\{ (\varepsilon_{0}, \overline{S}_0),(\varepsilon_{1}, \overline{S}_1), ...,(\varepsilon_{n-1}, \overline{S}_{n-1})\}$
    }
    
    $E_k = \{(x_i, y_i) \in E \mid \mathrm{max\_index}(y_i)=k\}$
    
    $P \gets \emptyset$ \atcp{(A) Patch extraction}
    
	\For{$x_j \:\mathrm{in}\: E_k$}{
		$S_j \gets \mathrm{GradCAM}(x_j, k, f_\mathrm{M}, l_\mathrm{gradcam})$
		
		$P^{*}_{x_j} \gets \mathrm{slice}(x_j, n_\mathrm{s})$
		
		$G^{*} \gets \mathrm{create\_masks}(x_j, n_\mathrm{s})$
		
		$P_{\psi} \gets \emptyset$
		
		\For{$g_o \:\mathrm{in}\: G^{*}$}{
			$\psi_{j,o} \gets f_{\psi}(S_j, g_o)$
			
			$P_{\psi} \gets P_{\psi} \cup \{\psi_{j,o}\}$  
		}
		$P_{x_j} \gets \mathrm{select\_top}(P^{*}_{x_j}, P_{\psi})$
		
		$P \gets P \cup P_{x_j}$
	}
    
	$X^{*} \gets \emptyset$  \atcp{(B) Concept-image composition}
	
	\For{$p_t \:\mathrm{in}\: P$}{
		$x_{p_t} \gets \mathrm{tile}(p_t, n_\mathrm{s})$
		
		$X^{*} \gets X^{*} \cup \{ x_{p_t} \}$
	}
    
	$A^{*} \gets \emptyset$ \atcp{(C) Concept clustering}
	
	\For{$x_{p_t} \:\mathrm{in}\: X^{*}$}{
		$a_{l_\mathrm{activ},x_{p_t}} \gets f_{l_\mathrm{activ}}(x_{p_t})$
		
		$A^{*} \gets A^{*} \cup \{a_{l_\mathrm{activ},x_{p_t}}\}$
	}
	$A^{*}_\mathrm{pca} \gets \mathrm{PCA}(A^{*}, n_\mathrm{pca})$
    
    $C^{*} \gets \mathrm{OPTICS}(A^{*}_\mathrm{pca})$
    
    \For{$c_i \:\mathrm{in}\: C^{*}$}{
    $\varepsilon_{c_i} \gets \mathrm{assemble\_example\_set}(c_i, X^{*})$
    }
    
    $R^{*} \gets \emptyset$ \atcp{(D) Random concept building}
    
    \For{$c_i \:\mathrm{in}\: C^{*}$}{
        $\varepsilon_{r_i} \gets \emptyset$
        
        \For{$x_j \:\mathrm{in}\: E\setminus{E_k}$}{
        $p_t \gets \mathrm{random\_crop}(x_j, n_\mathrm{s})$
        
        $x_{p_t} \gets \mathrm{tile}(p_t, n_\mathrm{s})$
        
        $\varepsilon_{r_i} \gets \varepsilon_{r_i} \cup {x_{p_t}}$ 
        }
        $R^{*} \gets \varepsilon_{r_i}$
    }
    \atcp{(E) Concept testing}
    \For{$c_i \:\mathrm{in}\: C^{*}$}{ 
    $\overline{S}_i \gets \mathrm{TCAV}(\varepsilon_{c_i}, R^{*}, l_\mathrm{activ})$
    }
    
    return ($\left \{(\varepsilon_{c_0}, \overline{S}_0), ..., (\varepsilon_{n_c-1}, \overline{S}_{n_c-1}) \right \}$)
    
    \label{alg:SPACE}
\end{algorithm}
\section{Extended experimental results: concrete cracks dataset}
\label{apd:Extended concrete cracks}

This appendix presents a summary of the performance metrics of the models trained in the concrete cracks dataset, and used in the initial runs. All models where trained until convergence and the sampling was weighted to compensate for the imbalance of the datasets.

As previously described, the concrete crack dataset \citep{conc} consists of two classes, each with \num{20000} concrete images of 227x227 pixels. Class 0 and 1 correspond to good and cracked samples of concrete. The main (and only) concept used for the labeling of the classes is well known, as the cracks span through complete images and there are no other significant features present in the dataset. In this appendix, we present the results of each model trained in this dataset, and the class metrics for the \textbf{Positive} class (which contains cracks) (See Table \ref{table:performance models concrete cracks}).

\begin{table}[ht!]
\caption{Performance metrics of models trained in the concrete cracks dataset.} \label{table:performance models concrete cracks}
\centering
\begin{tabular}{|lll|lll|lll|}

\hline

\rowcolor[HTML]{9B9B9B} 

\multicolumn{3}{|l|}{\cellcolor[HTML]{9B9B9B}}                                                                            & \multicolumn{3}{c|}{\cellcolor[HTML]{9B9B9B}average}                                                                    & \multicolumn{3}{c|}{\cellcolor[HTML]{9B9B9B}class}                                                                      \\ \hline

\rowcolor[HTML]{9B9B9B} 

\multicolumn{1}{|l|}{\cellcolor[HTML]{9B9B9B}CNN} & \multicolumn{1}{l|}{\cellcolor[HTML]{9B9B9B}seed} & acc & \multicolumn{1}{l|}{\cellcolor[HTML]{9B9B9B}precision} & \multicolumn{1}{l|}{\cellcolor[HTML]{9B9B9B}recall} & f1 & \multicolumn{1}{l|}{\cellcolor[HTML]{9B9B9B}precision} & \multicolumn{1}{l|}{\cellcolor[HTML]{9B9B9B}recall} & f1 \\ \hline

\multicolumn{1}{|l|}{vgg16}                                & \multicolumn{1}{l|}{0}                            & 0.999    & \multicolumn{1}{l|}{0.999}                             & \multicolumn{1}{l|}{0.999}                          & 0.999    & \multicolumn{1}{l|}{0.998}                             & \multicolumn{1}{l|}{1.000}                          & 0.999    \\ \hline

\multicolumn{1}{|l|}{vgg16}                                & \multicolumn{1}{l|}{1}                            & 0.999    & \multicolumn{1}{l|}{0.999}                             & \multicolumn{1}{l|}{0.999}                          & 0.999    & \multicolumn{1}{l|}{0.999}                             & \multicolumn{1}{l|}{1.000}                          & 0.999    \\ \hline

\multicolumn{1}{|l|}{vgg16}                                & \multicolumn{1}{l|}{2}                            & 0.999    & \multicolumn{1}{l|}{0.999}                             & \multicolumn{1}{l|}{0.999}                          & 0.999    & \multicolumn{1}{l|}{0.999}                             & \multicolumn{1}{l|}{1.000}                          & 0.999    \\ \hline

\multicolumn{1}{|l|}{vgg16}                                & \multicolumn{1}{l|}{3}                            & 0.999    & \multicolumn{1}{l|}{0.999}                             & \multicolumn{1}{l|}{0.999}                          & 0.999    & \multicolumn{1}{l|}{0.999}                             & \multicolumn{1}{l|}{1.000}                          & 0.999    \\ \hline

\multicolumn{1}{|l|}{vgg16}                                & \multicolumn{1}{l|}{4}                            & 0.999    & \multicolumn{1}{l|}{0.999}                             & \multicolumn{1}{l|}{0.999}                          & 0.999    & \multicolumn{1}{l|}{0.999}                             & \multicolumn{1}{l|}{1.000}                          & 0.999    \\ \hline

\multicolumn{1}{|l|}{vgg16}                                & \multicolumn{1}{l|}{5}                            & 0.999    & \multicolumn{1}{l|}{0.999}                             & \multicolumn{1}{l|}{0.999}                          & 0.999    & \multicolumn{1}{l|}{0.999}                             & \multicolumn{1}{l|}{0.999}                          & 0.999    \\ \hline

\multicolumn{1}{|l|}{vgg16}                                & \multicolumn{1}{l|}{6}                            & 0.999    & \multicolumn{1}{l|}{0.999}                             & \multicolumn{1}{l|}{0.999}                          & 0.999    & \multicolumn{1}{l|}{0.999}                             & \multicolumn{1}{l|}{1.000}                          & 0.999    \\ \hline

\multicolumn{1}{|l|}{vgg16}                                & \multicolumn{1}{l|}{7}                            & 0.999    & \multicolumn{1}{l|}{0.999}                             & \multicolumn{1}{l|}{0.999}                          & 0.999    & \multicolumn{1}{l|}{0.999}                             & \multicolumn{1}{l|}{0.999}                          & 0.999    \\ \hline

\multicolumn{1}{|l|}{vgg16}                                & \multicolumn{1}{l|}{8}                            & 0.999    & \multicolumn{1}{l|}{0.999}                             & \multicolumn{1}{l|}{0.999}                          & 0.999    & \multicolumn{1}{l|}{0.999}                             & \multicolumn{1}{l|}{1.000}                          & 0.999    \\ \hline

\multicolumn{1}{|l|}{vgg16}                                & \multicolumn{1}{l|}{9}                            & 0.999    & \multicolumn{1}{l|}{0.999}                             & \multicolumn{1}{l|}{0.999}                          & 0.999    & \multicolumn{1}{l|}{1.000}                             & \multicolumn{1}{l|}{0.999}                          & 0.999    \\ \hline

\multicolumn{1}{|l|}{resnet18}                             & \multicolumn{1}{l|}{0}                            & 0.999    & \multicolumn{1}{l|}{0.999}                             & \multicolumn{1}{l|}{0.999}                          & 0.999    & \multicolumn{1}{l|}{0.999}                             & \multicolumn{1}{l|}{1.000}                          & 0.999    \\ \hline

\multicolumn{1}{|l|}{resnet18}                             & \multicolumn{1}{l|}{1}                            & 0.999    & \multicolumn{1}{l|}{0.999}                             & \multicolumn{1}{l|}{0.999}                          & 0.999    & \multicolumn{1}{l|}{0.999}                             & \multicolumn{1}{l|}{1.000}                          & 0.999    \\ \hline

\multicolumn{1}{|l|}{resnet18}                             & \multicolumn{1}{l|}{2}                            & 1.000    & \multicolumn{1}{l|}{1.000}                             & \multicolumn{1}{l|}{1.000}                          & 1.000    & \multicolumn{1}{l|}{0.999}                             & \multicolumn{1}{l|}{1.000}                          & 1.000    \\ \hline

\multicolumn{1}{|l|}{resnet18}                             & \multicolumn{1}{l|}{3}                            & 0.999    & \multicolumn{1}{l|}{0.999}                             & \multicolumn{1}{l|}{0.999}                          & 0.999    & \multicolumn{1}{l|}{0.999}                             & \multicolumn{1}{l|}{1.000}                          & 0.999    \\ \hline

\multicolumn{1}{|l|}{resnet18}                             & \multicolumn{1}{l|}{4}                            & 0.999    & \multicolumn{1}{l|}{0.999}                             & \multicolumn{1}{l|}{0.999}                          & 0.999    & \multicolumn{1}{l|}{0.999}                             & \multicolumn{1}{l|}{1.000}                          & 0.999    \\ \hline

\multicolumn{1}{|l|}{resnet18}                             & \multicolumn{1}{l|}{5}                            & 0.999    & \multicolumn{1}{l|}{0.999}                             & \multicolumn{1}{l|}{0.999}                          & 0.999    & \multicolumn{1}{l|}{0.999}                             & \multicolumn{1}{l|}{1.000}                          & 0.999    \\ \hline

\multicolumn{1}{|l|}{resnet18}                             & \multicolumn{1}{l|}{6}                            & 0.999    & \multicolumn{1}{l|}{0.999}                             & \multicolumn{1}{l|}{0.999}                          & 0.999    & \multicolumn{1}{l|}{0.999}                             & \multicolumn{1}{l|}{1.000}                          & 0.999    \\ \hline

\multicolumn{1}{|l|}{resnet18}                             & \multicolumn{1}{l|}{7}                            & 0.999    & \multicolumn{1}{l|}{0.999}                             & \multicolumn{1}{l|}{0.999}                          & 0.999    & \multicolumn{1}{l|}{0.999}                             & \multicolumn{1}{l|}{1.000}                          & 0.999    \\ \hline

\multicolumn{1}{|l|}{resnet18}                             & \multicolumn{1}{l|}{8}                            & 0.999    & \multicolumn{1}{l|}{0.999}                             & \multicolumn{1}{l|}{0.999}                          & 0.999    & \multicolumn{1}{l|}{0.999}                             & \multicolumn{1}{l|}{1.000}                          & 0.999    \\ \hline

\multicolumn{1}{|l|}{resnet18}                             & \multicolumn{1}{l|}{9}                            & 1.000    & \multicolumn{1}{l|}{1.000}                             & \multicolumn{1}{l|}{1.000}                          & 0.999    & \multicolumn{1}{l|}{0.999}                             & \multicolumn{1}{l|}{1.000}                          & 1.000    \\ \hline

\multicolumn{1}{|l|}{densenet121}                          & \multicolumn{1}{l|}{0}                            & 0.999    & \multicolumn{1}{l|}{0.999}                             & \multicolumn{1}{l|}{0.999}                          & 0.999    & \multicolumn{1}{l|}{0.999}                             & \multicolumn{1}{l|}{1.000}                          & 0.999    \\ \hline

\multicolumn{1}{|l|}{densenet121}                          & \multicolumn{1}{l|}{1}                            & 0.999    & \multicolumn{1}{l|}{0.999}                             & \multicolumn{1}{l|}{0.999}                          & 0.999    & \multicolumn{1}{l|}{0.999}                             & \multicolumn{1}{l|}{1.000}                          & 0.999    \\ \hline

\multicolumn{1}{|l|}{densenet121}                          & \multicolumn{1}{l|}{2}                            & 0.999    & \multicolumn{1}{l|}{0.999}                             & \multicolumn{1}{l|}{0.999}                          & 0.999    & \multicolumn{1}{l|}{0.999}                             & \multicolumn{1}{l|}{1.000}                          & 0.999    \\ \hline

\multicolumn{1}{|l|}{densenet121}                          & \multicolumn{1}{l|}{3}                            & 0.999    & \multicolumn{1}{l|}{0.999}                             & \multicolumn{1}{l|}{0.999}                          & 0.999    & \multicolumn{1}{l|}{0.999}                             & \multicolumn{1}{l|}{1.000}                          & 0.999    \\ \hline

\multicolumn{1}{|l|}{densenet121}                          & \multicolumn{1}{l|}{4}                            & 0.999    & \multicolumn{1}{l|}{0.999}                             & \multicolumn{1}{l|}{0.999}                          & 0.999    & \multicolumn{1}{l|}{0.999}                             & \multicolumn{1}{l|}{1.000}                          & 0.999    \\ \hline

\multicolumn{1}{|l|}{densenet121}                          & \multicolumn{1}{l|}{5}                            & 0.999    & \multicolumn{1}{l|}{0.999}                             & \multicolumn{1}{l|}{0.999}                          & 0.999    & \multicolumn{1}{l|}{0.999}                             & \multicolumn{1}{l|}{1.000}                          & 0.999    \\ \hline

\multicolumn{1}{|l|}{densenet121}                          & \multicolumn{1}{l|}{6}                            & 0.999    & \multicolumn{1}{l|}{0.999}                             & \multicolumn{1}{l|}{0.999}                          & 0.999    & \multicolumn{1}{l|}{0.999}                             & \multicolumn{1}{l|}{1.000}                          & 0.999    \\ \hline

\multicolumn{1}{|l|}{densenet121}                          & \multicolumn{1}{l|}{7}                            & 0.999    & \multicolumn{1}{l|}{0.999}                             & \multicolumn{1}{l|}{0.999}                          & 0.999    & \multicolumn{1}{l|}{0.999}                             & \multicolumn{1}{l|}{1.000}                          & 0.999    \\ \hline

\multicolumn{1}{|l|}{densenet121}                          & \multicolumn{1}{l|}{8}                            & 0.999    & \multicolumn{1}{l|}{0.999}                             & \multicolumn{1}{l|}{0.999}                          & 0.999    & \multicolumn{1}{l|}{0.999}                             & \multicolumn{1}{l|}{1.000}                          & 0.999    \\ \hline

\multicolumn{1}{|l|}{densenet121}                          & \multicolumn{1}{l|}{9}                            & 1.000    & \multicolumn{1}{l|}{1.000}                             & \multicolumn{1}{l|}{1.000}                          & 1.000    & \multicolumn{1}{l|}{1.000}                             & \multicolumn{1}{l|}{1.000}                          & 1.000    \\ \hline

\end{tabular}
\end{table}

Similarly, we provide the results of analyzing every SPACE and ACE run, and aggregating them based on the alignment of the extracted concepts (See Table \ref{table:concrete cracks}).

\begin{table}[ht!]
\caption{Experimental results concrete crack dataset, class positive.} \label{table:concrete cracks}
\centering
\begin{tabular}{|l|l|l|l|l|}

\hline

\rowcolor[HTML]{C0C0C0} 

model        & parameters                            & concepts & aligned & aligned ratio \\ \hline

DenseNet-121 & ACE: $n_{\mathrm{SLIC}}=[15, 50, 80]$ & 14.0     & 6.3     & 0.45          \\ \hline

DenseNet-121 & ACE: $n_{\mathrm{SLIC}}=[15]$         & 17.4     & 4.1     & 0.24          \\ \hline

DenseNet-121 & ACE: $n_{\mathrm{SLIC}}=[200]$        & 19.5     & 2.7     & 0.14          \\ \hline

DenseNet-121 & ACE: $n_{\mathrm{SLIC}}=[50]$         & 15.2     & 4.9     & 0.33          \\ \hline

DenseNet-121 & ACE: $n_{\mathrm{SLIC}}=[80]$         & 14.4     & 4.0     & 0.29          \\ \hline

DenseNet-121 & SPACE: $n_{\mathrm{s}}=4$             & 3.0      & 3.0     & 1.00          \\ \hline

DenseNet-121 & SPACE: $n_{\mathrm{s}}=5$             & 6.8      & 6.4     & 0.94          \\ \hline

DenseNet-121 & SPACE: $n_{\mathrm{s}}=6$             & 6.0      & 6.0     & 1.00          \\ \hline

DenseNet-121 & SPACE: $n_{\mathrm{s}}=7$             & 6.8      & 6.7     & 0.99          \\ \hline

DenseNet-121 & SPACE: $n_{\mathrm{s}}=8$             & 12.4     & 11.8    & 0.96          \\ \hline

ResNet-18    & ACE: $n_{\mathrm{SLIC}}=[15, 50, 80]$ & 13.4     & 8.3     & 0.63          \\ \hline

ResNet-18    & ACE: $n_{\mathrm{SLIC}}=[15]$         & 13.3     & 7.8     & 0.59          \\ \hline

ResNet-18    & ACE: $n_{\mathrm{SLIC}}=[200]$        & 18.8     & 2.6     & 0.14          \\ \hline

ResNet-18    & ACE: $n_{\mathrm{SLIC}}=[50]$         & 13.8     & 7.0     & 0.52          \\ \hline

ResNet-18    & ACE: $n_{\mathrm{SLIC}}=[80]$         & 12.6     & 5.3     & 0.42          \\ \hline

ResNet-18    & SPACE: $n_{\mathrm{s}}=4$             & 3.9      & 3.6     & 0.86          \\ \hline

ResNet-18    & SPACE: $n_{\mathrm{s}}=5$             & 6.6      & 6.1     & 0.89          \\ \hline

ResNet-18    & SPACE: $n_{\mathrm{s}}=6$             & 7.9      & 7.1     & 0.89          \\ \hline

ResNet-18    & SPACE: $n_{\mathrm{s}}=7$             & 7.3      & 6.4     & 0.89          \\ \hline

ResNet-18    & SPACE: $n_{\mathrm{s}}=8$             & 13.0     & 11.5    & 0.88          \\ \hline

VGG-16       & ACE: $n_{\mathrm{SLIC}}=[15, 50, 80]$ & 15.0     & 12.4    & 0.83          \\ \hline

VGG-16       & ACE: $n_{\mathrm{SLIC}}=[15]$         & 13.1     & 10.4    & 0.79          \\ \hline

VGG-16       & ACE: $n_{\mathrm{SLIC}}=[200]$        & 17.9     & 4.2     & 0.23          \\ \hline

VGG-16       & ACE: $n_{\mathrm{SLIC}}=[50]$         & 14.7     & 11.4    & 0.78          \\ \hline

VGG-16       & ACE: $n_{\mathrm{SLIC}}=[80]$         & 16.5     & 9.8     & 0.60          \\ \hline

VGG-16       & SPACE: $n_{\mathrm{s}}=4$             & 5.4      & 5.1     & 0.92          \\ \hline

VGG-16       & SPACE: $n_{\mathrm{s}}=5$             & 10.6     & 10.1    & 0.95          \\ \hline

VGG-16       & SPACE: $n_{\mathrm{s}}=6$             & 11.0     & 10.4    & 0.94          \\ \hline

VGG-16       & SPACE: $n_{\mathrm{s}}=7$             & 9.6      & 8.7     & 0.90          \\ \hline

VGG-16       & SPACE: $n_{\mathrm{s}}=8$             & 16.9     & 15.0    & 0.87          \\ \hline

\end{tabular}
\end{table}

\clearpage
\section{Extended experimental results: metal nut dataset}
\label{apd:Extended metal nut}

This appendix presents a summary of the performance metrics of the models trained in the metal nut dataset, and used in the initial runs. All models where trained until convergence and the sampling was weighted to compensate for the imbalance of the datasets.

The MVTec metal nuts dataset \citep{8954181} is originally an anomaly detection dataset, which was reframed as a classification task. Each image is of size 700x700 pixels and contains a metal nut in front of a black background. Classes 0 to 4 are bent, color, flipped, ok, and scratched metal nuts, respectively. The numbers of images for each of the classes are imbalanced (e.g., class 3 has 242 images, whereas class 1 has only 22 images in total), which is why resampling and data augmentation were used for the training data.
In this appendix, we present the results of each model trained in this dataset, and the class metrics for the \textbf{color} class (which contain metal nuts stained with red, blue, or black marks) (See Table \ref{table:performance models metal nut}).

\begin{table}[ht!]
\caption{Performance metrics of models trained in the metal nut dataset.} \label{table:performance models metal nut}
\centering
\begin{tabular}{|lll|lll|lll|}

\hline

\rowcolor[HTML]{9B9B9B} 

\multicolumn{3}{|l|}{\cellcolor[HTML]{9B9B9B}}                                                                            & \multicolumn{3}{c|}{\cellcolor[HTML]{9B9B9B}average}                                                                    & \multicolumn{3}{c|}{\cellcolor[HTML]{9B9B9B}class}                                                                      \\ \hline

\rowcolor[HTML]{9B9B9B} 

\multicolumn{1}{|l|}{\cellcolor[HTML]{9B9B9B}CNN} & \multicolumn{1}{l|}{\cellcolor[HTML]{9B9B9B}seed} & acc & \multicolumn{1}{l|}{\cellcolor[HTML]{9B9B9B}precision} & \multicolumn{1}{l|}{\cellcolor[HTML]{9B9B9B}recall} & f1 & \multicolumn{1}{l|}{\cellcolor[HTML]{9B9B9B}precision} & \multicolumn{1}{l|}{\cellcolor[HTML]{9B9B9B}recall} & f1 \\ \hline

\multicolumn{1}{|l|}{vgg16}                                & \multicolumn{1}{l|}{0}                            & 1.000    & \multicolumn{1}{l|}{1.000}                             & \multicolumn{1}{l|}{1.000}                          & 1.000    & \multicolumn{1}{l|}{1.000}                             & \multicolumn{1}{l|}{1.000}                          & 1.000    \\ \hline

\multicolumn{1}{|l|}{vgg16}                                & \multicolumn{1}{l|}{1}                            & 1.000    & \multicolumn{1}{l|}{1.000}                             & \multicolumn{1}{l|}{1.000}                          & 1.000    & \multicolumn{1}{l|}{1.000}                             & \multicolumn{1}{l|}{1.000}                          & 1.000    \\ \hline

\multicolumn{1}{|l|}{vgg16}                                & \multicolumn{1}{l|}{2}                            & 1.000    & \multicolumn{1}{l|}{1.000}                             & \multicolumn{1}{l|}{1.000}                          & 1.000    & \multicolumn{1}{l|}{1.000}                             & \multicolumn{1}{l|}{1.000}                          & 1.000    \\ \hline

\multicolumn{1}{|l|}{vgg16}                                & \multicolumn{1}{l|}{3}                            & 1.000    & \multicolumn{1}{l|}{1.000}                             & \multicolumn{1}{l|}{1.000}                          & 1.000    & \multicolumn{1}{l|}{1.000}                             & \multicolumn{1}{l|}{1.000}                          & 1.000    \\ \hline

\multicolumn{1}{|l|}{vgg16}                                & \multicolumn{1}{l|}{4}                            & 1.000    & \multicolumn{1}{l|}{1.000}                             & \multicolumn{1}{l|}{1.000}                          & 1.000    & \multicolumn{1}{l|}{1.000}                             & \multicolumn{1}{l|}{1.000}                          & 1.000    \\ \hline

\multicolumn{1}{|l|}{vgg16}                                & \multicolumn{1}{l|}{5}                            & 1.000    & \multicolumn{1}{l|}{1.000}                             & \multicolumn{1}{l|}{1.000}                          & 1.000    & \multicolumn{1}{l|}{1.000}                             & \multicolumn{1}{l|}{1.000}                          & 1.000    \\ \hline

\multicolumn{1}{|l|}{vgg16}                                & \multicolumn{1}{l|}{6}                            & 1.000    & \multicolumn{1}{l|}{1.000}                             & \multicolumn{1}{l|}{1.000}                          & 1.000    & \multicolumn{1}{l|}{1.000}                             & \multicolumn{1}{l|}{1.000}                          & 1.000    \\ \hline

\multicolumn{1}{|l|}{vgg16}                                & \multicolumn{1}{l|}{7}                            & 1.000    & \multicolumn{1}{l|}{1.000}                             & \multicolumn{1}{l|}{1.000}                          & 1.000    & \multicolumn{1}{l|}{1.000}                             & \multicolumn{1}{l|}{1.000}                          & 1.000    \\ \hline

\multicolumn{1}{|l|}{vgg16}                                & \multicolumn{1}{l|}{8}                            & 1.000    & \multicolumn{1}{l|}{1.000}                             & \multicolumn{1}{l|}{1.000}                          & 1.000    & \multicolumn{1}{l|}{1.000}                             & \multicolumn{1}{l|}{1.000}                          & 1.000    \\ \hline

\multicolumn{1}{|l|}{vgg16}                                & \multicolumn{1}{l|}{9}                            & 0.991    & \multicolumn{1}{l|}{0.992}                             & \multicolumn{1}{l|}{0.991}                          & 0.991    & \multicolumn{1}{l|}{1.000}                             & \multicolumn{1}{l|}{0.955}                          & 0.977    \\ \hline

\multicolumn{1}{|l|}{resnet18}                             & \multicolumn{1}{l|}{0}                            & 1.000    & \multicolumn{1}{l|}{1.000}                             & \multicolumn{1}{l|}{1.000}                          & 1.000    & \multicolumn{1}{l|}{1.000}                             & \multicolumn{1}{l|}{1.000}                          & 1.000    \\ \hline

\multicolumn{1}{|l|}{resnet18}                             & \multicolumn{1}{l|}{1}                            & 1.000    & \multicolumn{1}{l|}{1.000}                             & \multicolumn{1}{l|}{1.000}                          & 1.000    & \multicolumn{1}{l|}{1.000}                             & \multicolumn{1}{l|}{1.000}                          & 1.000    \\ \hline

\multicolumn{1}{|l|}{resnet18}                             & \multicolumn{1}{l|}{2}                            & 1.000    & \multicolumn{1}{l|}{1.000}                             & \multicolumn{1}{l|}{1.000}                          & 1.000    & \multicolumn{1}{l|}{1.000}                             & \multicolumn{1}{l|}{1.000}                          & 1.000    \\ \hline

\multicolumn{1}{|l|}{resnet18}                             & \multicolumn{1}{l|}{3}                            & 1.000    & \multicolumn{1}{l|}{1.000}                             & \multicolumn{1}{l|}{1.000}                          & 1.000    & \multicolumn{1}{l|}{1.000}                             & \multicolumn{1}{l|}{1.000}                          & 1.000    \\ \hline

\multicolumn{1}{|l|}{resnet18}                             & \multicolumn{1}{l|}{4}                            & 1.000    & \multicolumn{1}{l|}{1.000}                             & \multicolumn{1}{l|}{1.000}                          & 1.000    & \multicolumn{1}{l|}{1.000}                             & \multicolumn{1}{l|}{1.000}                          & 1.000    \\ \hline

\multicolumn{1}{|l|}{resnet18}                             & \multicolumn{1}{l|}{5}                            & 1.000    & \multicolumn{1}{l|}{1.000}                             & \multicolumn{1}{l|}{1.000}                          & 1.000    & \multicolumn{1}{l|}{1.000}                             & \multicolumn{1}{l|}{1.000}                          & 1.000    \\ \hline

\multicolumn{1}{|l|}{resnet18}                             & \multicolumn{1}{l|}{6}                            & 1.000    & \multicolumn{1}{l|}{1.000}                             & \multicolumn{1}{l|}{1.000}                          & 1.000    & \multicolumn{1}{l|}{1.000}                             & \multicolumn{1}{l|}{1.000}                          & 1.000    \\ \hline

\multicolumn{1}{|l|}{resnet18}                             & \multicolumn{1}{l|}{7}                            & 1.000    & \multicolumn{1}{l|}{1.000}                             & \multicolumn{1}{l|}{1.000}                          & 1.000    & \multicolumn{1}{l|}{1.000}                             & \multicolumn{1}{l|}{1.000}                          & 1.000    \\ \hline

\multicolumn{1}{|l|}{resnet18}                             & \multicolumn{1}{l|}{8}                            & 1.000    & \multicolumn{1}{l|}{1.000}                             & \multicolumn{1}{l|}{1.000}                          & 1.000    & \multicolumn{1}{l|}{1.000}                             & \multicolumn{1}{l|}{1.000}                          & 1.000    \\ \hline

\multicolumn{1}{|l|}{resnet18}                             & \multicolumn{1}{l|}{9}                            & 1.000    & \multicolumn{1}{l|}{1.000}                             & \multicolumn{1}{l|}{1.000}                          & 1.000    & \multicolumn{1}{l|}{1.000}                             & \multicolumn{1}{l|}{1.000}                          & 1.000    \\ \hline

\multicolumn{1}{|l|}{densenet121}                          & \multicolumn{1}{l|}{0}                            & 1.000    & \multicolumn{1}{l|}{1.000}                             & \multicolumn{1}{l|}{1.000}                          & 1.000    & \multicolumn{1}{l|}{1.000}                             & \multicolumn{1}{l|}{1.000}                          & 1.000    \\ \hline

\multicolumn{1}{|l|}{densenet121}                          & \multicolumn{1}{l|}{1}                            & 1.000    & \multicolumn{1}{l|}{1.000}                             & \multicolumn{1}{l|}{1.000}                          & 1.000    & \multicolumn{1}{l|}{1.000}                             & \multicolumn{1}{l|}{1.000}                          & 1.000    \\ \hline

\multicolumn{1}{|l|}{densenet121}                          & \multicolumn{1}{l|}{2}                            & 1.000    & \multicolumn{1}{l|}{1.000}                             & \multicolumn{1}{l|}{1.000}                          & 1.000    & \multicolumn{1}{l|}{1.000}                             & \multicolumn{1}{l|}{1.000}                          & 1.000    \\ \hline

\multicolumn{1}{|l|}{densenet121}                          & \multicolumn{1}{l|}{3}                            & 1.000    & \multicolumn{1}{l|}{1.000}                             & \multicolumn{1}{l|}{1.000}                          & 1.000    & \multicolumn{1}{l|}{1.000}                             & \multicolumn{1}{l|}{1.000}                          & 1.000    \\ \hline

\multicolumn{1}{|l|}{densenet121}                          & \multicolumn{1}{l|}{4}                            & 1.000    & \multicolumn{1}{l|}{1.000}                             & \multicolumn{1}{l|}{1.000}                          & 1.000    & \multicolumn{1}{l|}{1.000}                             & \multicolumn{1}{l|}{1.000}                          & 1.000    \\ \hline

\multicolumn{1}{|l|}{densenet121}                          & \multicolumn{1}{l|}{5}                            & 0.991    & \multicolumn{1}{l|}{0.992}                             & \multicolumn{1}{l|}{0.991}                          & 0.991    & \multicolumn{1}{l|}{1.000}                             & \multicolumn{1}{l|}{1.000}                          & 1.000    \\ \hline

\multicolumn{1}{|l|}{densenet121}                          & \multicolumn{1}{l|}{6}                            & 1.000    & \multicolumn{1}{l|}{1.000}                             & \multicolumn{1}{l|}{1.000}                          & 1.000    & \multicolumn{1}{l|}{1.000}                             & \multicolumn{1}{l|}{1.000}                          & 1.000    \\ \hline

\multicolumn{1}{|l|}{densenet121}                          & \multicolumn{1}{l|}{7}                            & 1.000    & \multicolumn{1}{l|}{1.000}                             & \multicolumn{1}{l|}{1.000}                          & 1.000    & \multicolumn{1}{l|}{1.000}                             & \multicolumn{1}{l|}{1.000}                          & 1.000    \\ \hline

\multicolumn{1}{|l|}{densenet121}                          & \multicolumn{1}{l|}{8}                            & 1.000    & \multicolumn{1}{l|}{1.000}                             & \multicolumn{1}{l|}{1.000}                          & 1.000    & \multicolumn{1}{l|}{1.000}                             & \multicolumn{1}{l|}{1.000}                          & 1.000    \\ \hline

\multicolumn{1}{|l|}{densenet121}                          & \multicolumn{1}{l|}{9}                            & 1.000    & \multicolumn{1}{l|}{1.000}                             & \multicolumn{1}{l|}{1.000}                          & 1.000    & \multicolumn{1}{l|}{1.000}                             & \multicolumn{1}{l|}{1.000}                          & 1.000    \\ \hline

\end{tabular}
\end{table}

Similarly, we provide the results of analyzing every SPACE and ACE run, and aggregating them based on the alignment of the extracted concepts (See Table \ref{table:metal nut}).

\begin{table}[ht!]
\caption{Experimental results metal nut dataset, class color.} \label{table:metal nut}
\centering
\begin{tabular}{|l|l|l|l|l|}

\hline

\rowcolor[HTML]{C0C0C0} 

model        & parameters                            & concepts & aligned & aligned ratio \\ \hline

DenseNet-121 & ACE: $n_{\mathrm{SLIC}}=[15, 50, 80]$ & 10.3     & 3.7     & 0.36          \\ \hline

DenseNet-121 & ACE: $n_{\mathrm{SLIC}}=[15]$         & 4.5      & 0.1     & 0.02          \\ \hline

DenseNet-121 & ACE: $n_{\mathrm{SLIC}}=[200]$        & 19.4     & 4.0     & 0.20          \\ \hline

DenseNet-121 & ACE: $n_{\mathrm{SLIC}}=[50]$         & 10.8     & 1.6     & 0.15          \\ \hline

DenseNet-121 & ACE: $n_{\mathrm{SLIC}}=[80]$         & 13.9     & 1.7     & 0.12          \\ \hline

DenseNet-121 & SPACE: $n_{\mathrm{s}}=5$             & 1.0      & 1.0     & 1.00          \\ \hline

DenseNet-121 & SPACE: $n_{\mathrm{s}}=6$             & 1.4      & 1.3     & 0.95          \\ \hline

DenseNet-121 & SPACE: $n_{\mathrm{s}}=7$             & 1.4      & 1.2     & 0.92          \\ \hline

DenseNet-121 & SPACE: $n_{\mathrm{s}}=8$             & 2.0      & 0.9     & 0.43          \\ \hline

ResNet-18    & ACE: $n_{\mathrm{SLIC}}=[15, 50, 80]$ & 10.6     & 2.4     & 0.23          \\ \hline

ResNet-18    & ACE: $n_{\mathrm{SLIC}}=[15]$         & 3.9      & 0.6     & 0.15          \\ \hline

ResNet-18    & ACE: $n_{\mathrm{SLIC}}=[200]$        & 14.4     & 1.7     & 0.12          \\ \hline

ResNet-18    & ACE: $n_{\mathrm{SLIC}}=[50]$         & 10.5     & 1.2     & 0.12          \\ \hline

ResNet-18    & ACE: $n_{\mathrm{SLIC}}=[80]$         & 13.7     & 2.0     & 0.15          \\ \hline

ResNet-18    & SPACE: $n_{\mathrm{s}}=5$             & 1.0      & 1.0     & 1.00          \\ \hline

ResNet-18    & SPACE: $n_{\mathrm{s}}=6$             & 1.0      & 0.9     & 0.90          \\ \hline

ResNet-18    & SPACE: $n_{\mathrm{s}}=7$             & 1.0      & 0.8     & 0.80          \\ \hline

ResNet-18    & SPACE: $n_{\mathrm{s}}=8$             & 1.2      & 0.9     & 0.70          \\ \hline

VGG-16       & ACE: $n_{\mathrm{SLIC}}=[15, 50, 80]$ & 10.3     & 3.8     & 0.37          \\ \hline

VGG-16       & ACE: $n_{\mathrm{SLIC}}=[15]$         & 4.5      & 0.5     & 0.11          \\ \hline

VGG-16       & ACE: $n_{\mathrm{SLIC}}=[200]$        & 13.8     & 1.5     & 0.12          \\ \hline

VGG-16       & ACE: $n_{\mathrm{SLIC}}=[50]$         & 11.2     & 1.7     & 0.16          \\ \hline

VGG-16       & ACE: $n_{\mathrm{SLIC}}=[80]$         & 13.4     & 1.9     & 0.14          \\ \hline

VGG-16       & SPACE: $n_{\mathrm{s}}=5$             & 1.0      & 1.0     & 1.00          \\ \hline

VGG-16       & SPACE: $n_{\mathrm{s}}=6$             & 1.1      & 0.9     & 0.85          \\ \hline

VGG-16       & SPACE: $n_{\mathrm{s}}=7$             & 1.5      & 0.7     & 0.57          \\ \hline

VGG-16       & SPACE: $n_{\mathrm{s}}=8$             & 1.5      & 1.0     & 0.65          \\ \hline

\end{tabular}
\end{table}

\clearpage
\section{Extended experimental results: leather dataset}
\label{apd:Extended leather}

\begin{figure}[h]

\centering
\begin{minipage}{1.0\textwidth}
    \centering
    \begin{subfigure}[t]{0.98\linewidth}
        \centering
        \begin{subfigure}[t]{0.16\linewidth}
            \centering
            \includegraphics[width=1.0\linewidth]{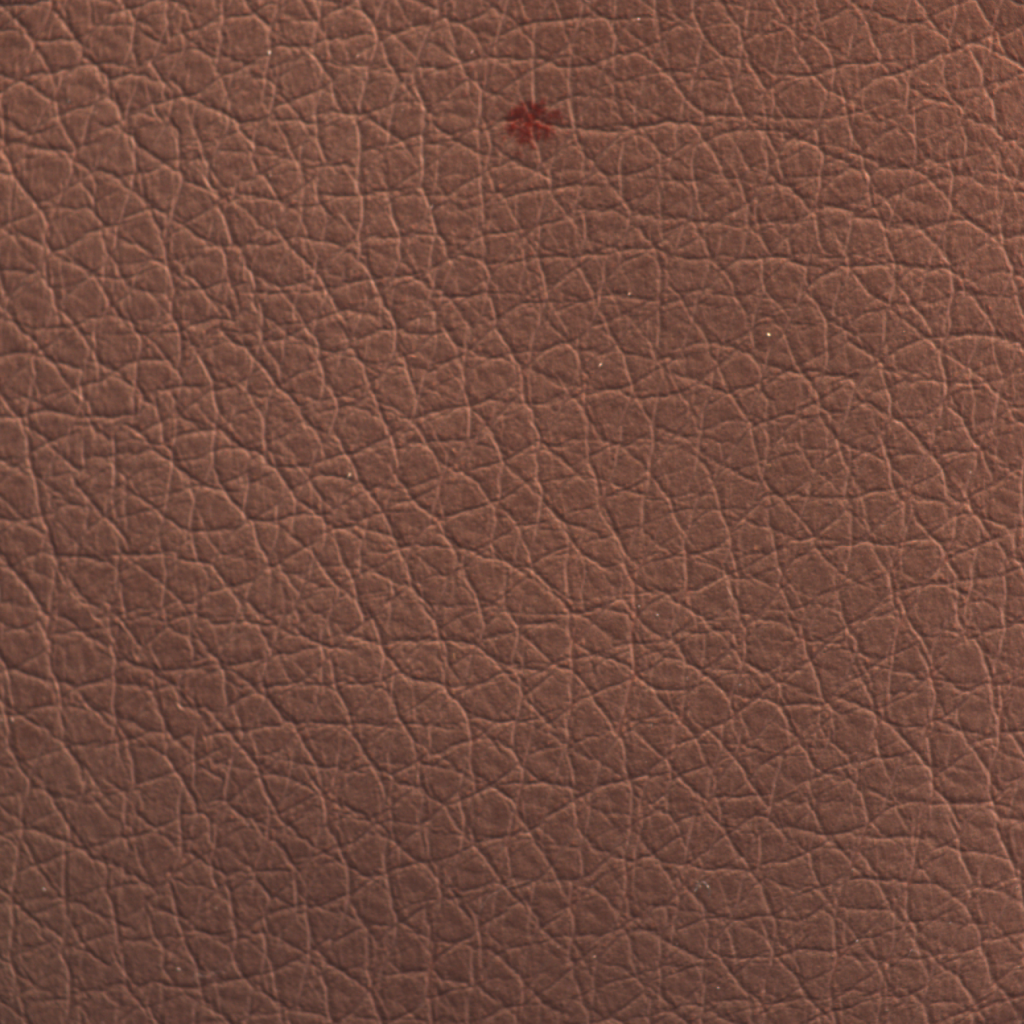}
            \caption{color}\label{fig:leather-color}
        \end{subfigure}
        \begin{subfigure}[t]{0.16\linewidth}
            \centering
            \includegraphics[width=1.0\linewidth]{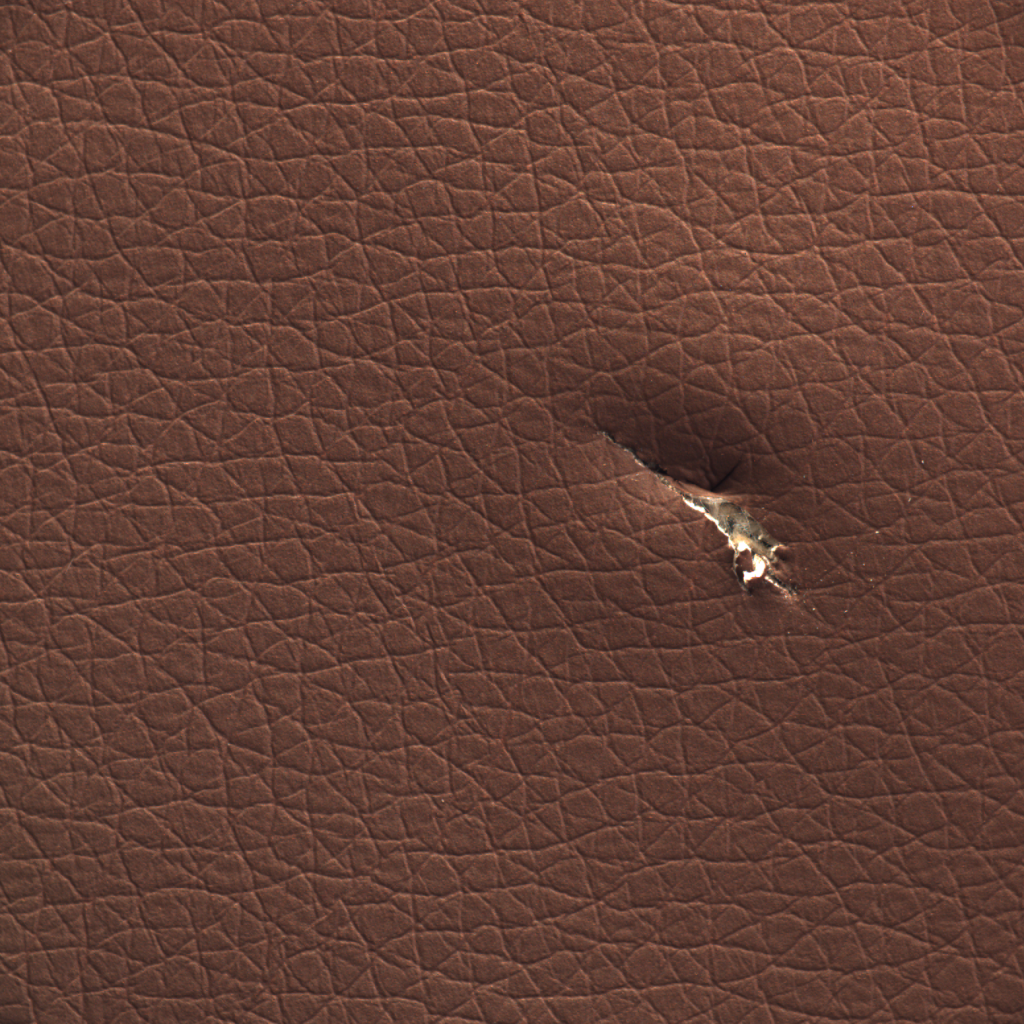}
            \caption{cut}\label{fig:leather-cut}
        \end{subfigure}
        \begin{subfigure}[t]{0.16\linewidth}
            \centering
            \includegraphics[width=1.0\linewidth]{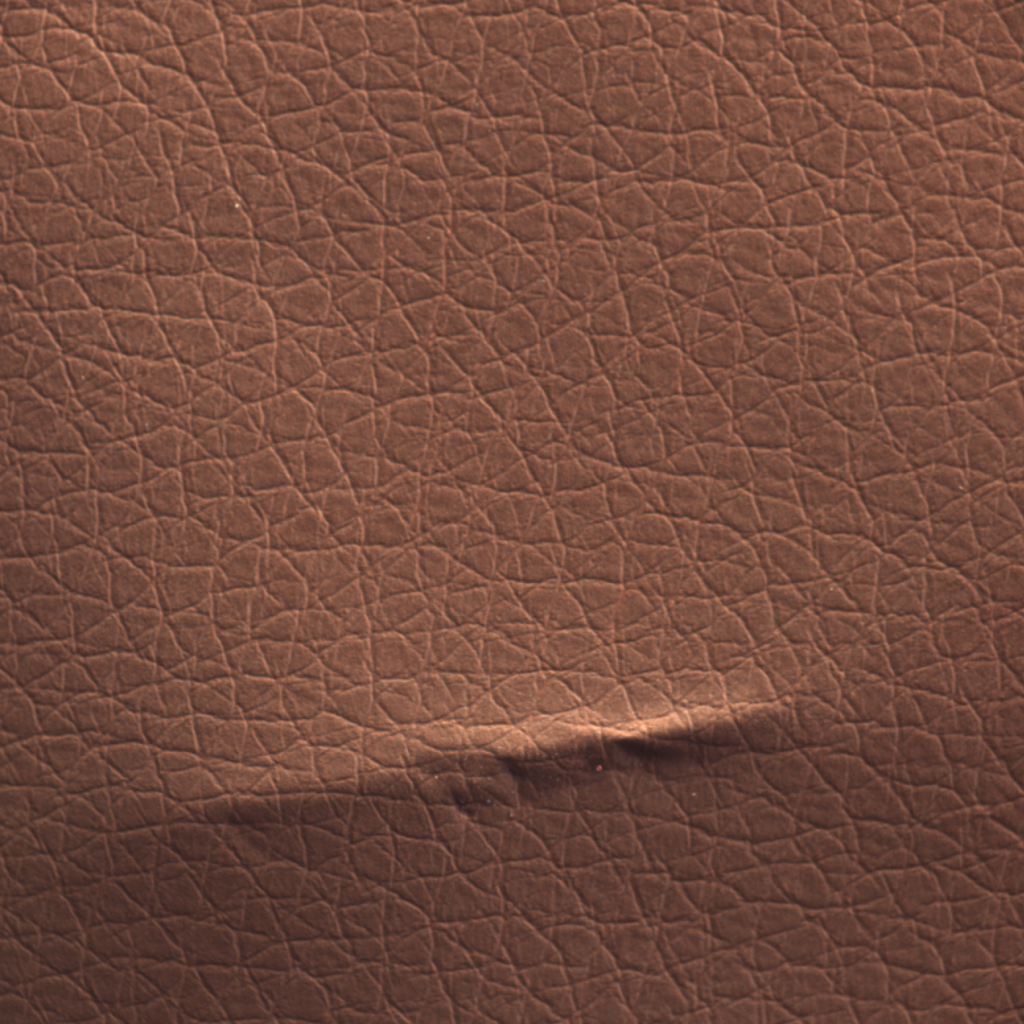}
            \caption{fold}\label{fig:leather-fold}
        \end{subfigure}
        \begin{subfigure}[t]{0.16\linewidth}
            \centering
            \includegraphics[width=1.0\linewidth]{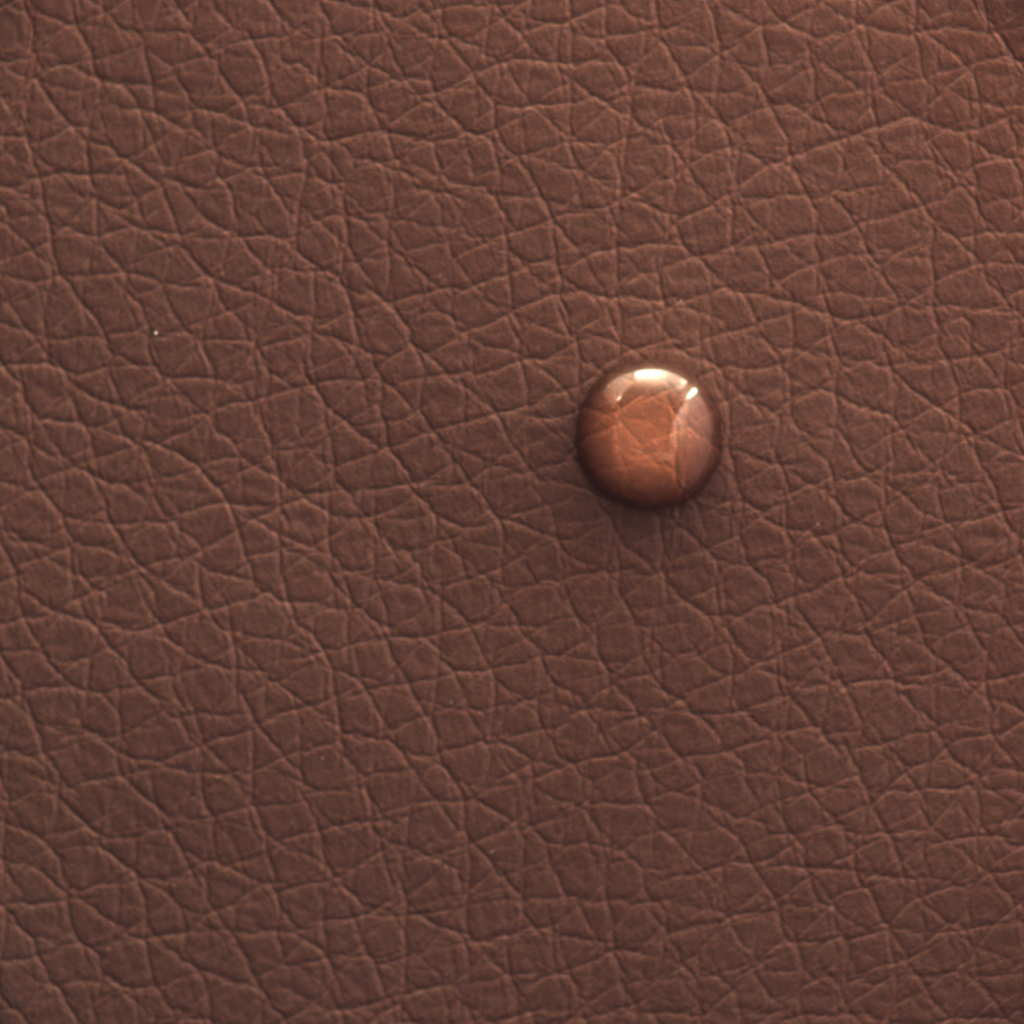}
            \caption{glue}\label{fig:leather-glue}
        \end{subfigure}
        \begin{subfigure}[t]{0.16\linewidth}
            \centering
            \includegraphics[width=1.0\linewidth]{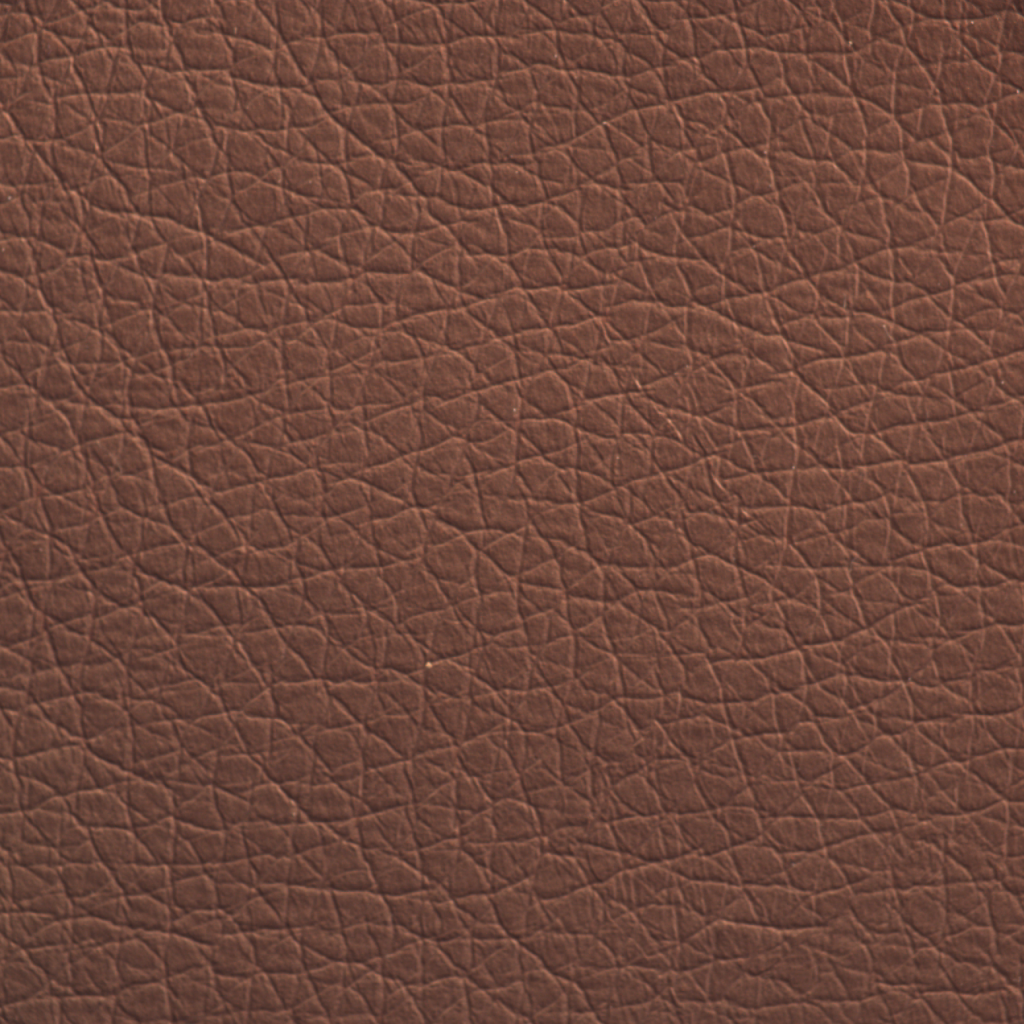}
            \caption{ok}\label{fig:leather-ok}
        \end{subfigure}
        \begin{subfigure}[t]{0.16\linewidth}
            \centering
            \includegraphics[width=1.0\linewidth]{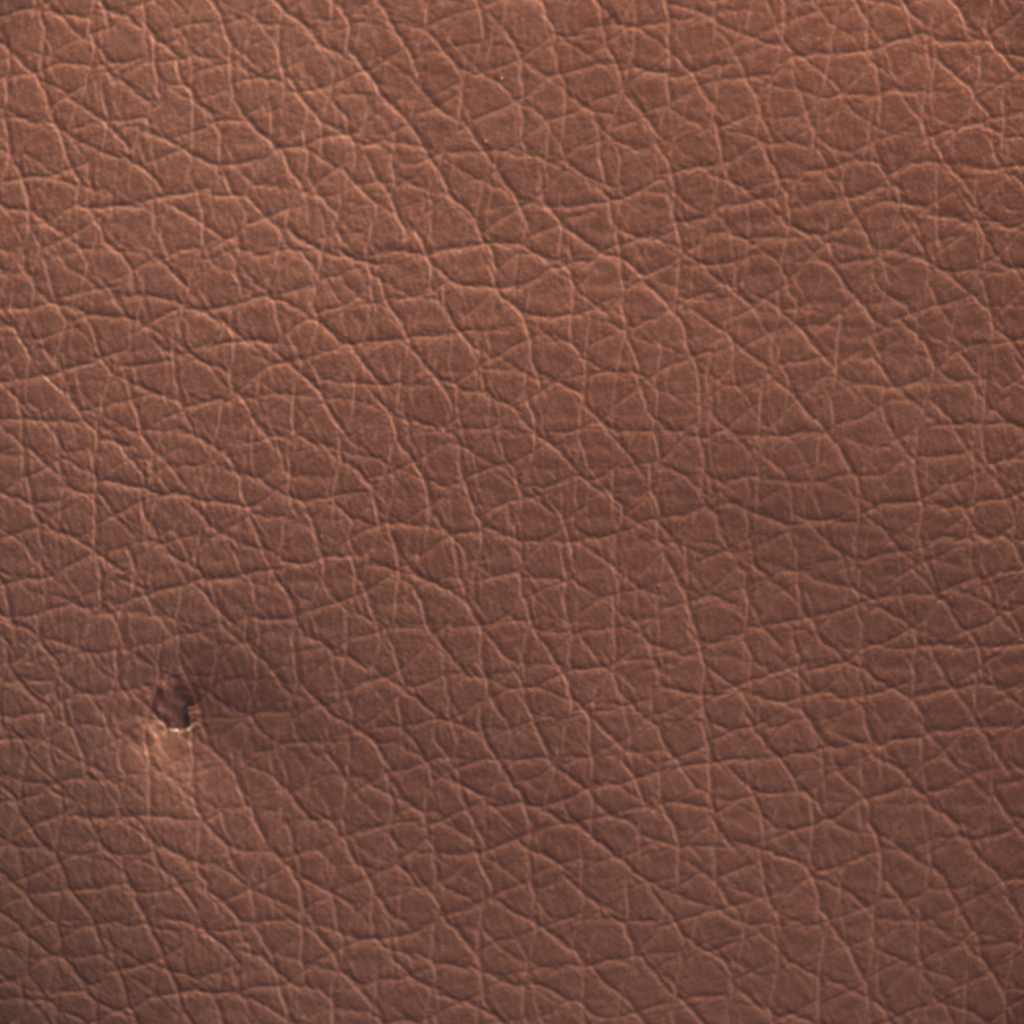}
            \caption{poke}\label{fig:leather-poke}
        \end{subfigure}
    \end{subfigure}
\end{minipage}
\caption{Examples of each class from the MVTec leather dataset.}
\label{fig:leather}
\end{figure}

The MVTec leather dataset \citep{8954181} is originally an anomaly detection dataset, which was reframed as a classification task.
This fourth dataset was evaluated in the same way as the concrete crack dataset and the metal nut dataset.
Each image is of size 700x700 pixels and contains a leather texture. Classes 0 to 5 are color, cut, fold, glue, good, and poke, respectively containing small defects. Example images of each class are presented in Figure \ref{fig:leather}. The numbers of images for each of the classes are imbalanced (e.g., class 0 has 19 images, whereas class 4 has 32 images in total), which is why resampling and data augmentation were used for the training data.

In this appendix, we present the results of each model trained in this dataset, and the class metrics for the \textbf{color} class (where the leather texture has a red mark in it) (See Table \ref{table:performance models leather}).

\begin{table}[ht!]
\caption{Performance metrics of models trained in the leather dataset.} \label{table:performance models leather}
\centering
\begin{tabular}{|lll|lll|lll|}

\hline

\rowcolor[HTML]{9B9B9B} 

\multicolumn{3}{|l|}{\cellcolor[HTML]{9B9B9B}}                                                                            & \multicolumn{3}{c|}{\cellcolor[HTML]{9B9B9B}average}                                                                    & \multicolumn{3}{c|}{\cellcolor[HTML]{9B9B9B}class}                                                                      \\ \hline

\rowcolor[HTML]{9B9B9B} 

\multicolumn{1}{|l|}{\cellcolor[HTML]{9B9B9B}CNN} & \multicolumn{1}{l|}{\cellcolor[HTML]{9B9B9B}seed} & acc & \multicolumn{1}{l|}{\cellcolor[HTML]{9B9B9B}precision} & \multicolumn{1}{l|}{\cellcolor[HTML]{9B9B9B}recall} & f1 & \multicolumn{1}{l|}{\cellcolor[HTML]{9B9B9B}precision} & \multicolumn{1}{l|}{\cellcolor[HTML]{9B9B9B}recall} & f1 \\ \hline

\multicolumn{1}{|l|}{vgg16}                                & \multicolumn{1}{l|}{0}                            & 1.000    & \multicolumn{1}{l|}{1.000}                             & \multicolumn{1}{l|}{1.000}                          & 1.000    & \multicolumn{1}{l|}{1.000}                             & \multicolumn{1}{l|}{1.000}                          & 1.000    \\ \hline

\multicolumn{1}{|l|}{vgg16}                                & \multicolumn{1}{l|}{1}                            & 0.992    & \multicolumn{1}{l|}{0.992}                             & \multicolumn{1}{l|}{0.992}                          & 0.992    & \multicolumn{1}{l|}{0.950}                             & \multicolumn{1}{l|}{1.000}                          & 0.974    \\ \hline

\multicolumn{1}{|l|}{vgg16}                                & \multicolumn{1}{l|}{2}                            & 0.992    & \multicolumn{1}{l|}{0.992}                             & \multicolumn{1}{l|}{0.992}                          & 0.992    & \multicolumn{1}{l|}{0.950}                             & \multicolumn{1}{l|}{1.000}                          & 0.974    \\ \hline

\multicolumn{1}{|l|}{vgg16}                                & \multicolumn{1}{l|}{3}                            & 1.000    & \multicolumn{1}{l|}{1.000}                             & \multicolumn{1}{l|}{1.000}                          & 1.000    & \multicolumn{1}{l|}{1.000}                             & \multicolumn{1}{l|}{1.000}                          & 1.000    \\ \hline

\multicolumn{1}{|l|}{vgg16}                                & \multicolumn{1}{l|}{4}                            & 0.992    & \multicolumn{1}{l|}{0.992}                             & \multicolumn{1}{l|}{0.992}                          & 0.992    & \multicolumn{1}{l|}{0.950}                             & \multicolumn{1}{l|}{1.000}                          & 0.974    \\ \hline

\multicolumn{1}{|l|}{vgg16}                                & \multicolumn{1}{l|}{5}                            & 0.992    & \multicolumn{1}{l|}{0.992}                             & \multicolumn{1}{l|}{0.992}                          & 0.992    & \multicolumn{1}{l|}{0.950}                             & \multicolumn{1}{l|}{1.000}                          & 0.974    \\ \hline

\multicolumn{1}{|l|}{vgg16}                                & \multicolumn{1}{l|}{6}                            & 0.992    & \multicolumn{1}{l|}{0.992}                             & \multicolumn{1}{l|}{0.992}                          & 0.992    & \multicolumn{1}{l|}{0.950}                             & \multicolumn{1}{l|}{1.000}                          & 0.974    \\ \hline

\multicolumn{1}{|l|}{vgg16}                                & \multicolumn{1}{l|}{7}                            & 0.992    & \multicolumn{1}{l|}{0.992}                             & \multicolumn{1}{l|}{0.992}                          & 0.992    & \multicolumn{1}{l|}{0.950}                             & \multicolumn{1}{l|}{1.000}                          & 0.974    \\ \hline

\multicolumn{1}{|l|}{vgg16}                                & \multicolumn{1}{l|}{8}                            & 0.984    & \multicolumn{1}{l|}{0.985}                             & \multicolumn{1}{l|}{0.984}                          & 0.984    & \multicolumn{1}{l|}{0.950}                             & \multicolumn{1}{l|}{1.000}                          & 0.974    \\ \hline

\multicolumn{1}{|l|}{vgg16}                                & \multicolumn{1}{l|}{9}                            & 0.992    & \multicolumn{1}{l|}{0.992}                             & \multicolumn{1}{l|}{0.992}                          & 0.992    & \multicolumn{1}{l|}{0.950}                             & \multicolumn{1}{l|}{1.000}                          & 0.974    \\ \hline

\multicolumn{1}{|l|}{resnet18}                             & \multicolumn{1}{l|}{0}                            & 0.992    & \multicolumn{1}{l|}{0.992}                             & \multicolumn{1}{l|}{0.992}                          & 0.992    & \multicolumn{1}{l|}{0.950}                             & \multicolumn{1}{l|}{1.000}                          & 0.974    \\ \hline

\multicolumn{1}{|l|}{resnet18}                             & \multicolumn{1}{l|}{1}                            & 0.992    & \multicolumn{1}{l|}{0.992}                             & \multicolumn{1}{l|}{0.992}                          & 0.992    & \multicolumn{1}{l|}{0.950}                             & \multicolumn{1}{l|}{1.000}                          & 0.974    \\ \hline

\multicolumn{1}{|l|}{resnet18}                             & \multicolumn{1}{l|}{2}                            & 0.992    & \multicolumn{1}{l|}{0.992}                             & \multicolumn{1}{l|}{0.992}                          & 0.992    & \multicolumn{1}{l|}{0.950}                             & \multicolumn{1}{l|}{1.000}                          & 0.974    \\ \hline

\multicolumn{1}{|l|}{resnet18}                             & \multicolumn{1}{l|}{3}                            & 0.992    & \multicolumn{1}{l|}{0.992}                             & \multicolumn{1}{l|}{0.992}                          & 0.992    & \multicolumn{1}{l|}{0.950}                             & \multicolumn{1}{l|}{1.000}                          & 0.974    \\ \hline

\multicolumn{1}{|l|}{resnet18}                             & \multicolumn{1}{l|}{4}                            & 0.992    & \multicolumn{1}{l|}{0.992}                             & \multicolumn{1}{l|}{0.992}                          & 0.992    & \multicolumn{1}{l|}{0.950}                             & \multicolumn{1}{l|}{1.000}                          & 0.974    \\ \hline

\multicolumn{1}{|l|}{resnet18}                             & \multicolumn{1}{l|}{5}                            & 0.992    & \multicolumn{1}{l|}{0.992}                             & \multicolumn{1}{l|}{0.992}                          & 0.992    & \multicolumn{1}{l|}{0.950}                             & \multicolumn{1}{l|}{1.000}                          & 0.974    \\ \hline

\multicolumn{1}{|l|}{resnet18}                             & \multicolumn{1}{l|}{6}                            & 1.000    & \multicolumn{1}{l|}{1.000}                             & \multicolumn{1}{l|}{1.000}                          & 1.000    & \multicolumn{1}{l|}{1.000}                             & \multicolumn{1}{l|}{1.000}                          & 1.000    \\ \hline

\multicolumn{1}{|l|}{resnet18}                             & \multicolumn{1}{l|}{7}                            & 0.992    & \multicolumn{1}{l|}{0.992}                             & \multicolumn{1}{l|}{0.992}                          & 0.992    & \multicolumn{1}{l|}{0.950}                             & \multicolumn{1}{l|}{1.000}                          & 0.974    \\ \hline

\multicolumn{1}{|l|}{resnet18}                             & \multicolumn{1}{l|}{8}                            & 1.000    & \multicolumn{1}{l|}{1.000}                             & \multicolumn{1}{l|}{1.000}                          & 1.000    & \multicolumn{1}{l|}{1.000}                             & \multicolumn{1}{l|}{1.000}                          & 1.000    \\ \hline

\multicolumn{1}{|l|}{resnet18}                             & \multicolumn{1}{l|}{9}                            & 1.000    & \multicolumn{1}{l|}{1.000}                             & \multicolumn{1}{l|}{1.000}                          & 1.000    & \multicolumn{1}{l|}{1.000}                             & \multicolumn{1}{l|}{1.000}                          & 1.000    \\ \hline

\multicolumn{1}{|l|}{densenet121}                          & \multicolumn{1}{l|}{0}                            & 0.992    & \multicolumn{1}{l|}{0.992}                             & \multicolumn{1}{l|}{0.992}                          & 0.992    & \multicolumn{1}{l|}{0.950}                             & \multicolumn{1}{l|}{1.000}                          & 0.974    \\ \hline

\multicolumn{1}{|l|}{densenet121}                          & \multicolumn{1}{l|}{1}                            & 0.992    & \multicolumn{1}{l|}{0.992}                             & \multicolumn{1}{l|}{0.992}                          & 0.992    & \multicolumn{1}{l|}{0.950}                             & \multicolumn{1}{l|}{1.000}                          & 0.974    \\ \hline

\multicolumn{1}{|l|}{densenet121}                          & \multicolumn{1}{l|}{2}                            & 0.992    & \multicolumn{1}{l|}{0.992}                             & \multicolumn{1}{l|}{0.992}                          & 0.992    & \multicolumn{1}{l|}{0.950}                             & \multicolumn{1}{l|}{1.000}                          & 0.974    \\ \hline

\multicolumn{1}{|l|}{densenet121}                          & \multicolumn{1}{l|}{3}                            & 0.992    & \multicolumn{1}{l|}{0.992}                             & \multicolumn{1}{l|}{0.992}                          & 0.992    & \multicolumn{1}{l|}{0.950}                             & \multicolumn{1}{l|}{1.000}                          & 0.974    \\ \hline

\multicolumn{1}{|l|}{densenet121}                          & \multicolumn{1}{l|}{4}                            & 0.984    & \multicolumn{1}{l|}{0.985}                             & \multicolumn{1}{l|}{0.984}                          & 0.984    & \multicolumn{1}{l|}{0.950}                             & \multicolumn{1}{l|}{1.000}                          & 0.974    \\ \hline

\multicolumn{1}{|l|}{densenet121}                          & \multicolumn{1}{l|}{5}                            & 0.992    & \multicolumn{1}{l|}{0.992}                             & \multicolumn{1}{l|}{0.992}                          & 0.992    & \multicolumn{1}{l|}{0.950}                             & \multicolumn{1}{l|}{1.000}                          & 0.974    \\ \hline

\multicolumn{1}{|l|}{densenet121}                          & \multicolumn{1}{l|}{6}                            & 0.992    & \multicolumn{1}{l|}{0.992}                             & \multicolumn{1}{l|}{0.992}                          & 0.992    & \multicolumn{1}{l|}{0.950}                             & \multicolumn{1}{l|}{1.000}                          & 0.974    \\ \hline

\multicolumn{1}{|l|}{densenet121}                          & \multicolumn{1}{l|}{7}                            & 0.992    & \multicolumn{1}{l|}{0.992}                             & \multicolumn{1}{l|}{0.992}                          & 0.992    & \multicolumn{1}{l|}{0.950}                             & \multicolumn{1}{l|}{1.000}                          & 0.974    \\ \hline

\multicolumn{1}{|l|}{densenet121}                          & \multicolumn{1}{l|}{8}                            & 0.992    & \multicolumn{1}{l|}{0.992}                             & \multicolumn{1}{l|}{0.992}                          & 0.992    & \multicolumn{1}{l|}{0.950}                             & \multicolumn{1}{l|}{1.000}                          & 0.974    \\ \hline

\multicolumn{1}{|l|}{densenet121}                          & \multicolumn{1}{l|}{9}                            & 0.992    & \multicolumn{1}{l|}{0.992}                             & \multicolumn{1}{l|}{0.992}                          & 0.992    & \multicolumn{1}{l|}{0.950}                             & \multicolumn{1}{l|}{1.000}                          & 0.974    \\ \hline

\end{tabular}
\end{table}

Similarly, we provide the results of analyzing every SPACE and ACE run, and aggregating them based on the alignment of the extracted concepts (See Table \ref{table:leather}).

\begin{table}[ht!]
\caption{Experimental results leather dataset, class color.} \label{table:leather}
\centering
\begin{tabular}{|l|l|l|l|l|}

\hline

\rowcolor[HTML]{C0C0C0} 

model        & parameters                            & concepts & aligned & aligned ratio \\ \hline

DenseNet-121 & ACE: $n_{\mathrm{SLIC}}=[15, 50, 80]$ & 9.7      & 0.9     & 0.09          \\ \hline

DenseNet-121 & ACE: $n_{\mathrm{SLIC}}=[15]$         & 3.6      & 0.1     & 0.03          \\ \hline

DenseNet-121 & ACE: $n_{\mathrm{SLIC}}=[200]$        & 23.2     & 0.2     & 0.01          \\ \hline

DenseNet-121 & ACE: $n_{\mathrm{SLIC}}=[50]$         & 11.8     & 0.4     & 0.04          \\ \hline

DenseNet-121 & ACE: $n_{\mathrm{SLIC}}=[80]$         & 16.4     & 0.6     & 0.04          \\ \hline

DenseNet-121 & SPACE: $n_{\mathrm{s}}=5$             & 1.0      & 1.0     & 1.00          \\ \hline

DenseNet-121 & SPACE: $n_{\mathrm{s}}=6$             & 1.7      & 1.1     & 0.70          \\ \hline

DenseNet-121 & SPACE: $n_{\mathrm{s}}=7$             & 1.3      & 0.8     & 0.65          \\ \hline

DenseNet-121 & SPACE: $n_{\mathrm{s}}=8$             & 2.0      & 1.0     & 0.53          \\ \hline

ResNet-18    & ACE: $n_{\mathrm{SLIC}}=[15, 50, 80]$ & 9.1      & 1.0     & 0.11          \\ \hline

ResNet-18    & ACE: $n_{\mathrm{SLIC}}=[15]$         & 3.3      & 0.1     & 0.03          \\ \hline

ResNet-18    & ACE: $n_{\mathrm{SLIC}}=[200]$        & 22.4     & 0.0     & 0.00          \\ \hline

ResNet-18    & ACE: $n_{\mathrm{SLIC}}=[50]$         & 11.9     & 0.5     & 0.04          \\ \hline

ResNet-18    & ACE: $n_{\mathrm{SLIC}}=[80]$         & 15.4     & 0.0     & 0.00          \\ \hline

ResNet-18    & SPACE: $n_{\mathrm{s}}=5$             & 1.6      & 0.7     & 0.45          \\ \hline

ResNet-18    & SPACE: $n_{\mathrm{s}}=6$             & 1.7      & 0.7     & 0.32          \\ \hline

ResNet-18    & SPACE: $n_{\mathrm{s}}=7$             & 1.5      & 0.3     & 0.13          \\ \hline

ResNet-18    & SPACE: $n_{\mathrm{s}}=8$             & 2.0      & 0.7     & 0.30          \\ \hline

VGG-16       & ACE: $n_{\mathrm{SLIC}}=[15, 50, 80]$ & 9.5      & 1.1     & 0.11          \\ \hline

VGG-16       & ACE: $n_{\mathrm{SLIC}}=[15]$         & 3.2      & 0.5     & 0.15          \\ \hline

VGG-16       & ACE: $n_{\mathrm{SLIC}}=[200]$        & 19.0     & 0.0     & 0.00          \\ \hline

VGG-16       & ACE: $n_{\mathrm{SLIC}}=[50]$         & 11.5     & 0.2     & 0.02          \\ \hline

VGG-16       & ACE: $n_{\mathrm{SLIC}}=[80]$         & 14.8     & 0.2     & 0.01          \\ \hline

VGG-16       & SPACE: $n_{\mathrm{s}}=5$             & 1.7      & 0.8     & 0.58          \\ \hline

VGG-16       & SPACE: $n_{\mathrm{s}}=6$             & 2.2      & 1.1     & 0.50          \\ \hline

VGG-16       & SPACE: $n_{\mathrm{s}}=7$             & 2.4      & 1.0     & 0.44          \\ \hline

VGG-16       & SPACE: $n_{\mathrm{s}}=8$             & 3.2      & 1.1     & 0.38          \\ \hline

\end{tabular}
\end{table}

In contrast to other datasets, not all runs of SPACE and ACE were able to extract concepts containing the red marks (characteristics of the class). Specially when comparing the number of aligned concepts extracted through patches of smaller scale ($n_{\mathrm{s}}=8$, $n_{\mathrm{SLIC}}=80$), SPACE outperformed ACE. This phenomenon arises as SPACE prioritizes the important regions of the images. In comparison, ACE rescaling multiplies the impact of the patch geometry, which becomes a common factor of clustering.
\end{appendices}

\end{document}